\documentclass{clv3}
\usepackage[letterpaper]{geometry}
\usepackage{times}
\usepackage{latexsym}
\usepackage{amsmath}
\usepackage{multirow}
\usepackage{url}
\usepackage{nicefrac}

\usepackage{mathtools}
\usepackage{tabularx}
\usepackage{array}
\usepackage{amssymb}
\usepackage{bbold}
\usepackage{float}
\usepackage[export]{adjustbox}
\usepackage[caption = false]{subfig}
\usepackage{array,url,kantlipsum}
\usepackage{verbatim}
\usepackage[scientific-notation=true]{siunitx}
\usepackage[inline]{enumitem}
\usepackage{makecell}
\usepackage{enumitem}
\usepackage{algorithm}
\usepackage{algpseudocode}
\usepackage[mathscr]{euscript}
\usepackage[cal=boondoxo]{mathalfa}
\usepackage{pgfplots}
\usepackage{pgfplotstable}
\pgfplotsset{compat=1.11,
  /pgfplots/ybar legend/.style={
    /pgfplots/legend image code/.code={%
      \draw[##1,/tikz/.cd,bar width=2pt,yshift=-0.3em,bar shift=0pt]
    plot coordinates {(0cm,0.8em)};},
  },
}

\usepackage{mathrsfs}
\usepackage{amsbsy}

\newcommand{\RNum}[1]{\uppercase\expandafter{\romannumeral #1\relax}}
\newcommand{\TABLE}[1]{Table~\ref{#1}}
\newcommand{\SEC}[1]{Section~\ref{#1}}
\newcommand{\FIG}[1]{Figure~\ref{#1}}
\newcommand{\EQ}[1]{Equation~\ref{#1}}
\newcommand{\ALG}[1]{Algorithm~\ref{#1}}

\DeclarePairedDelimiter\abs{\lvert}{\rvert}

\newcommand{\eg}{e.g., }
\newcommand{\ie}{i.e., }

\newcolumntype{Y}{>{\centering\arraybackslash}X}
\makeatletter
\newcommand{\mypm}{\mathbin{\mathpalette\@mypm\relax}}
\newcommand{\@mypm}[2]{\ooalign{%
    \raisebox{.1\height}{$#1+$}\cr
    \smash{\raisebox{-.6\height}{$#1-$}}\cr}}
\makeatother

\dochead{Principal Word Vectors}
\runningtitle{Word Embedding through Principal Component Analysis}

\runningauthor{Basirat}

\begin{document}

\title{Principal Word Vectors}


\maketitle
\begin{abstract}
  We generalize principal component analysis for embedding words into a vector space. 
  The generalization is made in two major levels. 
  The first is to generalize the concept of corpus as a counting process which is defined by three key elements vocabulary set, feature (annotation) set, and context.
  This generalization enables the principal word embedding method to generate word vectors with regard to different types of contexts and different types of annotations provided for a corpus. 
  The second is to generalize the transformation step used in most of the word embedding methods. 
  To this end, we define two levels of transformations.
  The first is a quadratic transformation, which accounts for different types of weighting over the vocabulary units and contextual features. 
  Second is an adaptive non-linear transformation, which reshapes the data distribution to be meaningful to principal component analysis. 
  The effect of these generalizations on the word vectors are intrinsically studied with regard to the spread and the discriminability of the word vectors. 
  We also provide an extrinsic evaluation on the contribution of the principal word vectors on a word similarity benchmark and the task of dependency parsing. 
  Our experiments are finalized by a comparison between the principal word vectors and other sets of word vectors generated with popular word embedding methods. 
  The results obtained from our intrinsic evaluation metrics show that the spread and the discriminability of the principal word vectors are higher than that of other word embedding methods. 
  The results obtained from the extrinsic evaluation metrics show that the principal word vectors are better than some of the word embedding methods and on par with popular methods of word embedding. 
\end{abstract}

\section{Introduction}
\label{sec:introduction}

The distributional representation of words, also known as word embeddings or word vectors, is a fundamental technique to most modern approaches of the natural language processing \cite{collobert2011natural, chen14, dyer:2015acl} 
In this representation, words are modelled through real-valued feature vectors that capture global syntactic and semantic dependencies between words in a corpus. 
This representation enables the application of powerful machine learning techniques, developed for continuous data, on the discrete and symbolic observations of words.  

Several word embedding techniques have been proposed by researchers \cite{schutze1992dimensions,lund1996producing,landauer1997solution,sahlgren2006word,collobert2011natural,Mikolov13vector_space,pennington2014glove,lebret-collobert:2014:EACL}.
\namecite{lebret-collobert:2014:EACL} claim that ``a simple spectral method as PCA can generate word embeddings as good as with deep-learning  architectures''.
In their method, word vectors are basically the principal components of a probability co-occurrence matrix which undergoes a Hellinger transformation. 
\namecite{basirat-nivre:2017:NoDaLiDa} show that the Hellinger transformation is not the best transformation for certain tasks.
However, they do not provide any systematic method to find the best transformation function. 

In this paper, we provide a solution to this problem and generalize the main idea of word embedding through PCA in two ways.
The first is to give the method enough flexibility to take different types of contexts and contextual features into account. 
This lead us to formulate a concept of contextual word vector that encodes both the contexts and the contextual features of a word into a vector. 
The second is to use generalized principal component analysis to provide the method with two levels of transformation, 
a quadratic transformation, which provides for a weighting mechanism over words and features, followed by an adaptive non-linear transformation that reshapes the distribution of contextual word vectors. 
The generalized principal word vectors are built through the singular value decomposition of a mean centred sample matrix of contextual word vectors. 
Depending on the size of the contextual features and vocabulary set, this matrix can be very large and computationally very expensive to process. 
We adopt the randomized method of singular value decomposition proposed by \namecite{Halko2011finding} and adapt it to perform the matrix factorization in an efficient way. 
The proposed method is able to preserve the sparsity of the data and estimate the singular value decomposition of the mean centred data without explicitly constructing the mean centred data matrix which, is inherently a dense matrix and needs a lot of memory. 
These generalizations contribute to an efficient and flexible word embedding method, called principal word embedding, which is able to generate a set of word vectors, called principal word vectors, from a raw or annotated corpus. 

We organize this paper as follows.
In \SEC{sec:related_work}, we outline the previous work done in this area and explain how the previous methods are connected to the principal word embedding. 
The contextual word vectors and their distribution are studied in \SEC{sec:contextual_word_vector}.
This is followed by a short study of generalized principal component analysis in \SEC{sec:pca}. 
The principal word vectors and their parameters are introduced in \SEC{sec:princ_word_vec}.
\SEC{sec:exp_setting} describes our experimental settings and evaluation metrics which are then used in \SEC{sec:experiments} to study and evaluate the principal word vectors in two major ways.
First, we study the effective parameters of principal word embedding such as contextual features, weighting mechanism, and the transformation function. 
This study is on the basis of the spread and discriminability of the word vectors.
Next, we compare the performance of word vectors in different tasks with other popular methods of word embeddings. 
The comparisons are on the basis of the contribution of word vectors in word similarity benchmark \cite{faruqui-2014:SystemDemo}, and in the task of dependency parsing \cite{nivre2004incrementality}. 

\section{Related Work} 
\label{sec:related_work}
The distributional representation of words is developed in the area of distributional semantics \cite{schutze1992dimensions,lund1996producing,landauer1997solution,sahlgren2006word,pennington2014glove,lebret-collobert:2014:EACL,basirat-nivre:2017:NoDaLiDa}, 
and in the area of language modeling \cite{collobert2011natural,Mikolov13vector_space,NIPS2013_5021}.
\namecite{levy2014neural} show that the algorithms developed in both areas are highly connected to each other. 
The general idea shared between the two areas is that similar words tend to appear in similar contexts. 
In a bird's eye view, the word embedding algorithms generate a set of word vectors through the application of dimensionality reduction techniques to a co-occurrence matrix, 
which counts the frequency of words in different contexts. 
The co-occurrence matrix is the key data structure in this approach that encodes the contextual environments of words. 
This matrix is built either explicitly \cite{pennington2014glove,lebret-collobert:2014:EACL,basirat-nivre:2017:NoDaLiDa} or implicitly \cite{collobert2011natural,Mikolov13vector_space,sahlgren2006word} during scanning a training corpus. 

Principal word embedding is rooted in the methods that are developed in the area of distributional semantics. 
In other words, the principal word vectors can be considered as vectors in a distributional semantic space. 
A distributional semantic space is a finite dimensional linear space whose dimensions correspond to the contextual environment of words. 
Word similarities in a distributional semantic space are reflected through the similarities between vectors associated with them. 
The remainder of this section is devoted to a review of the word embedding methods developed in the area of distributional semantics and an elaboration of how these methods are connected to the principal word embedding method.

In distributional semantics, the process of embedding words into a vector space consists of three main steps.
The first is to scan a corpus and build a co-occurrence matrix which counts the frequency of seeing words in different contexts.
The columns of a co-occurrence matrix are associated with words and represent the frequency of seeing words with different contextual features corresponding to the rows of the matrix.
So each column can be seen as a vector representation of its corresponding word.
The second is to perform a transformation on the elements of the co-occurrence matrix. 
The third is to extract a set of low dimensional vectors from the column vectors which depending on the number of context units can be high dimensional. 
The definition of context is a key factor in the type of information encoded into the column vectors formed in the first step. 
The two major types of contexts used in the literature are the window-based context and the dependency context.
A window-based context is formed by all words in a sequence (or window) of surrounding words. 
A dependency-based context is formed by words that are in certain dependency relation with the word in interest. 
The relative position of words is either completely ignored by these types of context or in some cases they are modelled through some weighting mechanisms with respect to the positional distance between the words in a sentence. 

The window-based context is used in many word embedding methods. 
Among these methods are \texttt{HAL} \cite{lund1996producing}, \texttt{word2vec} \cite{Mikolov13vector_space}, \texttt{HPCA} \cite{lebret-collobert:2014:EACL}, and \texttt{GloVe} \cite{pennington2014glove}. 
These methods are designed to process raw corpora and the contextual features in these methods are basically word forms. 
\namecite{pado2007depssm} define the context of a word in a sentence as all words in the sentence which are in a dependency relation with the target word. 
This idea is further developed by \namecite{levy-goldberg:2014:P14-2} where they generalize the skip-gram model used in \texttt{word2vec} with dependency context. 
Although \namecite{pado2007depssm} show that the dependency context can result in higher accuracies in certain tasks, \namecite{kiela2014systematic} argue that better results can be obtained from the window-based context than the dependency context if the vectors are extracted from a fairly large corpus with a small window size. 
The idea of using rich contextual features is explored by \namecite{kiela2014systematic}.
They use different types of contextual features such as lemma, part-of-speech tags, and CCG supertags.
This can also be seen in \texttt{LSA} \cite{landauer1997solution} where the contextual feature and the context of a word are the label and the content of the document it belongs to.
The principal word embedding generalizes the window-based context and dependency context to the \emph{context function}.
In addition, it defines the concept \emph{feature variable} that links context function and contextual features together. 
This gives the word embedding method enough flexibility to make use of different types of contexts and contextual features provided by both raw and annotated corpora. 

The goal of the transformation step is to reduce the dominance of high-frequency words in the co-occurrence matrix. 
The logarithm function is among the transformation functions which is widely used for this aim. 
\namecite{salton1988term} propose using a weighting mechanism called \emph{tf-idf} to reduce the negative effect of the most frequently used words. 
The \emph{tf-idf} weighting mechanism is then generalized to \emph{tf-icf} \cite{tf-icf:4041501}. 
Pointwise mutual information \cite{church1990word} combines weighting with a logarithmic transformation. 
\namecite{levy2014neural} and \namecite{melamud-goldberger:2017:Short} provide a detailed study about the effect of pointwise mutual transformation on the word embedding. 
\cite{lebret-collobert:2014:EACL} propose to use the Hellinger transformation on a \emph{probability} co-occurrence matrix. 
The principal word embedding provides a systematic mechanism that allows us to define different types of weighting approaches together with non-linear transformation function. 
The weighting is carried out through two weight matrices which affect the contextual features and the words. 
In \SEC{subsubsection:weight_matrices}, we show that different weighting methods in the literature are special cases of our approach.
The non-linear transformation step in principal word embedding is not restricted to a single function such as logarithmic and square root transformation.
The principal word embedding uses an adaptive method of transformation that is able to reshape the data distribution taking into consideration the distribution of the training data. 

Among the popular methods of dimensionality reduction, principal component analysis is extensively used to reduce the dimensionality of column vectors in a co-occurrence matrix \cite{lund1996producing,landauer1997solution,lebret-collobert:2014:EACL,basirat-nivre:2017:NoDaLiDa}. 
\namecite{ICML2012Dahl_364} and \namecite{Hinton504} use the restricted Boltzmann Machine and \namecite{Mikolov13vector_space} use an auto-encoder for this aim. 
It can be shown that both auto-encoders and restricted Boltzmann machine are connected to the principal component analysis \cite{jolliffe2002principal,Hinton504}.
Mixture models \cite{hofmann1999probabilistic,blei2003latent}, and non-linear methods \cite{Roweis2323LLE,Hinton504} are among the other approaches that have been used for reducing the dimensionality of a semantic space. 
\texttt{GloVe} \cite{pennington2014glove} formulates the problem of dimensionality reduction as a regression problem. 
Although \texttt{GloVe} does not explicitly use PCA, \namecite{basirat-nivre:2017:NoDaLiDa} show that the regression formulation is equivalent to the kernel principal component analysis of the column vectors.
\texttt{HPCA} \cite{lebret-collobert:2014:EACL} uses the singular value decomposition to estimate the principal components of the co-occurrence matrix. 
\namecite{lebret-collobert:2014:EACL} ignore the centring step in PCA in order to take the advantage of the data sparsity. 
\texttt{RSV} \cite{basirat-nivre:2017:NoDaLiDa} uses the same approach to compute the principal components.
However, it performs the mean subtraction step to centre the column vectors around their mean. 
This results in a dense matrix which require a lot of memory space. 
The principal word embedding uses a modified version of the randomised SVD algorithm \cite{Halko2011finding} to estimate the principal word vectors. 
The modified algorithm is able to estimate the singular factors of the \emph{mean centred} co-occurrence matrix without performing the actual mean subtraction. 
This enables the algorithm to take advantage of the sparsity of the data and estimate the principal word vectors from the mean centred data. 

\section{Contextual Word Vector}
\label{sec:contextual_word_vector}
In this section, we introduce contextual word vectors as the initial frequency-based representation of words in a corpus and study their distribution as a mixture model. 
These vectors then undergo the principal component analysis to construct the low-dimensional principal word vectors. 
A contextual word vector associated with a word is a real-valued vector whose elements are the frequency of seeing the word in different contexts formed in a corpus.
This definition has three main concepts: \emph{word}, \emph{context}, and \emph{corpus}.
The word is one the most fundamental concepts in linguistics. 
In terms of syntax, the word is defined as the minimum syntactic unit of language \cite{syntax_book_Matthews}. 
Any arrangement of words that follow certain grammatical rules forms a sentence. 
The set of words in a language forms a vocabulary set and a set of sentences, sampled from all possible sentences in a language and indexed by natural numbers, forms a corpus.\footnote{The indexing represents the order of sentences in the corpus. In other words, a corpus is a subset of $S\times\mathbb{N}$, where $S$ is the set of all possible sentences in a language.}
Corpora are released in raw or annotated forms. 
A raw corpus consists of a collection of sentences with or without word and sentence boundaries. 
An annotated corpus provides some abstract information about the linguistic units such as characters, words, sentences, and documents in the corpus.
This information is usually represented through some symbols which are associated with the elements of the corpus. 
For example, a corpus can be annotated with part-of-speech tags at the word level, or document genre at the document level. 
Regardless of the type of annotation, we refer to a set of annotation symbols used in a corpus as a \emph{feature} set. 
In this research, we do not distinguish between raw corpora and annotated corpora. 
We consider a raw corpus as an annotated corpus whose words are annotated by their forms. 
In addition, we consider the word as the most basic element of a corpus and generalize the annotation information provided for the more abstract units such as sentences and documents to their constituent words.
For example, a genre associated with a document in a corpus is associated with all words forming the document. 
This generalization allow us to develop a model of word embedding that can benefit from different levels of annotation provided by the corpora. 
It also makes it possible to formalize a corpus with a set of vocabulary and a set of annotation features. 
Denoting a vocabulary set consisting of $n$ words as $V = \{v_1, \dots, v_n\}$, and  a feature set consisting of $m$ features as $F=\{f_1, \dots, f_m\}$, 
the corpus $E$ of size $T$ is defined as a set of triples (word, feature, and index), $E=\{e_1,\dots,e_T\}$ where $e_i\in V\times F\times \mathbb{N}$. 
The index $i$ in each triple $e_i\in E$ represents the relative position of words in the corpus. 
We use the vocabulary lookup function $\mathcal{V}:E\to V$, the feature lookup function $\mathcal{F}:E\to F$, and the index function $\mathcal{I}:E\to\mathbb{N}$, to map each element $e\in E$ onto its corresponding entries in $V$ and $F$, and $\mathbb{N}$ respectively.
In other words, each element $e_i\in E$ is a triple $(v,f,i)$ such that $\mathcal{V}(e_i)=v$, $\mathcal{F}(e_i)=f$, and $\mathcal{I}(e_i)=i$. 

We define context as a certain type of connection between the elements of a corpus. 
For example, at the surface level, one can define a neighbourhood connection between words (\eg $e_{i-1}$ is the immediate preceding of $e_i$).
Another example is at the syntax level, where a dependency connection is defined between the elements of a corpus that form a sentence. 
One can also extend the domain of connections to discourse and document. 
More formally, we define the context of each element $e\in E$ as a function $\mathcal{C}:E\to \mathcal{P}(E)$, where $\mathcal{P}(E)$ is the power set of $E$. 
This function is referred to as context function or context for short. 
In this definition, the context function connects each element in the corpus to a set of elements in the same corpus. 
The power function in the range of context gives it enough flexibility to model different types of connections between the corpus elements.
A limited variant of context function is the singleton context which maps individual elements onto each other \ie $E$, \ie $\mathcal{C}:E\to E$. 
In other words, the singleton context of a word in a corpus is another word in the same corpus. 
In the remainder of this section, we use the definition of singleton context to formulate the concept of contextual word vector. 
Later on, in \SEC{subsec:feature_variables}, we elaborate on how more complicated contexts can be formed from multiple singleton contexts. 

Given the singleton context function $\mathcal{C}:E\to E$, we associate each feature $f_i\in F$ with a Bernoulli variable $\mathbf{f}_i(e_t)$, called \emph{feature variable}, whose values for the given word $e_t\in E$ is defined as below:
\begin{equation}
  \mathbf{f}_i(e_t) = 
  \begin{cases}
    1 & \mathcal{F}(\mathcal{C}(e_t)) = f_i \\
    0 & \mbox{otherwise}
  \end{cases}
  \label{eq:context_variable}
\end{equation}
Similarly, we associate each word $v_j\in V$ with a Bernoulli random variable $\mathbf{v}_j(e_t)$, called \emph{word variable}, whose values for the given word $e_t\in E$ is defined as below:
\begin{equation}
  \mathbf{v}_j(e_t) = 
  \begin{cases}
    1 & \mathcal{V}(e_t) = v_j \\
    0 & \mbox{otherwise}
  \end{cases}
  \label{eq:word_variable}
\end{equation}

For a given set of feature variables $\mathbf{F}=\{\mathbf{f}_1,\dots,\mathbf{f}_m\}$ corresponding to the feature set $F$, 
a set of word variables $\mathbf{V}=\{\mathbf{v}_1,\dots,\mathbf{v}_n\}$ corresponding to the vocabulary set $V$, and 
a training corpus $E=\{e_1, \dots, e_T\}$, 
we define the contextual word vector $\boldsymbol{\mathcal{v}}^{(j)}$ associated with word $v_j\in V$ as an $m$-dimensional random vector whose $i$th element ($i=1,\dots,m$) is the frequency of seeing feature $f_i\in F$ in the context of the word $v_j\in V$ in the training corpus $E$:
\begin{equation}
  \boldsymbol{\mathcal{v}}^{(j)}_i=\sum_{t=1}^{T}\mathbf{f}_i(e_t)\mathbf{v}_j(e_t)
  \label{eq:element_of_cwvec}
\end{equation}
Each element $\boldsymbol{\mathcal{v}}^{(j)}_i$ of the contextual word vector follows a binomial distribution as below:
\begin{equation}
  \boldsymbol{\mathcal{v}}^{(j)}_i \sim \mbox{B}(n(v_j), p(f_i|v_j))
  \label{eq:binomial_dist}
\end{equation}
where $n(v_j)=\sum_{t=1}^T\mathbf{v}_j(e_t)$ is the marginal frequency of seeing $v_j$ in $E$, and $p(f_i|v_j) = p(\mathbf{f}_i = 1 | \mathbf{v}_j = 1)$.
This is because the product $\mathbf{f}_i(e_t)\mathbf{v}_j(e_t)$ in \EQ{eq:element_of_cwvec} follows the Bernoulli distribution, therefore its sum is a binomial random variable. 
So the contextual word vectors $\boldsymbol{\mathcal{v}}^{(j)}$ is a vector of the binomial random variable, \ie $\boldsymbol{\mathcal{v}}^{(j)}=(\boldsymbol{\mathcal{v}}^{(j)}_1,\dots,\boldsymbol{\mathcal{v}}^{(j)}_m)$, with the restriction $\Sigma_{k=1}^{m}\mathcal{v}^{(j)}_k=n(v_j)$, where $n(v_j)$ is the marginal frequency of $v_j$ in the corpus. 
This shows that the contextual word vector $\boldsymbol{\mathcal{v}}^{(j)}$ follow the multinomial distribution as below:
\begin{equation}
  f(\mathcal{v}^{(j)}_1,\dots,\mathcal{v}^{(j)}_m;n(v_j),p(f_1|v_j),\dots,p(f_m|v_j)) = \binom{n(v_j)}{\mathcal{v}^{(j)}_1,\dots,\mathcal{v}^{(j)}_m}\prod_{i=1}^{m}p(f_i|v_j)^{\mathcal{v}^{(j)}_i}
  \label{eq:multinomial_contextual_word_vector}
\end{equation}
where $\mathcal{v}^{(j)}=(\mathcal{v}^{(j)}_1,\dots,\mathcal{v}^{(j)}_m)^T$ is a realization of $\boldsymbol{\mathcal{v}}^{(j)}$ \ie $\mathcal{v}^{(j)}_i\in\mathbb{Z}^+\cup\{0\}$ and $\sum_{i=1}^{m} \mathcal{v}^{(j)}_i = n(v_j)$ ($i=1,\dots,m$).
In the following, we use the term contextual word vector to refer to the vector of random variables denoted by $\boldsymbol{\mathcal{v}}$ and to its realization denoted by $\mathcal{v}$. 
The distinction between the two concepts is made through the notation, bold versus normal format. 

We represent the set of contextual word vectors associated with all words in the vocabulary set through a mixture model. 
This mixture model then undergo the principal component analysis to form the low-dimensional word vectors. 
Given that each contextual word vector follows a multinomial distribution which is characterized by the occurrence of the corresponding word in the corpus, 
the set of contextual word vectors $\{\boldsymbol{\mathcal{v}}^{(1)},\dots,\boldsymbol{\mathcal{v}}^{(n)}\}$ associated with all words in the vocabulary set $V$ forms a mixture model with the following probability mass function:
\begin{equation}
  p( \boldsymbol{\mathcal{v}}=(\mathcal{v}_1,\dots,\mathcal{v}_m) ) = \sum_{j=1}^{n} p(v_j)f(\mathcal{v}_1,\dots,\mathcal{v}_m;n(v_j),p(f_1|v_j),\dots,p(f_m|v_j))
  \label{eq:mixture_model}
\end{equation}
where $p(v_j) = p(\mathbf{v}_j = 1)$ is the marginal distribution of the word $v_j$ in the corpus. 
We refer to $\boldsymbol{\mathcal{v}}$ as the mixture of contextual word vectors. 
The mixture of contextual word vectors weights each contextual word vector with the frequency of its corresponding word in the corpus. 
In order to have a better view of this mixture model, we estimate the overall mean vector and covariance matrix of the distribution in \EQ{eq:mixture_model}. 

The mean vector of the mixture of contextual word vectors with the distribution in \EQ{eq:mixture_model} is 
$\boldsymbol{\mu}^{\boldsymbol{\mathcal{v}}}=\sum_{k=1}^{n}p(v_k)\boldsymbol{\mu}^{\boldsymbol{\mathcal{v}}^{(k)}}$. 
Given that $p(v_k)$ follows the Zipfian distribution, we have $p(v_k)\approx\frac{C}{k}$ where $k$ is the word index in a list of words sorted in a descending order of their frequencies and $C=\frac{1}{\sum_{k=1}^{n}{k^{-1}}}$.
Using the Zipf's law and the fact that $\boldsymbol{\mathcal{v}}^{(k)}$ follows a binomial distribution (see \EQ{eq:binomial_dist}), the $i$th element of the mean vector $\boldsymbol{\mu}^{\boldsymbol{\mathcal{v}}}$ will be as below:
\begin{equation}
  \begin{split}
    \boldsymbol{\mu}^{\boldsymbol{\mathcal{v}}}_i 
      & = \sum_{k=1}^{n}p(v_k)\boldsymbol{\mu}^{\boldsymbol{\mathcal{v}}^{(k)}} \\
      & = \sum_{k=1}^{n}n(v_k)p(v_k)p(f_i|v_k) \\
      & \approx TC^2\sum_{k=1}^{n}\frac{p(f_i|v_k)}{k^2} \\
  \end{split}
  \label{eq:expectation_of_word_vectors_zipf}
\end{equation}
where we use $\boldsymbol{\mu}^{\boldsymbol{\mathcal{v}}^{(k)}}=n(v_k)p(f_i|v_k)$, $p(v_k)\approx\frac{C}{k}$, and $n(v_k)\approx\frac{TC}{k}$ and $T$ is the corpus size.
The product $TC^2$ in \EQ{eq:expectation_of_word_vectors_zipf} can be approximated by $\frac{T}{(\log{n})^2}$, 
where we approximate $C\approx\frac{1}{\log{n}}$, using the harmonic series approximation $\sum_{k=1}^{n}{k^{-1}}\approx \log{n}$.
So, the $i$th elements of the mean vectors $\boldsymbol{\mu}^{\boldsymbol{\mathcal{v}}}$ can be approximated by
\begin{equation}
  \boldsymbol{\mu}^{\boldsymbol{\mathcal{v}}}_i \approx \frac{T}{(\log{n})^2}\sum_{k=1}^{n}\frac{p(f_i|v_k)}{k^2}
  \label{eq:expectation_of_word_vectors_zipf_approx}
\end{equation}
\EQ{eq:expectation_of_word_vectors_zipf_approx} shows that the elements of the mean vector are affected by three factors: the corpus $T$, the vocabulary size $n$, and the distribution of features over words $p(f_i|v_k)$. 
We study the effect of these parameters on the elements of mean vector in two steps.
First we study the parameters $T$ and $n$ then we show how the mean value is affected by the distributions of features over words. 

The parameters $T$ and $n$ are dependent on each other \cite{ASI:ASI20524}. 
A large corpus is expected to have large vocabulary set too. 
On the basis of the Heaps' law, the number of distinct words in a corpus is a function of the corpus. 
Using the same notation as before, the Heaps' law relate the corpus size $T$ and the vocabulary size $n$ as $T=Kn^\beta$, where $10\le K \le 100$ is an integer and $\beta\in(0,1)$.
\FIG{fig:tau_vs_n} shows how the values of $\tau=\frac{T}{(\log{n})^2}$ appeared in \EQ{eq:expectation_of_word_vectors_zipf_approx} vary with respect to $n$. 
Unsurprisingly, the value of $\tau$ increases as the vocabulary size $n$ increases. 
This shows that the elements of the mean vector of mixture of contextual word vectors are highly affected by the product $\tau=\frac{T}{(\log{n})^2}$ in \EQ{eq:expectation_of_word_vectors_zipf_approx}.
\begin{figure}[h!]
  \begin{center}
    \begin{tikzpicture}[trim axis left]
      \begin{axis}[
          domain=1000:100000,
          samples=100,
          enlarge x limits=false,
          xlabel={$n$},
          ylabel={$\tau$},
          no markers,
        ]
        \addplot +[thick] {100*(x)^(0.6)/(ln(x))^2};
      \end{axis}
    \end{tikzpicture}
  \end{center}
  \caption{The variation of $\tau=\frac{T}{(\log{n})^2}$ versus $n$, where we use Heaps' law to estimate $T\approx Kn^\beta$ with $K=100$, and $\beta=0.6$.}
  \label{fig:tau_vs_n}
\end{figure}

Now we study how the element of the mean vector of mixture of contextual word vectors are affected by the distribution of features over words. 
For a fixed value of $T$ and $n$, \EQ{eq:expectation_of_word_vectors_zipf_approx} shows that the mean vector strictly depends on the distribution of the features over the words. 
The presence of the $k^2$ in the denominator of $\frac{p(f_i|v_k)}{k^2}$ in \EQ{eq:expectation_of_word_vectors_zipf_approx} shows that the mean vector is more affected with the words with small values of $k$, 
the most frequent words. 
In other words, the mean values of those features which are seen with the most frequent words are reasonably higher than those features which are seen with the less frequent words.
In many practical cases, where the size of the feature set is large enough (\eg word forms or n-grams of word forms are used as features), the distribution of the features over words are expected to be close to the Zipfian distribution. 
In these cases, we have $p(f_i|v_k)\approx\frac{I}{i}$, where $i$ is the index of feature $f_i$ in the list of features sorted in the descending order of their frequency seen with $v_j$ and $I$ is the probability normalizing factor with the value of $I=\frac{1}{\sum_{i=1}^{m}i^{-1}}$. 
Replacing the Zipfian approximation of $p(f_i|v_k)\approx\frac{I}{i}$ in to the \EQ{eq:expectation_of_word_vectors_zipf_approx}, we have 
\begin{equation}
  \begin{split}
    \boldsymbol{\mu}^{\boldsymbol{\mathcal{v}}}_i 
      & \approx \frac{TI}{i(\log{n})^2}\sum_{k=1}^{n}\frac{1}{k^2} \\
      & \approx \frac{\pi^2T}{6i\log{m}(\log{n})^2} \\
  \end{split}
  \label{eq:expectation_of_word_vectors_zipf_zipf_approx}
\end{equation}
where we use $I\approx\frac{1}{\log{m}}$ and $\sum_{k=1}^{n}{k^{-2}}\approx\frac{\pi^2}{6}$.
\EQ{eq:expectation_of_word_vectors_zipf_zipf_approx} shows that if the features are distributed with the Zipfian distribution over the words then the elements of the mean vector of the mixture of contextual word vectors are directly proportional to the corpus size $T$ and inversely proportional to the logarithm of the size of the feature set and the logarithm of the size of the vocabulary set. 
The effect of the vocabulary size is to a large extent cancelled out by the presence of the corpus size $T$ in the nominator of the fraction, since the vocabulary size $n$ is directly proportional to the corpus size $T$. 
As we mentioned above, on the basis of the Heaps' law, the corpus size is still larger than the vocabulary size. 
So we cannot completely eliminate the effect of these parameters on the elements of the mean vectors. 
Nevertheless, it would be safe to conclude that as we increase the size of the feature set, the mean vector will be more close to zero. 
The presence of index $i$ in denominator of \EQ{eq:expectation_of_word_vectors_zipf_zipf_approx} shows that the $\boldsymbol{\mu}^{\boldsymbol{\mathcal{v}}}_i$ becomes smaller as $i$ increases. 
This basically means that the elements of the mean vector related to the most frequent features, for which $i$ is small, are much higher than the other features.

Nevertheless, the assumption about the Zipfian distribution of features over words might not be a valid assumption for all circumstances. 
For example, if a small set of part-of-speech tags is used as the feature set, then the distribution is more likely not to follows the Zipfian distribution and be more close to the uniform distribution. 
In order to estimate the mean values in a more general way, we use the maximum entropy principle which states that the best distribution to model an observed the data set is one with maximum entropy. 
When we have no knowledge about the data, the best distribution to model them is the uniform distribution. 
Using the maximum entropy principle, if we assume that feature $f_i$ is uniformly distributed over all words with probability $p(f_i|v_k)\approx\frac{1}{m}$, where $m$ is the size of the feature set, the $i$th element of the mean vector in \EQ{eq:expectation_of_word_vectors_zipf_zipf_approx} is approximated as below:
\begin{equation}
  \boldsymbol{\mu}^{\boldsymbol{\mathcal{v}}}_i \approx \frac{T\pi^2}{6m(\log{n})^2}
  \label{eq:mean_cwvec_uniform_dist}
\end{equation}
where we use $\sum_{k=1}^{n}{k^{-2}}\approx\frac{\pi^2}{6}$, and $C\approx\frac{1}{\log{n}}$.
\begin{figure}[h!]
  \begin{center}
    \begin{tikzpicture}[trim axis left]
      \begin{axis}[
          domain=1000:100000,
          samples=100,
          enlarge x limits=false,
          xlabel={$n$},
          ylabel={$\boldsymbol{\mu}^{\boldsymbol{\mathcal{v}}}_i$},
          no markers,
        ]
        \addplot +[thick] {100*(x)^(0.6)*(3.14^2)/(6*x*(ln(x))^2)};
      \end{axis}
    \end{tikzpicture}
  \end{center}
  \caption{An approximiation of the elements of the mean vector $\boldsymbol{\mu}^{\boldsymbol{\mathcal{v}}}_i\approx \frac{T\pi^2}{6n(\log{n})^2}$ with respect to the size of the feature set and the size of the vocabulary set which are assume to be equal, $n$. The corpus size $T$ is approximated on the basis of the Heaps' law $T\approx Kn^\beta$ with $K=100$, and $\beta=0.6$.}
  \label{fig:elemt_of_mean_of_cwvec}
\end{figure}
In order to show how the elements of the mean vector vary with respect to the corpus size $T$ and the vocabulary set $n$, we assume that the size of feature set, $m$, is equal to the size of the vocabulary set, $n$. 
This basically means that the corpus is a raw corpus, \ie the features are the word forms.
In this case the value of $\boldsymbol{\mu}^{\boldsymbol{\mathcal{v}}}_i$ can be approximated by $\frac{T\pi^2}{6n(\log{n})^2}$.
\FIG{fig:elemt_of_mean_of_cwvec} shows how the values of $\boldsymbol{\mu}^{\boldsymbol{\mathcal{v}}}_i$ vary with respect to $n$. 
In general, the figure shows that the value of $\boldsymbol{\mu}^{\boldsymbol{\mathcal{v}}}_i$ are so small even with a relatively small set of the vocabulary set.
This means that the mean vector of contextual word vectors are close to the null vector. 

\begin{figure}[h!]
  \centering
  \includegraphics[scale=0.5]{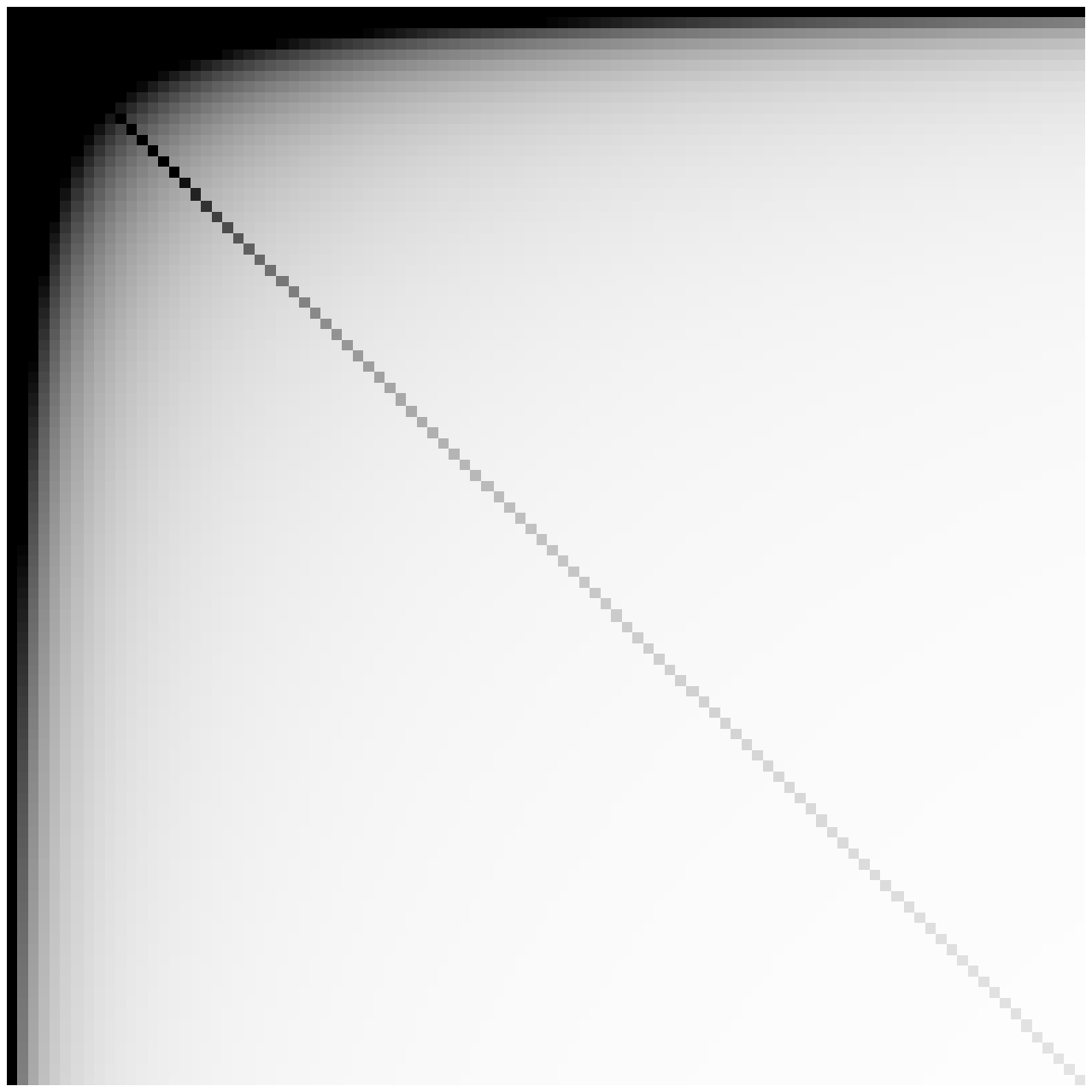}
  \label{fig:sample_cov_mat}
  \caption{The covariance matrix of the mixture of contextual word vectors with the assumption that the features are distributed over words with Zipfian distribution. The intensity of black shows the absolute value of the covariances.}
\end{figure}

Now we turn our attention on the covariance of the mixture of contextual word vectors $\mathbf{\Sigma}^{\boldsymbol{\mathcal{v}}}$. 
Denoting $\mathbf{\Sigma}^{\boldsymbol{\mathcal{v}}^{(k)}}$ as the covariance matrix of the $k$th contextual word vector forming the mixture model in \EQ{eq:mixture_model}, 
the covariance matrix of the mixture of contextual word vectors is $\mathbf{\Sigma}^{\boldsymbol{\mathcal{v}}}=\sum_{k=1}^{n}p(v_k)\mathbf{\Sigma}^{\boldsymbol{\mathcal{v}}^{(k)}}$, where $p(v_k)$ is the marginal probability of observing the $k$th word in the vocabulary set.
Given that the elements of the contextual word vectors follow the binomial distribution in \EQ{eq:binomial_dist}, for the diagonal elements of $\mathbf{\Sigma}^{\boldsymbol{\mathcal{v}}^{(k)}}$ ($k=1,\dots,n$), we have $\mathbf{\Sigma}^{\boldsymbol{\mathcal{v}}^{(k)}}_{(i,i)}=n(v_k)p(f_i|v_k)(1-p(f_i|v_k))$.  
Using Zipf's law for the words, the diagonal element $(i,i)$ of the overall covariance matrix will be as below:
\begin{equation}
  \begin{split}
    \boldsymbol{\Sigma}^{\boldsymbol{\mathcal{v}}}_{(i,i)} 
      & = \sum_{k=1}^{n}p(v_k)\mathbf{\Sigma}^{\boldsymbol{\mathcal{v}}^{(k)}}_{(i,i)} \\
      & = \sum_{k=1}^{n}n(v_k)p(v_k)p(f_i|v_k)(1-p(f_i|v_k)) \\
      & \approx TC^2\sum_{k=1}^{n}\frac{p(f_i|v_k)}{k^2}(1-p(f_i|v_k))\\
  \end{split}
  \label{eq:diag_cov_of_word_vectors_zipf}
\end{equation}
where we use $p(v_k)\approx\frac{C}{k}$, and $n(v_k)\approx\frac{TC}{k}$ and $T$ is the corpus size.
For the off-diagonal elements of $\mathbf{\Sigma}^{\boldsymbol{\mathcal{v}}^{(k)}}$ we have $\mathbf{\Sigma}^{\boldsymbol{\mathcal{v}}^{(k)}}_{(i,j)}=-n(v_k)p(f_i|v_k)p(f_j|v_k)$ with $i \ne j$.
Using Zipf's law, the off-diagonal elements can be approximated as below:
\begin{equation}
  \begin{split}
    \boldsymbol{\Sigma}^{\boldsymbol{\mathcal{v}}}_{(i,j)} 
      & = \sum_{k=1}^{n}p(v_k)\mathbf{\Sigma}^{\boldsymbol{\mathcal{v}}^{(k)}}_{(i,j)} \\
      & = \sum_{k=1}^{n}-n(v_k)p(v_k)p(f_i|v_k)p(f_j|v_k) \\
      & \approx -TC^2\sum_{k=1}^{n}\frac{p(f_i|v_k)p(f_j|v_k)}{k^2}\\
  \end{split}
  \label{eq:off_diag_cov_of_word_vectors_zipf}
\end{equation}
Similar to the analysis of the mean vector of the mixture of contextual word vectors, we analyse the covariance matrix with two scenarios about the distribution of features over the words.
First, we assume that the distribution of features over words follows the Zipfian distribution, \ie $p(f_i|v_k)\approx\frac{I}{i}$ where $I=\sum_{i=1}^{m}i^{-1}$.
Second, we use the maximum entropy principle and assume that the features are uniformly distributed over words, \ie $p(f_i|v_k)\approx\frac{1}{m}$.
With the first scenario, the diagonal element $(i,i)$ is approximated as
\begin{equation}
  \begin{split}
    \boldsymbol{\Sigma}^{\boldsymbol{\mathcal{v}}}_{(i,i)} 
      & \approx \frac{TC^2I(i-I)}{i^2}\sum_{k=1}^{n}\frac{1}{k^2}\\
      & \approx \frac{T\pi^2}{6i\log{m}(\log{n})^2}\Big( 1 - \frac{1}{i\log{m}} \Big) \\
  \end{split}
  \label{eq:diag_cov_of_word_vectors_zipf_zipf}
\end{equation}
and the off-diagonal element $(i,j)$ with $(i\neq j)$ is approximated as
\begin{equation}
  \begin{split}
    \boldsymbol{\Sigma}^{\boldsymbol{\mathcal{v}}}_{(i,j)} 
      & \approx -\frac{I^2C^2T}{ij}\sum_{k=1}^{n}\frac{1}{k^2}\\
      & \approx -\frac{T\pi^2}{6ij(\log{m})^2(\log{n})^2}
  \end{split}
  \label{eq:off_diag_cov_of_word_vectors_zipf_zipf}
\end{equation}
where we use $I\approx\frac{1}{\log{m}}$, $C\approx\frac{1}{\log{n}}$, and $\sum_{k=1}^{n}\frac{1}{k^2}\approx\frac{\pi^2}{6}$. 
\EQ{eq:diag_cov_of_word_vectors_zipf_zipf} and \EQ{eq:off_diag_cov_of_word_vectors_zipf_zipf} can be simplified into 
\begin{equation}
  \boldsymbol{\Sigma}^{\boldsymbol{\mathcal{v}}}_{(i,j)} = \frac{-A(1-\alpha)}{ij}
  \label{eq:cov_of_word_vectors_zipf_zipf}
\end{equation}
where $A=\frac{T\pi^2}{6(\log{m}\log{n})^2}$ and $\alpha$ is defined as below:
\begin{equation}
  \alpha = 
  \begin{cases}
    i\log{m}  & i=j \\
    0         & i\ne j\\
  \end{cases}
  \label{eq:cov_of_word_vectors_zipf_zipf_alpha}
\end{equation}
\EQ{eq:cov_of_word_vectors_zipf_zipf} shows that the covariance between the feature variables is directly proportional to the parameter $A$ and inversely proportional to the indices $i$ and $j$. 
Given that $T$, $m$, and $n$ are positive integers, and assuming that $m>1$ and $n>1$, $A$ is always a positive real number and $\alpha$ is a real number larger than for $i=j$.
Thus, the diagonal elements are positive since $\alpha>1$ and $A>0$, and the off-diagonal elements of the covariance matrix are negative since $A>0$.
The values of $A$ are directly proportional to the corpus size $T$ and inversely proportional to the feature size $m$ and vocabulary size $n$.
As shown in \FIG{fig:tau_vs_n}, the ratio $\tau=\frac{T}{(\log{n})^2}$ increases with the vocabulary size $n$, which itself is a function of the corpus size $T$. 
Given that $A=\frac{\tau\pi^2}{6(\log{m})^2}$, with a constant value of $m$, $A$ also linearly increases with the corpus size and vocabulary size. 
With a constant value of $\tau$, the variation of the values of $A$ is inversely proportional to the logarithm of the size of the feature set, $m$. 
A small feature set results in relatively larger values of $A$ and a very large feature set can result in significantly smaller values of $A$. 
In addition to the parameter $A$, the covariance between feature variables are also affected by their indices in the list of features sorted in descending order of the feature frequencies.
The \emph{absolute} value of covariance between the feature variable associated with less frequent features, for which the feature indices are high, is much smaller than the \emph{absolute} value of covariance between the feature variables associated with the features with relatively higher frequency.
The diagonal elements of the covariance matrix are higher than the absolute value of the off-diagonal elements in the same row and column. 
\FIG{fig:sample_cov_mat} provides a visualization of what we described about the elements of the covariance matrix in \EQ{eq:cov_of_word_vectors_zipf_zipf}. 
The fact that the diagonal elements of the covariance matrix sharply decreases as the index $i$ increases show that the majority of the data spread is due to the disproportionate contribution of the most frequent features with high indices.
In terms of the eigen-spectrum, this picture of the distribution of contextual word vectors implies a sharp decrease in the spectrum of eigenvalues of their covariance matrix. 
In other words, most of the variation of the data is along a few number of top eigenvectors of the covariance matrix. 
Later on, in \SEC{subsec:num_dim}, we will make use of this property to determine the optimal number of dimensions for the principal word vectors. 

Alternatively, if we assume that the features are uniformly distributed over words \ie $p(f_i|v_k)\approx\frac{1}{m}$, 
the element $(i,j)$ of the covariance matrix then will be as below:
\begin{equation}
  \boldsymbol{\Sigma}^{\boldsymbol{\mathcal{v}}}_{(i,j)} = \frac{T\pi^2\alpha}{6(m\log{n})^2}
  \label{eq:cov_of_word_vectors_unif_zipf}
\end{equation}
where $\alpha$ is defined as:
\begin{equation}
  \alpha = 
  \begin{cases}
    m-1  & i=j \\
    -1         & i\ne j\\
  \end{cases}
  \label{eq:cov_of_word_vectors_unif_zipf_alpha}
\end{equation}
\EQ{eq:cov_of_word_vectors_unif_zipf} and \EQ{eq:cov_of_word_vectors_unif_zipf_alpha} show that the elements of the covariance matrix become smaller as the number of features increases. 
The diagonal elements of the covariance matrix are larger than the absolute values of the off-diagonal elements. 
If we use $m\approx m-1$ in \EQ{eq:cov_of_word_vectors_unif_zipf_alpha}, then the diagonal elements of the covariance matrix will be as small as the corresponding mean values in \EQ{eq:mean_cwvec_uniform_dist}.
As shown in \FIG{fig:elemt_of_mean_of_cwvec}, the elements of the mean vector can be very small when the number of features increases. 
Similarly, the elements of the covariance matrix can be very small as the number of features increases. 
This together with the fact that the diagonal elements of the covariance matrix are higher than the off-diagonal values shows that depending on the size of the feature set the elements of the covariance matrix can be very small.
This basically means that the contextual word vectors are tightly massed around their mean vector which itself is close to the null vector. 

We summarize this section as follows.
A contextual word vector associated with a word is a vector of random variables, associated with a set of features, that count the frequency of seeing the word with each feature in a corpus. 
The set of contextual word vectors associated with all words in a vocabulary set form a mixture model. 
For a given corpus of a certain size, both the mean vector and the covariance matrix of the mixture of contextual word vectors are highly affected by the number of features. 
Depending on the size of the feature set, the mean vector of the mixture of contextual word vectors can be very close to the null vector and the contextual word vectors be massed around the mean vector. 
The spectrum of the eigenvalues of the covariance matrix of the principal word vectors sharply decreases and most of the variation in the data is along a few eigenvectors associated with the top eigenvalues. 

\section{Principal Component Analysis}
\label{sec:pca}
Principal component analysis (PCA) is a method to study the structure of data matrix. 
PCA is studied from different views which depends on the way that the data is seen \cite{jolliffe2002principal}. 
In a geometric view, the data matrix is seen as vectors in an Euclidean space spanned by the column or row vectors in the matrix. 
In a statistical view, the data matrix is seen as a sample from a multivariate distribution.  
In this paper, we focus on the statistical view of PCA since it is more relevant to the way that we model the contextual word vectors in \SEC{sec:contextual_word_vector}. 

PCA, in its statistical view, deals with the study of the structure of the covariance between a vector of random variables $\mathbf{X}=(\mathbf{x}_1, \dots, \mathbf{x}_m)^T$. 
It looks for a vector of independent latent variables $\mathbf{Y}=(\mathbf{y}_1, \dots, \mathbf{y}_k)$ ($k \ll m$), inferred from the original variables, whose variances under projection of original variables are maximal. 
The latent variables, called \emph{principal components}, are linear functions of original variables, $\mathbf{Y}=A^T (\mathbf{X} - \mathbf{E[\mathbf{X}]})$, where $\mathbf{E[\mathbf{X}]}$ is the expected vector of the random variables, and the $m \times k$ matrix $A=[A_1 \dots A_k]$ is composed of the $k$ dominant eigenvectors of the covariance matrix $\boldsymbol{\Sigma}^{\mathbf{X}}=\mathbf{E}[(\mathbf{X}-\mathbf{E}[\mathbf{X}])(\mathbf{X}-\mathbf{E}[\mathbf{X}])^T]$, \ie $\boldsymbol{\Sigma}^{\mathbf{X}}A_j = \lambda_j A_j$.
The resulting random variables in $\mathbf{Y}$ are independent from each other with the diagonal covariance matrix $\boldsymbol{\Sigma}^{\mathbf{Y}}=\mathbf{E}[(\mathbf{Y}-\mathbf{E}[\mathbf{Y}])(\mathbf{Y}-\mathbf{E}[\mathbf{Y}])^T]$ whose diagonal entries are as $\boldsymbol{\Sigma}^{\mathbf{Y}}_{(j,j)} = \lambda_j$ ($j=1,\dots,k$), where $\lambda_j$ is the eigenvalue associated with $A_j$.

Given the $m \times n$ sample matrix $X$ drawn from the random vector $\mathbf{X}=(\mathbf{x}_1, \dots, \mathbf{x}_m)^T$, the matrix of eigenvectors $A$ can be efficiently computed by singular value decomposition of the \emph{mean-centred} sample matrix $\mathscr{X} = X - \overline{X}\mathbf{1}_n^T$, where $\overline{X} = \frac{1}{n}X\mathbf{1}_n$ is the sample mean vector, and $\mathbf{1}_n$ is the vector of $n$ ones.
The left singular vectors of $\mathscr{X}$ are the eigenvectors of the covariance matrix $\boldsymbol{\Sigma}^\mathbf{X}$, and the singular values are equal to $\sqrt{(n-1)\lambda}$, where $\lambda$ is the vector of eigenvalues of $\boldsymbol{\Sigma}^\mathbf{X}$.  
Given the singular value decomposition of $\mathscr{X} = U\Sigma V^T$, the $k$ dimensional principal components of $\mathscr{X}$ can be efficiently computed by $Y=U_k^T\mathscr{X}$, or equivalently by $Y=\Sigma_k V_k^T$, 
where the $m \times k$ matrix $U_k$ and the $n \times k$ matrix $V_k$ are the left and right singular vectors corresponding to the $k$ dominant singular values of $\mathscr{X}$ on the main diagonal of the $k \times k$ diagonal matrix $\Sigma_k$. 

PCA can be generalized in various ways \cite{jolliffe2002principal,gpca2016}.
One way of generalization is to add an $m \times m$ metric matrix $\Phi$, and an $n \times n$ weight matrix $\Omega$ to the classical definition of PCA. 
These matrices provide for using a priori knowledge in the PCA. 
The metric matrix $\Phi$ is used for weighting the random variables and the weight matrix $\Omega$, which is usually a diagonal matrix, is used to weight the observations. 
It is also quite common to transform the data before performing the singular value decomposition. 
This adds a degree of non-linearity to the classical PCA and is helpful when the non-linear relationship between the variables is of interest.  
Another way of generalization is to give the method enough flexibility to form the final low dimensional vectors $Y$ not only on the basis of the top eigenvalues but on the basis of an \emph{arbitrary} set of eigenvalues and their corresponding eigenvectors. 
This can be done through an $m\times m$ diagonal matrix which is used to weight the eigenvalues in $\Sigma$. 
We refer to this eigenvalue weighting matrix as $\Lambda$.
The diagonal elements of $\Lambda$ are non-negative real numbers that control the variance of data along the corresponding eigenvectors. 
The presence of zero values on the diagonal elements is basically equivalent with eliminating the dimension formed by the corresponding eigenvector. 
So, if the desired number of dimensions is to be $k \ll m$, then we must have exactly $k$ non-zero values on the diagonal elements of $\Lambda$.

\ALG{alg:gpca} is a generalized version of PCA, called GPCA, with respect to the parameters mentioned above. 
The inputs to the algorithm are:
\begin{itemize}
  \item the $m \times n$ matrix $X$ sampled from the vector of random variables,
  \item the $m \times m$ metric matrix $\Phi$,
  \item the $n \times n$ weight matrix $\Omega$,
  \item the $m \times m$ diagonal eigenvalue weighting matrix $\Lambda$, and
  \item the transformation function $\mathcal{f}$
\end{itemize}
In Line~\ref{alg:gpca:transformation}, the algorithm applies several transformations on the sample matrix $X$ using the weight matrices and the transformation function $\mathcal{f}$. 
In Line~\ref{alg:gpca:centring}, the column vectors of the transformed matrix $\mathcal{X}$ are centred around their mean. 
Then in Line~\ref{alg:gpca:svd}, the singular value decomposition of the mean centred data matrix is computed. 
The singular values from which the low dimensional vectors are computer are weighted and selected in Line~\ref{alg:gpca:eigen_weight} via the weights provided by the input matrix $\Lambda$.
Usually, in practice, the matrix $\Lambda$ is not feed to the algorithm as an input argument, but rather it is computed in the algorithm as a function of the matrix $\Sigma$. 
In this case, the number of dimensions should be specified in the input argument list. 
The final low dimensional vectors are computed in Line~\ref{alg:gpca:y}.
The number of dimensions of these vectors are equal to the number of positive elements in the diagonal matrix $\Lambda$. 
A drawback of using \ALG{alg:gpca} on a large sparse matrix, such as the sample matrix of contextual word vectors, is that the algorithms cannot take advantage of the sparsity in the input matrix. 
This is because the matrix $\mathscr{X}$ in Line \ref{alg:gpca:centring} will inevitably be a dense matrix after being centred and it needs a large amount of memory and CPU time to be processed by the subsequent SVD step in Line~\ref{alg:gpca:svd}. 
In \SEC{sec:svd}, we introduce an efficient algorithm to approximate the singular value decomposition $\mathscr{X}=U\Sigma V^T$ from the non-centred matrix $\mathcal{X}$ without explicitly forming the dense matrix $\mathscr{X}$ in Line \ref{alg:gpca:centring}. 
This algorithm enables us to take the advantage of the sparsity in the input matrix to estimate its principal components. 
\begin{algorithm}
  \caption{Generalized principal component analysis.}\label{alg:gpca}
  \begin{algorithmic}[1]
    \Procedure{GPCA}{$X,\Phi,\Omega,\Lambda,\mathcal{f}$} 
      \State Form $\mathcal{X} \gets \mathcal{f}(\Phi X \Omega)$ \label{alg:gpca:transformation}
      \State Compute the mean vector $E$ and form the centred data $\mathscr{X} \gets \mathcal{X} - E\mathbf{1}_n^T$ \label{alg:gpca:centring}
      \State Compute singular value decomposition $\mathscr{X}=U\Sigma V^T$ \label{alg:gpca:svd}
      \State Weight the singular values with $\Lambda$, $\Sigma_1 \gets \Lambda \Sigma$ \label{alg:gpca:eigen_weight}
      \State Form the low dimensional principal components $Y \gets \Sigma_1 V^T$ \label{alg:gpca:y}
      \State \Return $Y$
    \EndProcedure
  \end{algorithmic}
\end{algorithm}

\subsection{Centred Singular Value Decomposition}
\label{sec:svd}
Any real-valued $m\times n$ matrix $X$ can be decomposed into three matrices as $X = U\Sigma V^T$
where the $m \times m$ matrix $U$ and $n \times n$ matrix $V$ are real-valued orthonormal matrices (\ie $U^TU=I_{m}$ and $V^TV=I_{n}$), and $\Sigma$ is a diagonal $m \times n$ matrix whose diagonal elements are non-negative real values. 

As mentioned before, the set of principal components of a mean centred matrix can be efficiently computed from the singular values and singular vectors of the matrix. 
In this section, we propose a method of singular value decomposition for this aim. 
Denoting $X$ as an $m \times n$ matrix, $E$ as an $m$ dimensional vector, and $\mathbf{1}_n$ as an $n$ dimensional vector of ones,
\ALG{alg:rsvd}, inspired by the randomized matrix factorization method introduced by \namecite{Halko2011finding}, 
returns a rank-$k$ approximation of the singular value decomposition of the data matrix $\mathscr{X} = X - E\mathbf{1}_n^T$, whose columns are centred around the vector $E$. 
The algorithm works in three major steps.

First is to approximate a rank $K$ basis matrix $Q_1$ ($k<K\ll n$) that captures most of the information in the input matrix $X$. 
This is done through the first three lines in \ALG{alg:rsvd}. 
A basis matrix of a vector space contains a set of linearly independent vectors that span the vector space. 
In Line~\ref{alg:rsvd:omega} a standard random matrix in drawn which is used in Line~\ref{alg:rsvd:smapling} to form the sample matrix $X_1$. 
The columns of $X_1$ are independent random points in the range of $X$.
By way of illustration, the $i$th element of the $j$th sampled vector, $X_{1(i,j)}$, is the linear combination of the $i$th element of all vectors in $X$ and the random linear coefficients in the $j$th row of $\Omega$, \ie $X_{1(i,j)}=\sum_{k=1}^{n} X_{(i,k)}\Omega_{(k,j)}$.
Hence, $X_1$ can be used to estimate the basis matrix $Q_1$, as it is done in Line~ref{alg:rsvd:qr}. 
Line~\ref{alg:rsvd:qrupdate} uses the QR-update algorithm proposed by \namecite[p.~607]{Golub:MC} to update the QR factorization of $X_1=Q_1R_1$ with respect to the input vector $E$.
For a given QR factorization such as $Q_1R_1=X_1$ and two vectors $u$, and $v$, the QR-update algorithm compute the QR-factorization of $Q'R'=X_1+uv^T$ through updating the already available factors $Q_1$ and $R_1$. 
Replacing $u$ with $-E$ and $v$ with $\mathbf{1}_n$, the QR-update in Line~\ref{alg:rsvd:qrupdate} returns the basis matrix $Q$ that captures most of the information in the mean centred matrix $\mathscr{X} = X - E\mathbf{1}_n^T$, \ie $\mathscr{X} \approx QQ^T\mathscr{X}$.
Note that, we compute the basis matrix of the mean centred matrix $\mathscr{X}$ without explicitly building the matrix $\mathscr{X}$. 

Second is to project the matrix $\mathscr{X}$ to the space spanned by $Q$, \ie $Y = Q^T\mathscr{X}$.
This step in done in Line~\ref{alg:rsvd:projection} through the application of the distributive property of multiplication over addition, \ie $Y = Q^TX_1 - Q^TE\mathbf{1}_n^T$. 
Note that in Line~\ref{alg:rsvd:projection}, instead of applying the product operator on $Q$ and $\mathscr{X}=X-E\mathbf{1}_n^T$, we use $Q^TX_1 - Q^TE\mathbf{1}_n^T$. 
This is because the amount of memory required by the latter approach is significantly smaller than the amount of memory required by the former approach. 
In the former approach, one needs to build the $m\times n$ matrix $\mathscr{X}=X-E\mathbf{1}_n^T$ but in the latter approach we have two $m\times K$ matrices $Q^TX_1$ and $Q^TE\mathbf{1}_n^T$. 
Since $K\ll m$ and in many applications the number of observations $n$ is larger than or equal (or at least comparable) to the number of variables $m$ ($m\le n$), the latter approach is more efficient. 
Moreover, since the matrix $Q^TE\mathbf{1}_n^T$ repeats the column vectors $Q^TE$ in all its columns, instead of building this matrix one can subtract the column vector $Q^TE$ from all columns of $Q^TX_1$ in a loop. 

Finally, in the third step, the SVD factors of $\mathscr{X}$ are estimated from the $K \times n$ matrix $Y$ in two steps.  
First, the rank-$k$ SVD approximation of $Y$ is computed using a standard method of singular value decomposition, \ie $Y=U_{1}SV^T$ (Line~\ref{alg:rsvd:svd}). 
Then, the left singular vectors are updated by $U \leftarrow QU_{1}$ resulting in $USV^T=QY$ (Line~\ref{alg:rsvd:qu}).  
Replacing $Y$ with $Q^T\mathscr{X}$ and the fact that $\mathscr{X} \approx QQ^T\mathscr{X}$, we have $USV^T\approx\mathscr{X}$, which is the rank-$k$ approximation of $\mathscr{X}$ .
\begin{algorithm}[H]
  \begin{algorithmic}[1]
    \Procedure{Centred-SVD}{$X,E,k,K$} 
      \State Draw an $n \times K$ standard Gaussian matrix $\Omega$ \label{alg:rsvd:omega}
      \State Form the sample matrix $X_1 \gets X\Omega$ \label{alg:rsvd:smapling} 
      \State Compute the QR factorization $X_1 = Q_1R_1$ \label{alg:rsvd:qr}
      \State Compute the QR factorization $QR = Q_1R_1 - E\mathbf{1}_n^T$ using the QR-update algorithm \cite{Golub:MC} \label{alg:rsvd:qrupdate}
      \State Form $Y \gets Q^TX - Q^TE\mathbf{1}_n^T$   \label{alg:rsvd:projection}
      \State Compute the singular value decomposition of $Y = U_{1}SV^T$ \label{alg:rsvd:svd}
      \State $U \gets QU_{1}$         \label{alg:rsvd:qu}
      \State \Return $(U, \Sigma, V)$
    \EndProcedure
  \end{algorithmic}
  \caption{Singular value decomposition of $X-E\mathbf{1}_n^T=U\Sigma V^T$.}
\label{alg:rsvd}
\end{algorithm}

\section{Principal Word Vectors}
\label{sec:princ_word_vec}

We define \emph{principal word vectors} as the set principal components of a sample matrix of contextual word vectors.
We refer to this matrix as \emph{contextual matrix}. 
Formally, for a given corpus formed by $n$ word variables, and $m$ feature variables, the columns of the $m \times n$ contextual matrix $M$ is the contextual word vectors associated with the word variables. 
Denoting $\boldsymbol{\mathcal{v}}_i$ ($i=1,\dots,n$) as a contextual word vector associated with word $v_i\in V$, the contextual matrix  
$M=\left({\boldsymbol{\mathcal{v}}_{1}} \dots {\boldsymbol{\mathcal{v}}_{n}}\right)$. 

\ALG{alg:pwvec} summarizes three main steps to build a set of $k$ dimensional principal word vectors form the corpus $E$, characterized by a set of feature variables and a set of word variables. 
First is to builds a contextual matrix through scanning the input corpus and counting the frequency of seeing words with different features in their contexts (lines \ref{alg:pwvec:init} to \ref{alg:pwvec:endfor}). 
The two search steps in Line~\ref{alg:pwvec:feature_hash} and Line~\ref{alg:pwvec:vocab_hash} can be rapidly done by the hashing mechanism proposed by \namecite{Zobel2001Hash}.  
Hence, the time complexity of this step is mostly influenced by the corpus size $T$. 
However the memory space complexity of this step is influenced by size of the feature set, size of the vocabulary set, and the distribution of features over words. 
In practice, the sparsity of the contextual matrix $M$, which defined as the ratio of the number of zeros to the total number of elements, is large enough to consider it as a sparse matrix. 
In \SEC{subsec:feature_variables}, we study different types of feature variables used in the literature and we show how the combination of different typed of feature variables can affect the sparsity and distribution of the data in the contextual matrix.

Second is to set the parameters of the GPCA, the weigh matrices $\Phi$ and $\Omega$, and the transformation function $\mathcal{f}$.
These parameters can be set with a priori knowledge about the distribution of features and words, or they can be set on the basis of the statistics provided by the contextual matrix.
In this research, we don't deal with the former approach but we study the latter approach in \SEC{subsubsection:GPCA_params}. 
Third is to run the GPCA algorithm in \ALG{alg:gpca} on the contextual matrix using the above parameters. 
\begin{algorithm}
  \caption{The algorithm to generate a set of principal word vectors for words in a corpus. $E$ is a corpus characterized by 
    the set of feature variables $\mathbf{F}=\{\mathbf{f}_1,\dots,\mathbf{f}_m\}$, and the set of word variables $\mathbf{V}=\{\mathbf{v}_1,\dots,\mathbf{v}_n\}$.
    }
  \label{alg:pwvec}
  \begin{algorithmic}[1]
    \Procedure{Principal-Word-Vector}{$E,k$} 
    \State Initialize the $m\times n$ contextual matrix $M$ with zeros  \label{alg:pwvec:init}
    \For{ $t\gets1$ {\bf to} $T$}
      \State Find the feature index $i$ for which $\mathbf{f}_i(e_t)=1$ \label{alg:pwvec:feature_hash}
      \State Find the word index $j$ for which $\mathbf{v}_j(e_t)=1$ \label{alg:pwvec:vocab_hash}
      \State $M_{(i,j)} \gets M_{(i,j)} + 1$
      \EndFor     \label{alg:pwvec:endfor}
    \State Build the weight matrices $\Phi$ and $\Omega$
    \State Build the diagonal eigenvalue weighting matrix $\Lambda$ having $k$ positive elements
    \State Train the transformation function $\mathcal{f}$
    \State $Y \gets \Call{GPCA}{M,\Phi,\Omega,\Lambda,\mathcal{f}}$
    \State \Return $Y$
    \EndProcedure
  \end{algorithmic}
\end{algorithm}

%

\subsection{Feature Word Variables}
\label{subsec:feature_variables}
A set of feature variables are defined by
\begin{enumerate*}[label={\arabic*)}]
  \item a feature set, and 
  \item a context function. 
\end{enumerate*}
The features are basically the word features which describe the words occurrences in a corpus. 
Some examples of features are word's part-of-speech tags, word's lemmas, word's forms, or even the label of the document to which a word belongs. 
As elaborated in \SEC{sec:contextual_word_vector}, a context function is a mapping between the elements of a corpus, each elements is mapped onto a subset of elements in the corpus. 

A singleton context function provides a  mapping between the individual elements of the corpus. 
The two commonly used singleton context functions (or contexts) in the literature are:
\begin{enumerate*}[label={\arabic*)}]
  \item the neighbourhood context, and  
  \item the dependency context.  
\end{enumerate*}
Denoting $E=\{e_1,\dots,e_T\}$ as a corpus of size $T$, the neighbourhood context of word $e_t\in E$ with parameter $\tau\in\mathbb{Z}$ ($0 < t + \tau < T$) is $\mathcal{C}_n(e_t;\tau)=e_{t+\tau}$. 
Depending on the sign of parameter $\tau$, the neighbourhood context returns the word at $\tau$th position before or after $e_t$. 
We will refer to the neighbour context with $\tau<0$ as backward neighbourhood context and the neighbourhood context with $\tau>0$ as forward neighbourhood context. 
The dependency context is defined by the dependency relations between words which is modelled by a directed graph whose nodes are the words in $E\cup\{\mbox{root}\}$, where \emph{root} is a special node used as the root node of a dependency tree.
The dependency context of word $e\in E$ with parameter $\tau\in\{0\}\cup\mathbb{Z}^+$ is defined as below:
\begin{equation}
  \mathcal{C}_{d}(e;\tau) = 
  \begin{cases}
    \mathcal{C}_{d}(\mbox{Pa}(e);\tau-1) & \tau > 0 \\
    e & \tau = 0
  \end{cases}
  \label{eq:dependency_context}
\end{equation}
where function $\mbox{Pa}:E\to \{E, \mbox{root}\}$ returns the parent word of the input word $e\in E$ with respect to the dependency graph of $E$.

Depending on the feature set and context function in use, different sets of feature variables can be formed. 
For $k$ disjoint sets of feature variables $\{\mathbf{F}_1,\dots,\mathbf{F}_k\}$ with $\mathbf{F}_i=\{\mathbf{f}^{(i)}_1,\dots,\mathbf{f}^{(i)}_{m_i}\}$ ($i=1,\dots,k$), 
we propose to combine them together in three ways.
The first is to form a set of joint feature variables from the Cartesian product set of the original sets of feature variables \ie the combined set of feature variables is $\mathbf{F}=\mathbf{F}_1 \times \mathbf{F}_2 \times \dots \times \mathbf{F}_k$.
The size of the joint feature variables $\mathbf{F}=\{\mathbf{f}_1,\dots,\mathbf{f}_m\}$ is $m=\prod_{i=1}^{k}m_i$ and its $j$th element is $\mathbf{f}_j=\prod_{i=1}^{k} \mathbf{f}^{(i)}_{j_i}$,
where $j=1,\dots,m$, $j_i$  ($i=1,\dots,k$) is a feature index in $\mathbf{F}_i$ such that $j_1+\sum_{i=2}^{k} (j_i-1)\prod_{l=1}^{i-1}m_l=j$. 
In other words, $\mathbf{f}_j$ is equal to the product of the feature variables in the $j$th element of the Cartesian product $\prod_{i=1}^k\mathbf{F}_i$. 
This means that the joint feature variable $\mathbf{f}_j$ is a Bernoulli random variable and the contextual word vectors formed by them follows the multinomial distribution in \EQ{eq:multinomial_contextual_word_vector}. 
Since the size of joint feature variables in $\mathbf{F}$ is much higher than the size of the individual feature variables in each $\mathbf{F}_i$ ($i=1,\dots,k$), 
the contextual word vectors generated with the joint feature variable are expected to be more dense around their mean than the contextual word vectors generated with $\mathbf{F}_i$ ($i=1,\dots,k$). 
This due to the effect of size of the feature variables on the covariance matrix of the contextual word vectors elaborated in \SEC{sec:contextual_word_vector}. 

As an example, let's assume that the feature variable $\mathbf{F}$ is formed by the Cartesian product of two sets of feature variables $\mathbf{F}_1$ and $\mathbf{F}_2$ which are defined as follow.
$\mathbf{F}_1$ is defined by a set of part-of-speech tags as its feature set and the neighbourhood context with parameter $\tau=-1$ as its context function. 
$\mathbf{F}_2$ is defined by a set of supertags tags as its feature set and the dependency context with parameter $\tau=-1$ as its context function. 
Then, the joint feature variable $\mathbf{f}_l=(\mathbf{f}^{(1)}_i,\mathbf{f}^{(2)}_j)$, where $\mathbf{f}^{(1)}_i\in\mathbf{F}_1$ and $\mathbf{f}^{(2)}_j\in\mathbf{F}_2$, for a given token $e_t$ is $1$ if $\mathbf{f}^{(i)}_1(e)=1$ and $\mathbf{f}^{(j)}_2(e)=1$
, \ie the part-of-speech tag of the immediate preceding token, $e_{t-1}$, matches the part-of-speech tag associated with $\mathbf{f}^{(i)}_1$ \emph{and} the supertag of the parent word matches the supertag associated with $\mathbf{f}^{(j)}_2$.
The joint approach of feature combination can be seen as an extension of the $n$-gram models which are mostly used in language modelling. 
The difference is that an $n$-gram model is defined as a \emph{contiguous} sequence of items, but in the joint approach, the variables generated by the joint approach are not necessarily a contiguous sequence.
This is because of the different context functions used in the feature variables constituting the set of joint feature variables.
By way of illustration, in the previous example, the part-of-speech tag of the immediate preceding word and the supertag of the parent word are taken into consideration which are not necessarily in a contiguous sequence defined by the words. 
In addition to this, the joint feature variables provide us with a systematic way to combine different types of contextual features together. 
The elements of the contextual word vectors built with the joint feature variables basically show the frequency of seeing each word with all combinations of the features in the original sets of feature variables. 
Since not all combinations of feature variables are likely to occur in the corpus, 
one can eliminate those joint variables which are never seen in the corpus, \ie $\sum_{t=1}^T\mathbf{f}_j(e_t)=0$ where $\{e_1,\dots,e_T\}$ is the training corpus.


Using the same notation as before, the second way to combine multiple sets of feature variables is to form the union set of feature variables:
\begin{equation}
  \mathbf{F} = \bigcup_{i=1}^{k} \mathbf{F}_{i}
  \label{eq:union_context}
\end{equation}
The size of the union set of features variables $\mathbf{F}=\{\mathbf{f}_1,\dots,\mathbf{f}_m\}$ is $m=\sum_{i=1}^{k} m_i$ where $m_i$ is the size of the feature variable set $\mathbf{F}_{i}$. 
The contextual word vectors obtained from this approach are equivalent to the concatenation of the contextual word vectors obtained from each set the feature variables. 
Denoting the set of contextual word vectors obtained from each set of the feature variable $\mathbf{F}_{i}\in\mathbf{F}$ ($i=1,\dots,k$) as $\{\boldsymbol{\mathcal{v}}_1, \dots, \boldsymbol{\mathcal{v}}_k\}$, 
all associated with the same word, the corresponding union contextual word vector will be $\boldsymbol{\mathcal{v}} = (\boldsymbol{\mathcal{v}}_1^T, \dots, \boldsymbol{\mathcal{v}}_n^T)^T$.
A contextual word vector obtained by the union feature variables does not follow the multinomial distribution in \EQ{eq:multinomial_contextual_word_vector}. 
This is because the range of the feature lookup function and the context function of the union set of feature variables are changed to their corresponding power sets $\mathcal{P}(F)$ and $\mathcal{P}(E)$ respectively, 
\ie $\mathcal{F}:E\to\mathcal{P}(F)$ and $\mathcal{C}:E\to\mathcal{P}(E)$.
In other words, each $e_t\in E$ $(i=1,\dots,T)$ activates $n$ feature variables in the union contextual word vector $\boldsymbol{\mathcal{v}}$, each of which belongs to one of the contextual word vectors forming $\boldsymbol{\mathcal{v}}$. 
However the multinomial distribution in \EQ{eq:multinomial_contextual_word_vector} requires that each observation $e_t\in E$ activates only one feature variable. 
The contextual word vector $\boldsymbol{\mathcal{v}}$ obtained from the union of the feature variable sets $\mathbf{F}_{i}$ $(i=1,\dots,k)$ follows the joint distribution of the multinomial distributions associated with each $\mathbf{F}_i$ $(i=1,\dots,k)$.
Nevertheless, the individual feature variables in $\mathbf{F}$ follow the binomial distribution in \EQ{eq:binomial_dist}.
Thus, the mean vector and the covariance matrix of the mixture of union contextual word vectors will have the same properties as the mean vector and covariance matrix of mixture of contextual word vectors studied in \SEC{sec:contextual_word_vector}. 


The third way to combine multiple set of feature variables is to add the contextual word vectors obtained from each set together.
This addition approach of feature combination requires the size of the feature variables be equal with each other. 
This restriction is imposed by the algebraic definition of the vector addition stating that two or more vectors can be added is they have the same number of dimensions. 
Since the number of dimensions of a contextual word vector is equal to the size of its feature set, we have the above restriction on the addition approach of feature combination. 
In addition to this algebraic restriction, the addition approach of feature combination is restricted to the correspondences between the feature variables in the sets. 
In fact, the addition approach of feature combination makes sense if there is a one-to-one mapping between the elements of feature sets such that the corresponding feature variables are associated with the \emph{same feature} and the \emph{same type of contexts} (e.g. neighbourhood context) but different context parameters. 
For example, we can add the contextual word vectors built with the neighbourhood context with different parameter values $\tau\in\mathbb{Z}$ ($0 < t + \tau < T$).
However, it does not make sense to add a contextual word vectors built with the neighbourhood context to another contextual word vector built with the dependency context. 

It can be shown that the contextual word vectors obtained from the addition approach of feature combination follow a multinomial distribution if the coefficients $\alpha_\tau$ in \EQ{eq:window_based_context} are positive integers \ie 
$\alpha_\tau\in\mathbb{Z}^{+}$.
However, in general, the contextual word vectors obtained from the addition approach approximate the Gaussian distribution with $\boldsymbol{\mu}=\sum_{\tau=a}^{a+n-1}{\alpha}_{\tau}\boldsymbol{\mu}_{\tau}$ and $\boldsymbol{\Sigma}=\sum_{\tau=a}^{a+n-1}\alpha_{\tau}\boldsymbol{\Sigma}_{\tau}$, 
where $\boldsymbol{\mu}_{\tau}$ and $\boldsymbol{\Sigma}_{\tau}$ are the mean vectors and covariance matrices of the multinomial distributions corresponding to $\boldsymbol{\mathcal{v}}_{\tau}$ ($\tau=a,\dots,a+n-1$).


The addition approach of feature combination can be used to form the window-based context used in the literature. 
Denoting $\boldsymbol{\mathcal{v}}_{\tau}$ as a contextual word vector obtained from the neighbourhood context with parameter $\tau$, we define a window-based contextual word vector for word $v$ with three parameters $a$, $n$, $\boldsymbol{\alpha}$ as below: 
\begin{equation}
  \boldsymbol{\mathcal{v}} = \sum_{\tau=a}^{a+n-1} \alpha_\tau\boldsymbol{\mathcal{v}}_{\tau}
  \label{eq:window_based_context}
\end{equation}
where $a\in\mathbb{Z}$ indicates the beginning of the window, $n$ is the window length, and $\boldsymbol{\alpha}$ is a vector of weights $\alpha_\tau\in\mathbb{R}^+$ associated with each word in the window.
Depending on the values of $a$ and $n$, one can form asymmetric (backward, and forward) and symmetric window-based contextual word vectors corresponding to the asymmetric and symmetric window-based contexts. 
A backward window-based contextual word vector is formed with the backward window-based context of length $k\in\mathbb{N}$, $k$ preceding words, \ie $a=-k$, $n=k$, and $\alpha_\tau=\frac{1}{\abs{\tau}}$ ($i=1,\dots,k$) .
Similarly, A forward window-based contextual word vector is formed with the forward window-based context of length $k\in\mathbb{N}$, $k$ succeeding words, \ie $a=1$, $n=k$, and $\alpha_\tau=\frac{1}{\tau}$ ($i=1,\dots,k$). 
As for symmetric window-based contextual word vector, we set $a=-k$, $n=2k+1$, and $\alpha_{\abs{\tau}}=\frac{1}{\abs{\tau}}$ for $\tau\neq 0$ and $\alpha_0=0$.
The window-based context can be generalized to dependency context as well. 
We can form an ancestral context from addition of multiple contextual word vectors obtained from dependency context with different parameter values.

\subsection{The GPCA Parameters}
\label{subsubsection:GPCA_params}

In this section we study the GPCA parameters in the following order. 
First, we introduce different metric and weight matrices and study how the principal word vectors are influenced by these matrices. 
Then, we propose some transformation functions to expand the distribution of the mixture contextual word vectors described in \SEC{sec:contextual_word_vector}. 
Finally, we study different ways of weighting the eigenvalues which facilitates controlling the variance of the principal word vectors. 

\subsubsection{Metric and Weight Matrices}
\label{subsubsection:weight_matrices}
The metric matrix $\Phi$ and the weight matrix $\Omega$ provide GPCA with a priori knowledge about the feature variables and word variables respectively. 
The matrix $\Phi$ defines a metric on the feature variables in such a way that the distance between two contextual word vectors $\boldsymbol{\mathcal{v}}^{(i)}$ and $\boldsymbol{\mathcal{v}}^{(j)}$ is 
$(\boldsymbol{\mathcal{v}}^{(i)} - \boldsymbol{\mathcal{v}}^{(j)})^T\Phi(\boldsymbol{\mathcal{v}}^{(i)} - \boldsymbol{\mathcal{v}}^{(j)})$.
This is helpful to scale the values in the contextual word vectors.  
The matrix $\Omega$ is usually used to weight the observations and in our model it is used to weight the contextual word vectors. 
Although neither the metric matrix $\Phi$ nor the weight matrix $\Omega$ need to be diagonal, we restrict our research on some specific diagonal metric and weight matrices which are connected to the related work in the literature. 

\TABLE{table:metric_matrix} gives a list of diagonal metric and weight matrices that can be used with the contextual word vectors.
In the remaining parts of this section, we study how the contextual word vectors are affected by different combinations of $\Phi$ and $\Omega$ listed in \TABLE{table:metric_matrix}.
Denoting $X$ as the matrix of contextual word vectors, we define $Y=\Phi X \Omega$ as the product matrix which is formed in the line~\ref{alg:gpca:transformation} of \ALG{alg:gpca}.
We use pair $(\Phi,\Omega)$ to form different combinations of the metric and weight matrices in \TABLE{table:metric_matrix}.

Using $(\Phi,\Omega)=(\mathbf{I}_m,\mathbf{I}_n)$, the product matrix $Y=\mathbf{I}_m X \mathbf{I}_n$ will be equal to $X$.
The metric matrix \emph{iff} gives higher weights to the less frequent features. 
If we use (\emph{iff}, $\mathbf{I}_n$) then $Y_{(i,j)}\approx p(\mathbf{c}_i=1|\mathbf{v}_j)$, which is an estimation of the conditional probability of seeing the feature $c_i$ conditioned on $v_j$.
The weight matrix \emph{iwf} can be used to cancel out the Zipfian effect of the words distributions. 
\emph{iwf} reduces the disproportionate effect of very frequent words on the elements of contextual word vectors. 
If we use ($\mathbf{I}_m$, \emph{iwf}) then $Y_{(i,j)}\approx p(\mathbf{v}_j|\mathbf{c}_i)$, which is an estimation of the conditional probability of seeing the word $v_j$ conditioned on $c_i$.
If we use (\emph{iff},\emph{iwf}) then the elements of product matrix $Y$ will be as follows: $Y_{(i,j)}=\frac{n(c_i,v_j)}{n(c_i)n(v_j)}$ ($i=1,\dots,m$ and $j=1,\dots,n$).
Using the element wise transformation function $f(Y)=\max(0,\log(TY))$, where $T$ is the corpus size, 
the element ($i,j$) of $f(Y)$ will be equal to the pointwise mutual information between the contextual feature $c_i$ and the word $v_j$.
When we use (\emph{isf}, $\mathbf{I}_n$) the product matrix $Y$ will be equal to the correlation matrix of $X$.
So, \ALG{alg:gpca} performs PCA on the correlation matrix of contextual word vectors rather than their covariance matrix.
Using correlation matrix instead of covariance matrix is a solution to mitigate the heterogeneous metric problem in the observations, where the elements of observation vectors are represented in different metric systems (\eg kilogram, and meter). 
One can imagine similar problem in the elements of contextual word vectors, where we see imbalanced contribution of contextual features in the word vectors. 

\begin{table}[htbp]
  \renewcommand{\arraystretch}{2.2}
  \begin{tabularx}{\textwidth}{lll}
    \hline
    Name            & Definition &   \\ \hline
    $\mathbf{I}_m$  & $\Phi_{(i,i)} = 1$    & the $m \times m$ identity matrix \\
    \emph{iff}      & $\Phi_{(i,i)} = \dfrac{1}{n(c_i)}$      & The inverse of feature frequency \\ 
    \emph{isf}      & $\Phi_{(i,i)} = \dfrac{1}{\sigma_{\boldsymbol{\mathcal{v}}_i}}$   & The inverse of standard deviation of feature frequency \\ 
    $\mathbf{I}_n$  & $\Omega_{(i,i)} = 1$  & the $n\times n$ identity matrix \\
    \emph{iwf}      & $\Omega_{(i,i)} = \dfrac{1}{n(v_i)}$    & The inverse of word frequency \\ 
  \end{tabularx}
  \label{table:metric_matrix}
  \caption{The list of diagonal metric and weight matrices. $n(c_i)$ is the frequency of seeing the contextual features $c_i$ in the corpus $E=\{e_1,\dots,e_T\}$, \ie $n(c_i) = \sum_{t=1}^T \mathbf{c}_i(e_t)$ and $\mathbf{c}_i$ is the contextual variable corresponding to $c_i$.  $n(v_i)$ is the frequency of seeing word $v_i$ in the corpus $E=\{e_1,\dots,e_T\}$, \ie $n(v_i) = \sum_{t=1}^T \mathbf{v}_i(e_t)$ and $\mathbf{v}_i$ is the word variable corresponding to $v_i$.  $\sigma_{\boldsymbol{\mathcal{v}}_i}$ is the standard deviation of the $i$th random variable in the set of contextual word vectors, \ie $i$th row of the input matrix $X$ used in \ALG{alg:pwvec}.}
\end{table}

\subsubsection{Transformation Function}
\label{subsubsection:trans_func}
Data transformation is a common preprocessing step in the principal component analysis of special data. 
It is also seen as a way of non-linear principal component analysis of a set of random variables. 

As elaborated in \SEC{sec:contextual_word_vector}, the eigenvalues of the covariance matrix of the mixture of contextual word vectors decays sharply as the number of features increases. 
This depicts a long elliptical distribution for mixture of contextual word vectors which is highly stretched along a few top eigenvectors and highly tight along the remaining eigenvectors. 
The principal component analysis of the data with such a distribution is highly influenced by the data along the top eigenvectors and can not capture enough information about the data variation along the other eigenvectors. 
In order to mitigate this problem, we propose to compress the data along the top eigenvectors and expand them along the remaining eigenvectors while preserving the order of eigenvectors with respect to their eigenvalues. 
This transformation reshapes the data distribution from a long elliptical distribution to a more standard elliptical distribution which is more similar to the normal distribution with diagonal covariance matrix. 
This can be achieved through the application of any monotonically increasing concave functions that preserve the given order of the data and magnify small numbers in its domain.
Some examples of these transformation functions are the logarithm, the hyperbolic tangent, and the power transformation functions. 

The expansion effect of the transformation function increases the of entropy of the mixture of contextual word vectors. 
We use this property to tune the parameters of the transformation function. 
Denoting $\boldsymbol{\mathcal{v}}$ as the mixture of contextual word vectors and $\mathcal{f}(\boldsymbol{\mathcal{v}};\theta)$ as the transformation function defined with the parameter set $\theta$, the optimal parameter set $\hat{\theta}$ is one that maximizes the entropy of the data:
\begin{equation}
  \hat{\theta} = \arg\max_{\theta} H(\mathcal{f}(\boldsymbol{\mathcal{v}};\theta))
  \label{eq:obj_trans_func}
\end{equation}
The optimal value of $\hat{\theta}$ in \EQ{eq:obj_trans_func} can be estimated by iterative optimization techniques such as genetic algorithm or simulated annealing with the objective function $g(\theta)=H(\mathcal{f}(\boldsymbol{\mathcal{v}};\theta))$.
In order to compute the entropy of the mixture of contextual word vectors $\boldsymbol{\mathcal{v}}$ in \EQ{eq:obj_trans_func} we need to compute the probability density function $f(\mathcal{f}(\boldsymbol{\mathcal{v}};\theta)$, \ie $H(\mathcal{f}(\boldsymbol{\mathcal{v}};\theta)) = -\sum_{\boldsymbol{\mathcal{v}}} f(\mathcal{f}(\boldsymbol{\mathcal{v}};\theta)) \log{f(\mathcal{f}(\boldsymbol{\mathcal{v}};\theta))}$. 
We use kernel density estimate to approximate the probability density function $f$. 
In cases where the dimensionality of $\boldsymbol{\mathcal{v}}=(\boldsymbol{\mathcal{v}}_1,\dots,\boldsymbol{\mathcal{v}}_m)$ is too high, 
we partition $\boldsymbol{\mathcal{v}}$ into a limited number of vectors $\{\boldsymbol{b}_1,\dots,\boldsymbol{b}_k\}$ and estimate the entropy as the mean of the entropy obtained from each of these vectors. 
Each vector $\boldsymbol{b}_i$ of length $l$ is characterized by a set of random indices $s_i=\{i_1,\dots,i_l\}$, which is a subset of the permutation of $\{1,\dots,m\}$, \ie $s_i\subseteq\{1,\dots,m\}$ and  such the $\bigcup_{i=1}^k s_i = \{1,\dots,m\}$. 
The elements of $\boldsymbol{b}_i$ are the sequence of the elements of $\boldsymbol{\mathcal{v}}$ with the indices in $s_i$ , \ie $\boldsymbol{b}_i=(\boldsymbol{\mathcal{v}}_{i1},\dots,\boldsymbol{\mathcal{v}}_{il})$. 
Given the set of vectors, we estimate the entropy in \EQ{eq:obj_trans_func} as below:
\begin{equation}
  H(\mathcal{f}(\boldsymbol{\mathcal{v}};\theta)) \approx \frac{1}{k}\sum_{i=1}^{k} H(\mathcal{f}(\boldsymbol{b};\theta))
  \label{eq:ent_subvector}
\end{equation}
where $k$ is the number of partitions and $H(\mathcal{f}(\boldsymbol{b};\theta))=-\sum_{\boldsymbol{b}} f(\mathcal{f}(\boldsymbol{b};\theta)) \log{f(\mathcal{f}(\boldsymbol{b};\theta))}$. 
In this approach, the dependencies between the feature variables in each set $s_i$ $(i=1,\dots,l)$ are take into consideration but the dependencies between the feature variables in different sets are no taken into consideration. 
In other words, the set of feature variables $s_1,\dots,s_l$ are assumed to be independent of each other. 
Although this is not a correct assumption, we adopt this since it covers part of the dependencies between the features and enable us to estimate the entropies efficiently. 
In addition, the independence assumption about the feature variables enable us to process the vectors $\boldsymbol{b}_i$ $(i=1,\dots,l)$ in parallel. 
We leave the true computation of the entropies as future work. 

\subsection{Eigenvalue Weighting Matrix}
The eigenvalue weighting matrix $\Lambda$ is used in Line~\ref{alg:gpca:eigen_weight} in \ALG{alg:gpca} to control the variance of the principal components along the selected eigenvectors. 
It is also used to choose the desired number of dimensions of the principal components. 
A study of how to fill the elements of $\Lambda$ is provided by \namecite[Chapter~6.3]{jolliffe2002principal}.
In this section, we present two eigenvalue weighting matrices which are used in our experiments. 
First is the matrix used in the classic PCA. 
Assuming that the singular values on the diagonal elements of the matrix $\Sigma$ in \ALG{alg:gpca} are sorted in their descending order, the classic PCA algorithm defines the matrix $\Lambda$ as 
\begin{equation}
  \Lambda = \sqrt{n-1}
  \begin{bmatrix}
    \mathbf{I}_k  & \mathbf{0} \\
    \mathbf{0}    & \mathbf{0} \\
  \end{bmatrix}
  \label{eq:classic_Lambda}
\end{equation}
where $n$ is the number of observations (\ie number of words), $\mathbf{I}_k$ is the $k \times k$ identity matrix and $\mathbf{0}$ is the matrix of zeros. 
The constant coefficient $\sqrt{n-1}$ is to compute eigenvalues from singular values. 
This matrix chooses the top $k$ eigenvalues and their corresponding eigenvectors that account for most of the variation in the data matrix, \eg the contextual matrix. 
Second is the matrix inspired by the word embedding approach proposed by \namecite{basirat-nivre:2017:NoDaLiDa}. 
In this approach, the effect of eigenvalues on the data variance is completely ignored. 
Thus, the principal components have the same variance along all eigenvectors. 
This can be implemented through the eigenvalue weighting matrix whose elements are inversely proportional to the singular values in $\Sigma$:
\begin{equation}
  \Lambda = \alpha\sqrt{n-1}
  \begin{bmatrix}
    \Sigma_k^{-1}  & \mathbf{0} \\
    \mathbf{0}    & \mathbf{0} \\
  \end{bmatrix}
  \label{eq:normal_Lambda}
\end{equation}
where $\alpha$ is a constant, $n$ is the number of observations (\ie number of words), and $\Sigma_k^{-1}$ is inverse of the diagonal matrix of top $k$ eigenvalues. 
This matrix results in the same variance equal to $\alpha^2$ along the top $k$ eigenvectors. 
In other words the principal components are normally distributed with mean vector $\mathbf{0}$ and the covariance matrix $\alpha^2\mathbf{I}_k$. 
The fact that the elements of principal components are independent of each other and follow the normal distribution makes them suitable to be used in the neural networks \cite{lecun2012efficient}. 
The parameter $\alpha$ is recommended to be set as the standard deviation of the initial weights of the neural network. 
Replacing $\Lambda$ in Line~\ref{alg:gpca:eigen_weight} in \ALG{alg:gpca}, we have 
\begin{equation}
  \Sigma_{1} = \alpha\sqrt{n-1}
  \begin{bmatrix}
    \mathbf{I}_k  & \mathbf{0} \\
    \mathbf{0}    & \mathbf{0} \\
  \end{bmatrix}
  \label{eq:I_sigma1}
\end{equation}
Using this matrix in Line~\ref{alg:gpca:y} in \ALG{alg:gpca}, we have 
\begin{equation}
  Y = \alpha\sqrt{n-1} 
  \begin{bmatrix}
    V_k^T \\
    \mathbf{0} \\
  \end{bmatrix}
  \label{eq:y_normal}
\end{equation}
where $\alpha$ and $n$ are as above, and the $n\times k$ matrix $V_k$ is the first $k$ singular vectors in $V$.
Eliminating the useless matrix $\mathbf{0}$ in \EQ{eq:y_normal}, the principal word vectors are the top $k$ right singular vectors of the mean centred contextual matrix scaled by the coefficient $\alpha\sqrt{n-1}$.

\section{Experimental Setting}
\label{sec:exp_setting}

The most effective parameters on the principal word vectors are:
\begin{enumerate}
  \item feature variables,
  \item number of dimensions. 
  \item metric and weight matrices, 
  \item transformation function, and
\end{enumerate}
Each of these parameters can be defined in several ways (see \SEC{sec:princ_word_vec}), which result in a large parameter space to analyse exhaustively. 
We navigate through the parameters as follow.
First we define a default setting to initialize the parameters. 
Starting with the feature variables, we form different types of feature variables described in \SEC{subsec:feature_variables} and study how the principal word vectors are affected by feature variables.
Then, we study the effect of number of dimensions on the principal word vectors. 
The effect of the weighting matrices and the transformation function are studies together. 
In all experiments, we set all parameter with their default value except for those that are under study. 

The default setting of the parameters is defined as follows. 
The set of feature variables are initialized by word forms as their contextual feature and the backward neighbourhood context as their context function. 
The number of dimensions is set to $100$. 
The metric matrix and the weight matrix are both set with the identity matrix.
The eigenvalue weighting matrix $\Lambda$ is set as in \EQ{eq:classic_Lambda} with $k=100$. 
The transformation function is the identity function, \ie no transformation is done on the elements of contextual word vectors. 

The principal word vectors are studied and evaluated in two ways.
First is to study the spread of the principal word vectors. 
second is to study the discriminability of the word vectors. 
The spread of principal word vectors is measured through the determinant of the covariance matrix of the word vectors. 
The determinant of a covariance matrix, also known as the generalized variance, is equal to the product of the eigenvalues of the covariance matrix. 
Due to the disproportionate contribution of the top eigenvalue in the product, we use the logarithm of the generalized variance as the spread of principal word vectors.
Denoting $\lambda_i$ as the $i$th top eigenvalue of the covariance matrix of principal word vectors, the logarithm of generalized variance of principal word vectors is computed as below:
\begin{equation}
  \log{\mbox{GV}}=\sum_{i=1}^{k} \log{\lambda_i}
  \label{eq:log_gv}
\end{equation}

We use the Fisher discriminant ratio (FDR) to measure the discriminability of the word vectors.
The Fisher discriminant ratio measures the separation between a set of data points with regard to their categories. 
For a given number of data items together with their category labels, 
the Fisher discriminant rate is equal to the sum of the positive eigenvalues of the \emph{between class covariance matrix} of the data ($\Sigma_B$) multiplied by the inverse of the \emph{within class covariance matrix} of the data $\Sigma_W$, 
\ie $\mbox{FDR}=\sum \lambda_i$ where $\lambda_i>0$ is the $i$th top positive eigenvalue of $\Sigma_B\Sigma_W^{-1}$, which can be computed through the general eigenvalue equation $\Sigma_B A_i=\lambda_i \Sigma_W A_i$ with $A_i$ and $\lambda_i$ being the $i$th eigenvector and eigenvalues respectively. 
The within class covariance is computed as the sum of the covariance of data in each class weighted by the class probabilities, \ie $\Sigma_W = \sum_{i=1}^{c} p_i\Sigma_i$ where $p_i$ is the marginal probability of seeing the data belonging to the $i$th class, $\Sigma_i$ is the covariance of data belonging to the $i$th class and $c$ is the total number of classes. 
The between class covariance is computed as the covariance of the mean vectors of the classes, 
\ie $\Sigma_B=\mathbf{E}[(\boldsymbol{\mu}_i - \boldsymbol{\mu})(\boldsymbol{\mu}_i - \boldsymbol{\mu})^T]$ where $\boldsymbol{\mu}_i$ is the mean vector of the $i$th class and $\boldsymbol{\mu}$ is the overall mean of the data.

We use FDR to measure the syntactic and the semantic discriminability of the word vectors.
The syntactic discriminability of the word vectors are computed with respect to the syntactic categories of words, part-of-speech tags such as noun, verb, etc.
The semantic discriminability of words are computed with respect to the abstract named entities such persons, locations, organization, etc.
The set of universal part-of-speech tags \cite{nivre2016universal} is used to measure the syntactic discriminability of word vectors and 
the data provided by the CoNLL-2003 shared task \cite{TjongKimSang:2003:ICS:1119176.1119195} is used to measure the semantic discriminability of the word vectors. 
We use development set of the English part of the corpus of universal dependencies $v2.0$ and the \texttt{testa} file of the English part of the shared task. 
Since the syntactic and semantic categories of words vary with the contextual environment of words, in order to take the context of words in to consideration, 
we represent each word in the corpora with a large vector built with the concatenation of the word vectors associated with the word itself and its surroundings. 
More concretely, each word is represented by the concatenation of the word vectors associated with the words in a symmetric window of length $7$ where the word in interest is in the middle of the window, \ie $3$ preceding words, the labelled word at position $4$, and $3$ succeeding words. 

We also compare principal word vectors with word vectors obtained from popular methods of word embeddings. 
In addition to the spread and the discriminability of the word vectors, the comparisons are on the basis of the word similarity benchmarks developed by \namecite{faruqui-2014:SystemDemo}, and dependency parsing framework used by \namecite{basirat-nivre:2017:NoDaLiDa}. 
The word similarity benchmark evaluates a set of word vectors in $13$ different word similarity benchmarks.  
Each benchmark contains pairs of English word associated with their similarity rankings. 
The tool reports the correlation between the similarity rankings provided by the word similarity benchmark and the cosine similarity between the word vectors. 
In order to provide an overall view over the performance of word vectors, we report the average of the correlations obtained from all word similarity benchmarks. 
The dependency framework evaluates a set of word vectors with respect to their contributions on the accuracy of dependency parsing. 
\namecite{basirat-nivre:2017:NoDaLiDa} uses the Stanford dependency parser \cite{chen14} to evaluate different sets of word vectors. 
The parser is an arc-standard system \cite{nivre2004incrementality} with a feed-forward neural-network as its classifier. 
The parsing experiments are carried out on Wall Street Journal (WSJ) \cite{Marcus93_WSJ} annotated with Stanford typed dependencies (SD) \cite{de2010stanford}.
Sections $02$--$21$ of WSJ are used for training, and sections $22$ and $23$ are used as development set and test set respectively. 
The part-of-speech tags are assigned to words through ten-way jackknifing on the training sets using the Stanford part-of-speech tagger \cite{toutanova2003feature}.
We use the parser with $100$ dimensional word vectors and $400$ hidden units in the hidden layer of the neural network. 
The remaining parameters are set to their default values. 

The word vectors are extracted from the English corpus provided by the CoNLL-2017 shared task \cite{K17-3:2017} as additional data for training word embeddings. 
The corpus is annotated with universal part-of-speech tags and universal dependency relations using UDPipe \cite{udpipe:2017}. 
We use the annotations with no change but normalize the raw tokens as follow.
All sequences of digits are replaced with the special token \emph{<number>} and all tokens with frequency less that $50$ are replaced with \emph{<unknown>}. 
The normalized corpus contains $11,362,947,002$ tokens in total with $1,028,590$ unique tokens.

\section{Experiments}
\label{sec:experiments}

In this section, we study the results obtained from our experiments with principal word vectors. 
First, we study the effect of the feature variables, number of dimensions, and the GPCA parameters on the word vectors on the spread and discriminability of principal word vectors. 
Then, we compare principal word vectors with other sets of word vectors. 

\subsection{Feature Variables}
\label{sebsec:experiments_on_feature_variables}
In this section, we study the effect of feature variables on the spread and discriminability of the principal word vectors. 
As detailed in \SEC{sec:contextual_word_vector}, a set of feature variables is characterized by a set of contextual features and a context function. 
In \SEC{subsec:feature_variables}, we proposed different approaches to form new sets of features variables through combining different sets of feature variables. 
Our experiments on the feature variables are performed on the multiple combination of feature variables formed by the following contextual features and context functions:
\begin{itemize}
  \item contextual features: word forms, and part-of-speech tags.
  \item context function: the dependency context, and the neighbourhood context. 
\end{itemize}
The neighbourhood context is used with different values of $\tau=\mypm1,\dots,\mypm10$.
We combine the feature variables formed by the neighbourhood context in the following ways. 
First is to form different variants of the window-based context, including the backward, the forward, and the symmetric window-based context. 
Similarly, we form different variant of the union context, including the backward, the forward, and the symmetric union context.
We also form the joint set of feature variables from the feature variables formed by the word forms and the corresponding feature variables formed by the part-of-speech tags. 

We use the universal part-of-speech tags and the universal dependencies provided by our training corpus as the set of part-of-speech tags and the dependency context. 
All parameters of the principal word embedding model are set to their default values except for the feature variables and the number of dimensions. 
The feature variables are set as above and the number of dimensions is set to $15$. 
The reason we choose $15$ as the number of dimensions is the small size of the universal part-of-speech tag set, $17$. 
As mentioned in \SEC{sec:princ_word_vec}, the maximum number of dimensions generated by the word embedding model is always smaller than or equal to the size of the contextual feature set.
Hence, we set the number of dimensions $k$ to $15$ to cancel out make a fair comparison between the principal word vectors generated with the word forms and part-of-speech tags, although the principal word vectors generated with word forms can have thousands of dimensions. 
Later on, in \SEC{subsec:num_dim}, we study how the principal word vectors are affected by the number of dimensions. 

\subsubsection{Data Spread}
\FIG{fig:best_fisher_15dim} show the logarithm of generalized variance of the $15$-dimensional principal word vectors generated with different types of feature variables. 
As shown, regardless of the context in use, the principal word vectors generated with the part-of-speech tags (\FIG{subfig:pos_generalized_variance}) 
result in higher amount of the variance than the principal word vectors generated with word forms (\FIG{subfig:word_form_generalized_variance_15dim}), 
and the principal word vectors generated with the joint word forms and part-of-speech tags (\FIG{subfig:word_form_pos_generalized_variance}).
However, this is worth noting the results reported for the principal word vectors generated with the part-of-speech tags are due to the contribution of almost \emph{all} of the eigenvalues but the other results are due to the contribution of a small fraction of eigenvalues, \ie $15$ eigenvalues out of thousands of possible eigenvalues. 
The descending trend in the results obtained from the neighbourhood contexts shows that the word vectors become more dense as the parameter $\abs{\tau}$ increases. 
The principal word vectors generated with word forms and both backward and forward neighbourhood contexts result in the same values of generalized variance. 
This is because, for each value of $\tau>0$, the contextual matrix $M^{(-\tau)}$ obtained from the backward neighbourhood context is equal to the transpose of the contextual matrix $M^{(\tau)}$ obtained from the forward neighbourhood context \ie $M^{(-\tau)}={M^{(\tau)}}^T$.
In other words, for each pair of tokens $(e_t,e_{t+\tau})$ in the corpus with $\mathbf{v}_i(e_t)=1$ and $\mathbf{v}_j(e_{t+\tau})=1$, we have $\mathbf{f}_j(e_t)=1$ and $\mathbf{f}_i(e_{t+\tau})=1$, \ie $e_t=(v_i,f_j)$ and $e_{t+\tau}=(v_j,f_i)$, since the set of contextual features $F$ is equal to the set of vocabulary units $V$, \ie $f_i=v_i$, and $f_j=v_j$, hence $M^{(-\tau)}_{(i,j)}=M^{(\tau)}_{(j,i)}$.
Thus, the covariance matrix of the corresponding contextual matrices with the above property will be the same, hence resulting in the same sets of eigenvalues. 
The overlap between the results obtained from the forward and backward window-based context in \FIG{subfig:word_form_generalized_variance_15dim} is interpreted in the same way. 

The descending trend observed in the neighbourhood contexts changes to ascending once we combine the neighbour contexts together. 
The ascending trends observed in the window-based and union context show that the related feature combination approaches described in \SEC{subsec:feature_variables} are helpful to increase the spread of the principal word vector. 
\FIG{fig:best_fisher_15dim} shows that the highest amounts of data spread are obtained from the symmetric window-based and union context regardless of the type of contextual feature in use. 
The union contexts always result in higher amount of generalized variance than their corresponding window-based contexts. 

We summarize our observations on the effect of feature variables on the data spread as follows.
The spread of principal word vectors generated with the neighbourhood context decreases as the absolute value of parameter $\tau$ increases. 
The spread of the word vectors generated with the window-based and union feature combination approaches increases as the parameter $k$ increases. 
The union contexts result in higher mount of their corresponding window-based contexts. 
\begin{figure}[htp!]
  \begin{center}
    \subfloat[]{\scalebox{0.6} {
      \begin{tikzpicture}
  \begin{axis}[
      align =center,
      xlabel={x},
      xmin=0, xmax=11,
      ymin=230, ymax=355,
    legend cell align=left, legend style={ column sep=.5ex, font=\footnotesize },
    legend style={at={(0.00,1.02)},anchor=south west,legend columns=5},
    ]
    \addplot
    coordinates {
      (1, 298.049422) (2, 286.633196) (3, 269.153102) (4, 261.919251) (5, 249.947049) (6, 247.858300) (7, 241.938727) (8, 242.513225) (9, 236.731802) (10, 234.894504)
    } ;
    \addlegendentry{BN}

    \addplot
    coordinates {
      (1, 298.083728) (2, 286.526709) (3, 269.016483) (4, 261.907769) (5, 249.890969) (6, 247.798198) (7, 241.857690) (8, 242.427901) (9, 236.687325) (10, 234.849868)
    } ;
    \addlegendentry{FN}

    \addplot
    coordinates {
      (1, 298.050099) (2, 302.757234) (3, 304.724124) (4, 305.712394) (5, 306.434888) (6, 306.955533) (7, 307.352536) (8, 307.651435) (9, 307.895561) (10, 308.086447)

    } ;
    \addlegendentry{BW}

    \addplot
    coordinates {
      (1, 298.083520) (2, 302.660851) (3, 304.619696) (4, 305.611781) (5, 306.271217) (6, 306.787044) (7, 307.180137) (8, 307.476297) (9, 307.718701) (10, 307.907216)
    } ;
    \addlegendentry{FW}

    \addplot
    coordinates {
      (1, 318.084143) (2, 322.953173) (3, 325.066532) (4, 326.294976) (5, 327.082786) (6, 327.675084) (7, 328.112296) (8, 328.451162) (9, 328.707308) (10, 328.918819)
    } ;
    \addlegendentry{SW}

    \addplot
    coordinates {
      (1, 300.592203) (2, 312.838610) (3, 315.011622) (4, 316.373223) (5, 317.145859) (6, 317.759059) (7, 318.164043) (8, 318.501681) (9, 318.767608) (10, 318.989869)
    } ;
    \addlegendentry{BU}

    \addplot
    coordinates {
      (1, 300.940139) (2, 311.118538) (3, 313.365703) (4, 314.944131) (5, 315.796363) (6, 316.427208) (7, 316.849599) (8, 317.209489) (9, 317.482015) (10, 317.706735)
    } ;
    \addlegendentry{FU}

    \addplot
    coordinates {
      (1, 321.783969) (2, 331.691579) (3, 333.644911) (4, 334.848891) (5, 335.559498) (6, 336.140544) (7, 336.538960) (8, 336.878605) (9,337.172611) (9,337.232251) (10,337.240241)

    } ;
    \addlegendentry{SU}

    \addplot
    coordinates { (0, 267.063993) (11, 267.063993)};
    \addlegendentry{D}

  \end{axis}
\end{tikzpicture} 
      \label{subfig:word_form_generalized_variance_15dim}
      }
    }
    \hfill
    \subfloat[]{\scalebox{0.6} {
      \begin{tikzpicture}
  \begin{axis}[
      align =center,
      xlabel={x},
      xmin=0, xmax=11,
      ymin=230, ymax=355,
    legend cell align=left, legend style={ column sep=.5ex, font=\footnotesize },
    legend style={at={(0.00,1.02)},anchor=south west,legend columns=5},
    ]
    \addplot
    coordinates {
      (1, 317.157280) (2, 298.824235) (3, 278.118927) (4, 264.472850) (5, 252.574600) (6, 248.434260) (7, 241.212600) (8, 238.210167) (9, 233.557452) (10, 231.166731) 
    } ;
    \addlegendentry{BN}

    \addplot
    coordinates {
      (1, 319.469360) (2, 295.330725) (3, 273.754710) (4, 261.791831) (5, 249.289988) (6, 243.157605) (7, 237.247207) (8, 233.482871) (9, 228.476065) (10, 226.219079)
    } ;
    \addlegendentry{FN}

    \addplot
    coordinates {
      (1, 317.175826) (2, 319.737838) (3, 320.722541) (4, 321.307934) (5, 321.705124) (6, 321.965446) (7, 322.192873) (8, 322.375693) (9, 322.526807) (10, 322.639968)
    } ;
    \addlegendentry{BW}

    \addplot
    coordinates {
      (1, 319.488481) (2, 322.077095) (3, 323.205863) (4, 323.879094) (5, 324.335501) (6, 324.644824) (7, 324.874359) (8, 325.053647) (9, 325.200713) (10, 325.317281)
    } ;
    \addlegendentry{FW}

    \addplot
    coordinates {
      (1, 336.677381) (2, 338.854752) (3, 339.640877) (4, 340.241355) (5, 340.658192) (6, 340.915860) (7, 341.117766) (8, 341.270266) (9, 341.393247) (10, 341.488026)
    } ;
    \addlegendentry{SW}

    \addplot
    coordinates {
      (1, 317.175826) (2, 327.302809) (3, 329.272879) (4, 330.277674) (5, 330.878758) (6, 331.394528) (7, 331.746900) (8, 332.039161) (9, 332.279719) (10, 332.484468)
    } ;
    \addlegendentry{BU}

    \addplot
    coordinates {
      (1, 319.488481) (2, 328.945588) (3, 330.519786) (4, 331.484176) (5, 332.088020) (6, 332.497361) (7, 332.784831) (8, 333.027256) (9, 333.216396) (10, 333.382951)
    } ;
    \addlegendentry{FU}

    \addplot
    coordinates {
      (1, 343.154594) (2, 346.402655) (3, 347.584394) (4, 348.214473) (5, 348.631890) (6, 348.950624) (7, 349.189626) (8, 349.380751) (9, 349.532810) (10, 349.655231)
    } ;
    \addlegendentry{SU}

    \addplot
    coordinates { (0, 268.430759) (11, 268.430759)};
    \addlegendentry{D}

  \end{axis}
\end{tikzpicture} 
      \label{subfig:pos_generalized_variance}
      }
    }
    \hfill
    \subfloat[]{\scalebox{0.6} {
      \begin{tikzpicture}
  \begin{axis}[
      align =center,
      xlabel={x},
      xmin=0, xmax=11,
      ymin=230, ymax=355,
    legend cell align=left, legend style={ column sep=.5ex, font=\footnotesize },
    legend style={at={(0.00,1.02)},anchor=south west,legend columns=5},
    ]
    \addplot
    coordinates {
      (1, 298.617495) (2, 286.993586) (3, 269.409162) (4, 262.140710) (5, 250.068227) (6, 247.904657) (7, 241.998048) (8, 242.679344) (9, 236.995746) (10, 235.122067)
    } ;
    \addlegendentry{BN}

    \addplot
    coordinates {
      (1, 298.090529) (2, 286.286943) (3, 269.125959) (4, 261.901173) (5, 249.643435) (6, 248.065746) (7, 242.039403) (8, 242.725609) (9, 236.926707) (10, 235.180540)
    } ;
    \addlegendentry{FN}

    \addplot
    coordinates {
      (1, 298.617354) (2, 303.207255) (3, 305.111840) (4, 306.110736) (5, 306.849153) (6, 307.349126) (7, 307.731765) (8, 308.015343) (9, 308.251129) (10, 308.430277)
    } ;
    \addlegendentry{BW}

    \addplot
    coordinates {
      (1, 298.088637) (2, 302.666914) (3, 304.605602) (4, 305.657136) (5, 306.376816) (6, 306.880059) (7, 307.266658) (8, 307.553149) (9, 307.790177) (10, 307.976676)
    } ;
    \addlegendentry{FW}

    \addplot
    coordinates {
      (1, 318.447878) (2, 323.281892) (3, 325.383387) (4, 326.684690) (5, 327.508863) (6, 328.105055) (7, 328.534246) (8, 328.863183) (9, 329.114759) (10, 329.316475)
    } ;
    \addlegendentry{SW}

    \addplot
    coordinates {
      (1, 301.653499) (2, 313.664636) (3, 315.796641) (4, 317.127367) (5, 317.878670) (6, 318.483997) (7, 318.885732) (8, 319.222144) (9, 319.490571) (10, 319.712500)
    } ;
    \addlegendentry{BU}

    \addplot
    coordinates {
      (1, 301.489905) (2, 311.477687) (3, 313.802228) (4, 315.397602) (5, 316.269426) (6, 316.911150) (7, 317.337391) (8, 317.701014) (9, 317.976983) (10, 318.205060)
    } ;
    \addlegendentry{FU}

    \addplot
    coordinates {
      (1, 322.366616) (2, 332.268656) (3, 334.214279) (4, 335.433558) (5, 336.162627) (6, 336.753799) (7, 337.155431) (8, 337.499654) (9, 337.435254) (10, 337.354851)
    } ;
    \addlegendentry{SU}

    \addplot
    coordinates {(0,268.241582) (11,268.241582)};
    \addlegendentry{D}

  \end{axis}
\end{tikzpicture} 
      \label{subfig:word_form_pos_generalized_variance}
      }
    }

  \end{center}
  \caption{The logarithm of generalized varaiance of principal word vectors generated with different types of feature variables formed with different types of contextual features and context. The contextual features are (a) the word forms, (b) part-of-speech tags, and (c) the joint word-form and part-of-speech tag. The contexts are backward neighbourhood context (BN) forward neighbourhood context (FN), backward window-based context (BW), forward window-based context (FW), symmetric window-based context (SW), backward union context (BU), forward union context (FU), symmetric union context (SU), and dependency context (D). The horizontal axis $x$ refers to $\abs{\tau}$ for neighbouhood context and $k$ for window-based, and union context. The vertical axis $y$ shows the logarithm of generalized variance of principal word vectors.}
  \label{fig:best_fisher_15dim}
\end{figure}

\subsubsection{Data Discriminability}
In this part, we study the syntactic and semantic discriminability of the principal word vectors generated with different types of feature variables explained in \SEC{sebsec:experiments_on_feature_variables}. 
We conducted our experiments with respect to the feature sets. 
We start with the feature variables formed by the word forms and study the effect of context function on their syntactic and semantic discriminability. 
This scenario is repeated for the feature variables formed by part-of-speech tags. 
At the end, we study how the joint set of feature variables are affected by different types of context. 
The results for each series of experiments are depicted through two rows of figure which show the syntactic (top) and the semantic (bottom) discriminability of the word vectors. 

\FIG{fig:word_form_fisher} shows the values of syntactic and semantic discriminability of principal word vectors generated with the word forms and different types of context, the neighbourhood context, the window-based context, and the union context. 
\FIG{subfig:word_form_15dim_neighbourhood_syntactic_fisher} and \FIG{subfig:word_form_15dim_neighbourhood_semantic_fisher} show the syntactic and semantic discriminability obtained from the neighbourhood contexts with different value of $\tau$. 
In order to make the comparison between the contexts more convenient, the horizontal axes in these figures show the absolute value of $\tau$. 
The almost uniform trend in the results obtained from the backward neighbourhood context in \FIG{subfig:word_form_15dim_neighbourhood_syntactic_fisher} show that this context function is insensitive to the value of parameter $\tau$. 
However, the big fall in the values obtained from the forward neighbourhood context indicates that the syntactic discriminability of word vectors is negatively affected as $\tau$ increases. 
In general, we see that the backward neighbourhood context results in higher values of syntactic discriminability than the forward neighbourhood context. 
On the other hand, \FIG{subfig:word_form_15dim_neighbourhood_semantic_fisher} show that the semantic discriminability of the principal word vectors are almost insensitive to the value of parameter $\tau$. 
We see that the discrimination ratio obtained from the backward neighbourhood context starts with an upward trend but it falls down when $\tau$ is larger than $3$. 
However, the general trend of the graphs shows that the parameter $\tau$ is not an effective factor to the semantic discriminability of the word vectors generated with the word forms and the neighbourhood context. 
Moreover, we see that the semantic discrimination ratio does not differentiate between the forward context and the backward context. 
This is as opposed to what we observed in the syntactic discriminability of the word vectors generated with the neighbourhood context, where backward context was more informative than the forward context. 

\FIG{subfig:word_form_15dim_window-based_syntactic_fisher} and \FIG{subfig:word_form_15dim_window-based_semantic_fisher} show the syntactic and semantic discriminability of the principal word vectors generated with different types of window-based contexts, 
such as backward, forward, and symmetric window-based contexts with length $k=1,\dots,10$. 
The ascending trend in the results obtained from the forward and backward window-based contexts in \FIG{subfig:word_form_15dim_window-based_syntactic_fisher} show that 
the addition approach in \EQ{eq:window_based_context}, used to form a window-based context, preserves part of the \emph{syntactic} information provided by the constituent neighbourhood contexts forming the window-based context.
The symmetric window-based context is formed by the linear combination of both forward and backward window-based contexts. 
The word vectors generated with the symmetric window-based context are more influenced by the backward window-based context than the forward window-based context. 
In general, we see that the symmetric window-based context is as good as the backward window-based context and in terms of the syntactic discriminability there is no clear advantage to prioritize one over the other.
However, in terms of the computational resources, the backward window-based context with a small value of $k$ is more preferable because it results in higher sparsity in the contextual matrix and improves the computational time with smaller amount of memory. 
Similar to the neighbourhood context, the backward and symmetric window-based contexts result in higher amount of the syntactic discriminability than the forward window-based context. 
This is as opposed to the semantic discriminability of word vectors generated with the window-based context 
where both backward and forward window-based contexts result in almost similar values of Fisher discriminant ratio (see \FIG{subfig:word_form_15dim_window-based_semantic_fisher}). 
Despite the similar performance of the different types of window-based context on the semantic discriminability of the word vectors, 
we see that the symmetric window-based context consistently results in slightly higher amount of semantic discriminability. 
This agrees with the observations made by \namecite{DBLP:conf/cicling/LebretC15} where the symmetric window-based context with fairly large window size is preferred to the backward window-based context for the semantic oriented tasks.
Nevertheless, the overall trend of the variations in the results presented in \FIG{subfig:word_form_15dim_window-based_semantic_fisher} show that 
the semantic discrimination ratio of principal word vectors obtained with the symmetric window-based context is not highly sensitive to the window type and window-length $k$.  

\FIG{subfig:word_form_15dim_union_syntactic_fisher} and \FIG{subfig:word_form_15dim_union_semantic_fisher} show the syntactic and semantic discrimination ratios obtained from the principal word vectors generated with the union context with different length $k$. 
Both backward and symmetric contexts act similar to what we observed in the window-based context in both syntactic and semantic cases. 
\FIG{subfig:word_form_15dim_union_syntactic_fisher} shows that the contextual information provided by the backward context is more meaningful to determining the syntactic categories of words than the forward context. 
This is as opposed to the semantic discriminant ratios in \FIG{subfig:word_form_15dim_union_semantic_fisher} where the forward context with large value of $k$ is more meaningful to determine the semantic categories of words.
The best results of syntactic and semantic discriminant ratios are obtained from the symmetric union context with $k=1$ and $k=3$. 
We see almost no change in the values of semantic discriminability as $k$ becomes larger. 
However, increasing the value of $k$ has a negative effect on the syntactic discriminability of the principal word vectors and 

We summarize our observations on the feature variables formed by word forms and different types of context as follow.
The backward and symmetric contexts are more informative to determine the syntactic categories of words. 
The symmetric contexts are more informative to determine the semantic categories of words. 
\begin{figure}[htp]
  \begin{center}
    \subfloat[]{\scalebox{0.57} {
      \begin{tikzpicture}
  \begin{axis}[
      align = center, 
      title={neighbourhood - syntactic disc.},
      xlabel={$\abs{\tau}$},
      xtick={0,2,4,6,8,10},
      ylabel={FDR},
      xmin=0, xmax=+11,
      ymin=0.1, ymax=1.1,
      legend cell align={left},
      legend style={at={(0.98,0.1)}, anchor=south east},
    ]

    \addplot
    coordinates{(1,0.919284) (2,0.873628) (3,0.892075) (4,0.862335) (5,0.897066) (6,0.834253) (7,0.854856) (8,0.852366) (9,0.880550) (10,0.864829)} ;
    \addlegendentry{backward} 

    \addplot
    coordinates{(1,0.726389) (2,0.831409) (3,0.588623) (4,0.666617) (5,0.611682) (6,0.610961) (7,0.658304) (8,0.662770) (9,0.583317) (10,0.645100)} ;
    \addlegendentry{forward} 

%
    \addplot[mark=none, red]
    coordinates{(0,0.1359) (11,0.1359)} ;
    \addlegendentry{baseline}

  \end{axis}
\end{tikzpicture} 
      \label{subfig:word_form_15dim_neighbourhood_syntactic_fisher}
      }
    }
    \hfill
    \subfloat[]{\scalebox{0.57} {
      \begin{tikzpicture}
  \begin{axis}[
      align = center, 
      title={window-based - syntactic disc.},
      xlabel={$k$},
      xtick={0,2,4,6,8,10},
      ylabel={FDR},
      xmin=0, xmax=+11,
      ymin=0.1, ymax=1.1,
      legend cell align={left},
      legend style={at={(0.98,0.1)}, anchor=south east},
    ]
    
    \addplot
    coordinates{(1,0.919258) (2,0.929716) (3,0.931946) (4,0.931864) (5,0.944639) (6,0.943634) (7,0.942769) (8,0.942415) (9,0.941768) (10,0.941497)} ; 
    \addlegendentry{backward}
 
    \addplot
    coordinates{(1,0.726350) (2,0.755817) (3,0.763432) (4,0.767391) (5,0.850177) (6,0.843528) (7,0.840107) (8,0.838909) (9,0.837774) (10,0.837627)} ;
    \addlegendentry{forward}

    \addplot
    coordinates{(1,0.947627) (2,0.904232) (3,0.892630) (4,0.922112) (5,0.938885) (6,0.958665) (7,0.962978) (8,0.965037) (9,0.966509) (10,0.967685)} ;
    \addlegendentry{symmetric}

    \addplot[mark=none, red]
    coordinates{(0,0.1359) (11,0.1359)} ;
    \addlegendentry{baseline}


  \end{axis}
\end{tikzpicture} 
      \label{subfig:word_form_15dim_window-based_syntactic_fisher}
      }
    }
    \hfill
    \subfloat[]{\scalebox{0.57} {
      \begin{tikzpicture}
  \begin{axis}[
      align = center, 
      title={union - syntactic disc.},
      xlabel={$k$},
      xtick={0,2,4,6,8,10},
      ylabel={FDR},
      xmin=0, xmax=+11,
      ymin=0.1, ymax=1.1,
      legend cell align={left},
      legend style={at={(0.98,0.1)}, anchor=south east},
    ]
    \addplot
    coordinates{(1,0.922300) (2,0.950653) (3,0.954483) (4,0.939905) (5,0.957923) (6,0.961623) (7,0.963710) (8,0.965091) (9,0.966261) (10,0.967054)} ; 
    \addlegendentry{backward}
 
    \addplot
    coordinates{(1,0.747513) (2,0.781250) (3,0.849643) (4,0.854379) (5,0.857821) (6,0.859594) (7,0.861921) (8,0.863962) (9,0.865539) (10,0.866493)} ;
    \addlegendentry{forward}

    \addplot
    coordinates{(1,1.005219) (2,0.939335) (3,0.951830) (4,0.950708) (5,0.935463) (6,0.935034) (7,0.934435) (8,0.934168) (9,0.933854) (10,0.933412)} ;
    \addlegendentry{symmetric}

    \addplot[mark=none, red]
    coordinates{(0,0.1359) (11,0.1359)} ;
    \addlegendentry{baseline}

  \end{axis}
\end{tikzpicture} 
      \label{subfig:word_form_15dim_union_syntactic_fisher}
      }
    }
    \vfill
    \subfloat[]{\scalebox{0.57} {
      \begin{tikzpicture}
  \begin{axis}[
      align = center, 
      title={neighbourhood - semantic disc.},
      xtick={0,2,4,6,8,10},
      xlabel={$\abs{\tau}$},
      ylabel={FDR},
      xmin=0, xmax=+11,
      ymin=0.02, ymax=0.122,
      scaled y ticks=base 10:2
      legend cell align={left},
      legend style={at={(0.98,0.1)}, anchor=south east},
    ]

    \addplot
    coordinates{(1,0.098452) (2,0.103943) (3,0.109266) (4,0.100302) (5,0.100206) (6,0.097721) (7,0.100038) (8,0.103381) (9,0.100124) (10,0.102589)} ;
    \addlegendentry{backward}

    \addplot
    coordinates{(1,0.100644) (2,0.104622) (3,0.098109) (4,0.098626) (5,0.102189) (6,0.102444) (7,0.101272) (8,0.101927) (9,0.100032) (10,0.099506)} ;
    \addlegendentry{forward}

    \addplot[mark=none, red]
    coordinates{(0,0.0252) (11,0.0252)} ;
    \addlegendentry{baseline}

  \end{axis}
\end{tikzpicture} 
      \label{subfig:word_form_15dim_neighbourhood_semantic_fisher}
      }
    }
    \hfill
    \subfloat[]{\scalebox{0.57} {
      \begin{tikzpicture}
  \begin{axis}[
      align = center, 
      title={window-based - semantic disc.},
      xlabel={$k$},
      xtick={0,2,4,6,8,10},
      ylabel={FDR},
      xmin=0, xmax=+11,
      ymin=0.02, ymax=0.122,
      scaled y ticks=base 10:2
      legend cell align={left},
      legend style={at={(0.98,0.1)}, anchor=south east},
    ]
    \addplot
    coordinates{(1,0.098441) (2,0.105593) (3,0.105978) (4,0.105391) (5,0.106630) (6,0.105795) (7,0.105404) (8,0.105187) (9,0.104971) (10,0.104831) } ; 
    \addlegendentry{backward}
 
    \addplot
    coordinates{(1,0.100568) (2,0.107768) (3,0.105033) (4,0.103334) (5,0.101327) (6,0.100911) (7,0.100785) (8,0.100848) (9,0.100945) (10,0.101055)} ;
    \addlegendentry{forward}

    \addplot
    coordinates{(1,0.105845) (2,0.108903) (3,0.110846) (4,0.110777) (5,0.110750) (6,0.107576) (7,0.107660) (8,0.107657) (9,0.107646) (10,0.107643)} ;
    \addlegendentry{symmetric}

    \addplot[mark=none, red]
    coordinates{(0,0.0252) (11,0.0252)} ;
    \addlegendentry{baseline}

  \end{axis}
\end{tikzpicture} 
      \label{subfig:word_form_15dim_window-based_semantic_fisher}
      }
    }
    \hfill
    \subfloat[]{\scalebox{0.57} {
      \begin{tikzpicture}
  \begin{axis}[
      align = center, 
      title={union - semantic disc.},
      xlabel={$k$},
      xtick={0,2,4,6,8,10},
      ylabel={FDR},
      xmin=0, xmax=+11,
      ymin=0.02, ymax=0.122,
      scaled y ticks=base 10:2
      legend cell align={left},
      legend style={at={(0.98,0.1)}, anchor=south east},
    ]
    \addplot
    coordinates{(1,0.098646) (2,0.101198) (3,0.103384) (4,0.102993) (5,0.103468) (6,0.103785) (7,0.104007) (8,0.104210) (9,0.104395) (10,0.104488)} ; 
    \addlegendentry{backward}
 
    \addplot
    coordinates{(1,0.099859) (2,0.102864) (3,0.105179) (4,0.105784) (5,0.106558) (6,0.107172) (7,0.107695) (8,0.107999) (9,0.108246) (10,0.108462)} ;
    \addlegendentry{forward}

    \addplot
    coordinates{(1,0.105511) (2,0.112480) (3,0.113714) (4,0.113340) (5,0.112816) (6,0.112762) (7,0.112715) (8,0.112657) (9,0.112625) (10,0.112597)} ;
    \addlegendentry{symmetric}

    \addplot[mark=none, red]
    coordinates{(0,0.0252) (11,0.0252)} ;
    \addlegendentry{baseline}

  \end{axis}
\end{tikzpicture} 
      \label{subfig:word_form_15dim_union_semantic_fisher}
      }
    }
  \end{center}
  \caption{The syntactic (top) and semantic (bottom) Fisher discriminant ratio (FDR) of principal word vectors extracted with different types of features variables formed with word forms as contextual features and neighbourhood context (a,d), window-based context (b,e), and union context (c,f) as context function.}
  \label{fig:word_form_fisher}
\end{figure}

\begin{figure}[htp]
  \begin{center}
    \subfloat[]{\scalebox{0.57} {
      \begin{tikzpicture}
  \begin{axis}[
      align = center, 
      title={neighbourhood - syntactic disc.},
      xlabel={$\abs{\tau}$},
      xtick={0,2,4,6,8,10},
      ylabel={FDR},
      xmin=0, xmax=+11,
      ymin=0.1, ymax=1.1,
      legend cell align={left},
      legend style={at={(0.98,0.1)}, anchor=south east},
    ]
    \addplot
    coordinates{(1,0.819065) (2,0.799901) (3,0.782392) (4,0.839713) (5,0.809983) (6,0.967639) (7,0.956043) (8,1.014467) (9,0.961694) (10,1.029193)} ;
    \addlegendentry{backward} 

    \addplot
    coordinates{(1,0.918397) (2,0.780620) (3,0.541078) (4,0.619698) (5,0.606098) (6,0.617845) (7,0.650258) (8,0.668915) (9,0.614047) (10,0.629966)};
    \addlegendentry{forward} 
%
    \addplot[mark=none, red]
    coordinates{(0,0.1359) (11,0.1359)} ;
    \addlegendentry{baseline}

  \end{axis}
\end{tikzpicture} 
      \label{subfig:pos_15dim_neighbourhood_syntactic_fisher}
      }
    }
    \hfill
    \subfloat[]{\scalebox{0.57} {
      \begin{tikzpicture}
  \begin{axis}[
      align = center, 
      title={window-based - syntactic disc.},
      xlabel={$k$},
      xtick={0,2,4,6,8,10},
      ylabel={FDR},
      xmin=0, xmax=+11,
      ymin=0.1, ymax=1.1,
      legend cell align={left},
      legend style={at={(0.98,0.1)}, anchor=south east},
    ]
    \addplot
    coordinates{(1,0.817556) (2,0.811204) (3,0.819944) (4,0.839350) (5,0.852208) (6,0.859669) (7,0.861192) (8,0.861278) (9,0.861183) (10,0.860951)} ; 
    \addlegendentry{backward}
 
    \addplot
    coordinates{(1,0.927292) (2,0.868254) (3,0.870204) (4,0.870073) (5,0.869702) (6,0.869458) (7,0.869127) (8,0.868714) (9,0.868693) (10,0.868502)} ;
    \addlegendentry{forward}

    \addplot
    coordinates{(1,0.917643) (2,0.883848) (3,0.875147) (4,0.876904) (5,0.877590) (6,0.877914) (7,0.878114) (8,0.878099) (9,0.878170) (10,0.878332)} ;
    \addlegendentry{symmetric}

    \addplot[mark=none, red]
    coordinates{(0,0.1359) (11,0.1359)} ;
    \addlegendentry{baseline}

  \end{axis}
\end{tikzpicture} 
      \label{subfig:pos_15dim_window-based_syntactic_fisher}
      }
    }
    \hfill
    \subfloat[]{\scalebox{0.57} {
      \begin{tikzpicture}
  \begin{axis}[
      align = center, 
      title={union - syntactic disc.},
      xlabel={$k$},
      xtick={0,2,4,6,8,10},
      ylabel={FDR},
      xmin=0, xmax=+11,
      ymin=0.1, ymax=1.1,
      legend cell align={left},
      legend style={at={(0.98,0.1)}, anchor=south east},
    ]
    \addplot
    coordinates{(1,0.817556) (2,0.873214) (3,0.868156) (4,0.863413) (5,0.874027) (6,0.894493) (7,0.902199) (8,0.906032) (9,0.910853) (10,0.913254)} ; 
    \addlegendentry{backward}
 
    \addplot
    coordinates{(1,0.927292) (2,0.879062) (3,0.875111) (4,0.870950) (5,0.871502) (6,0.875412) (7,0.879110) (8,0.883317) (9,0.885643) (10,0.888239)} ;
    \addlegendentry{forward}

    \addplot
    coordinates{(1,0.971704) (2,0.998472) (3,1.001180) (4,1.004152) (5,1.005104) (6,1.009374) (7,1.012275) (8,1.015837) (9,1.018189) (10,1.020935)} ;
    \addlegendentry{symmetric}

    \addplot[mark=none, red]
    coordinates{(0,0.1359) (11,0.1359)} ;
    \addlegendentry{baseline}

  \end{axis}
\end{tikzpicture} 
      \label{subfig:pos_15dim_union_syntactic_fisher}
      }
    }
    \vfill
    \subfloat[]{\scalebox{0.57} {
      \begin{tikzpicture}
  \begin{axis}[
      align = center, 
      title={neighbourhood - semantic disc.},
      xtick={0,2,4,6,8,10},
      xlabel={$\abs{\tau}$},
      ylabel={FDR},
      xmin=0, xmax=+11,
      ymin=0.02, ymax=0.122,
      scaled y ticks=base 10:2
      legend cell align={left},
      legend style={at={(0.98,0.1)}, anchor=south east},
    ]

    \addplot
    coordinates{(1,0.106784) (2,0.094982) (3,0.110129) (4,0.105470) (5,0.108326) (6,0.106716) (7,0.110761) (8,0.110126) (9,0.109145) (10,0.105964)} ;
    \addlegendentry{backward}

    \addplot
    coordinates{(1,0.115364) (2,0.105665) (3,0.093769) (4,0.098296) (5,0.092935) (6,0.101004) (7,0.100890) (8,0.099065) (9,0.100142) (10,0.097518)} ;
    \addlegendentry{forward}

    \addplot[mark=none, red]
    coordinates{(0,0.0252) (11,0.0252)} ;
    \addlegendentry{baseline}

  \end{axis}
\end{tikzpicture} 
      \label{subfig:pos_15dim_neighbourhood_semantic_fisher}
      }
    }
    \hfill
    \subfloat[]{\scalebox{0.57} {
      \begin{tikzpicture}
  \begin{axis}[
      align = center, 
      title={window-based - semantic disc.},
      xtick={0,2,4,6,8,10},
      xlabel={$k$},
      ylabel={FDR},
      xmin=0, xmax=+11,
      ymin=0.02, ymax=0.122,
      scaled y ticks=base 10:2
      legend cell align={left},
      legend style={at={(0.98,0.1)}, anchor=south east},
    ]
    \addplot
    coordinates{(1,0.106784) (2,0.103898) (3,0.103183) (4,0.103309) (5,0.103598) (6,0.103932) (7,0.104010) (8,0.104020) (9,0.103987) (10,0.103989) } ; 
    \addlegendentry{backward}
 
    \addplot
    coordinates{(1,0.115270) (2,0.118418) (3,0.120050) (4,0.120127) (5,0.120166) (6,0.119989) (7,0.120002) (8,0.119984) (9,0.119981) (10,0.119979)} ;
    \addlegendentry{forward}

    \addplot
    coordinates{(1,0.111632) (2,0.108862) (3,0.107742) (4,0.107667) (5,0.107669) (6,0.107602) (7,0.107613) (8,0.107624) (9,0.107630) (10,0.107643)} ;
    \addlegendentry{symmetric}

    \addplot[mark=none, red]
    coordinates{(0,0.0252) (11,0.0252)} ;
    \addlegendentry{baseline}

  \end{axis}
\end{tikzpicture} 
      \label{subfig:pos_15dim_window-based_semantic_fisher}
      }
    }
    \hfill
    \subfloat[]{\scalebox{0.57} {
      \begin{tikzpicture}
  \begin{axis}[
      align = center, 
      title={union - semantic disc.},
      xlabel={$k$},
      xtick={0,2,4,6,8,10},
      ylabel={FDR},
      xmin=0, xmax=+11,
      ymin=0.02, ymax=0.122,
      scaled y ticks=base 10:2
      legend cell align={left},
      legend style={at={(0.98,0.1)}, anchor=south east},
    ]
    \addplot
    coordinates{(1,0.106784) (2,0.101761) (3,0.102790) (4,0.102843) (5,0.102523) (6,0.102733) (7,0.102590) (8,0.102682) (9,0.102780) (10,0.102930)} ; 
    \addlegendentry{backward}
 
    \addplot
    coordinates{(1,0.115270) (2,0.116726) (3,0.115142) (4,0.113764) (5,0.111364) (6,0.110121) (7,0.109401) (8,0.108774) (9,0.108398) (10,0.108143)} ;
    \addlegendentry{forward}

    \addplot
    coordinates{(1,0.106849) (2,0.107525) (3,0.108302) (4,0.108526) (5,0.108653) (6,0.108730) (7,0.108747) (8,0.108780) (9,0.108761) (10,0.108758)} ;
    \addlegendentry{symmetric}

    \addplot[mark=none, red]
    coordinates{(0,0.0252) (11,0.0252)} ;
    \addlegendentry{baseline}

  \end{axis}
\end{tikzpicture} 
      \label{subfig:pos_15dim_union_semantic_fisher}
      }
    }
  \end{center}
  \caption{The syntactic (top) and semantic (bottom) Fisher discriminant ratio (FDR) of principal word vectors extracted with different types of features variables formed with part-of-speech tags as contextual features and neighbourhood context (a,d), window-based context (b,e), and union context (c,f) as context function.}
  \label{fig:pos_fisher}
\end{figure}
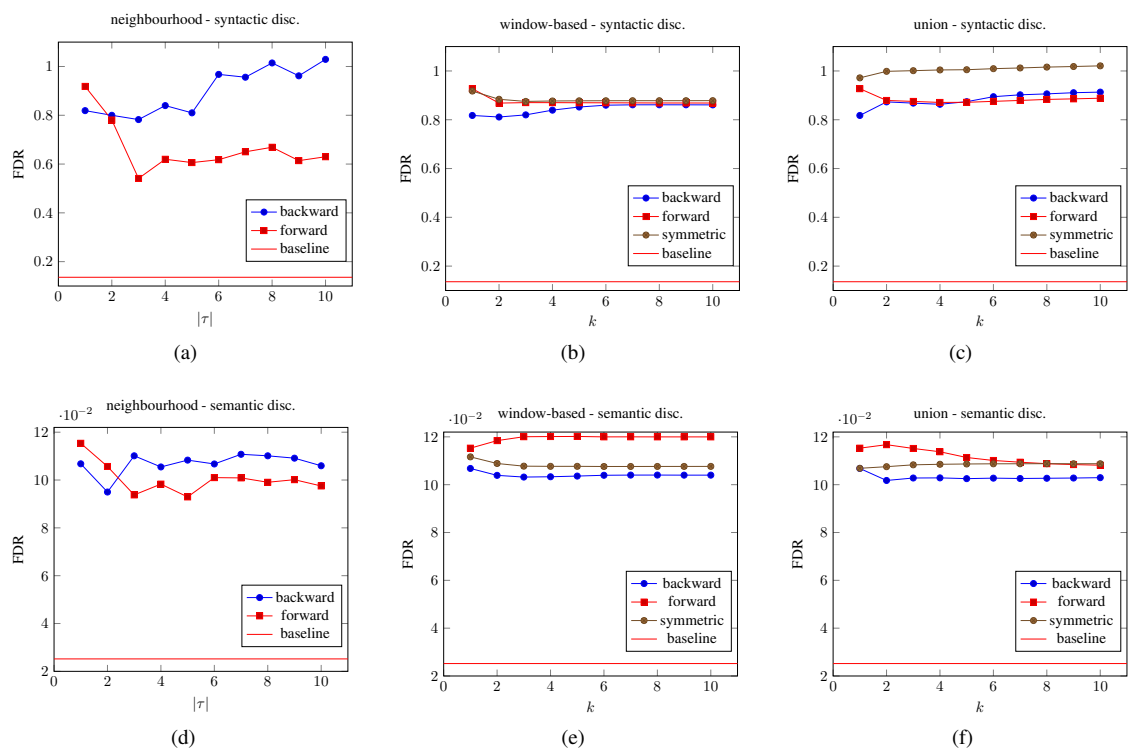

\FIG{fig:pos_fisher} shows the syntactic and semantic discriminant ratios obtained from the $15$-dimensional principal word vectors generated with part-of-speech tags as their contextual features and different types of contexts.  
\FIG{subfig:pos_15dim_neighbourhood_syntactic_fisher} and \FIG{subfig:pos_15dim_neighbourhood_semantic_fisher} show the results obtained form the neighbourhood context. 
As shown in \FIG{subfig:pos_15dim_neighbourhood_syntactic_fisher}, the forward neighbourhood context starts with a fairly high result but it drastically falls down as the value of $\tau$ increases. 
As opposed to this, the backward neighbourhood context starts with a small value of FDR but it dramatically increases as the parameter $\abs{\tau}$ increases. 
This shows that the syntactic category of the preceding words is more meaningful to the syntactic discriminability of the words than the syntactic category of the succeeding words. 
This observation that the backward context is more meaningful to the syntactic categories of words agrees with our previous observations with word forms too. 
The difference between the two observations is that the syntactic discriminability of word vectors generated with part-of-speech tags increases as the value of $\abs{\tau}$ becomes larger but the syntactic discriminability of word vectors generated with word forms is almost insensitive to the value of $\tau$. 
Nevertheless, \FIG{subfig:pos_15dim_neighbourhood_semantic_fisher} does not show the same observation for the semantic categories of words. 
\FIG{subfig:pos_15dim_neighbourhood_semantic_fisher} shows that the backward neighbourhood context with higher values of $\abs{\tau}$ does not necessarily lead to higher amount of semantic discriminability. 
The figure shows that the amount of semantic discriminability is almost insensitive to the value of $\tau$. 
The forward neighbourhood context with $\abs{\tau}=1$ results in the higher amount of semantic discriminability than the backward neighbourhood context with the same value of $\abs{\tau}$. 
This order changes as the value of $\abs{\tau}$ increases. 

\FIG{subfig:pos_15dim_window-based_syntactic_fisher} and \FIG{subfig:pos_15dim_neighbourhood_semantic_fisher} show the syntactic and semantic discriminability of word vectors generated with the backward, forward, and symmetric window-based contexts with length $k$. 
As shown in \FIG{subfig:pos_15dim_window-based_syntactic_fisher}, the difference between syntactic discriminability of the word vectors generated with different types of window-based contexts vanishes as the window length $k$ increases. 
Both forward and symmetric window-based contexts with length $k=1$ result in higher amount of syntactic discriminability than the backward window-based context. 
This is as opposed to the results obtained from word from where the backward and symmetric window-based context result in higher amount of syntactic discriminability (see \FIG{subfig:word_form_15dim_window-based_syntactic_fisher}).
\FIG{subfig:pos_15dim_window-based_syntactic_fisher} shows that increasing the value of window length $k$ has a positive effect on syntactic discriminability of word vectors generated with the backward context and a negative effect on the word vectors generated with the forward and symmetric contexts. 
However, we see that the results obtained from the backward window-based context never reaches above those obtained from the forward and symmetric window-based contexts. 
On other hand, increasing the window length $k$ has positive effect on the semantic discriminability of the word vectors generated with the forward window-based context. 
We also see a descending trend in the results obtained from backward and the symmetric contexts as the window size $k$ increases. 

\FIG{subfig:pos_15dim_union_syntactic_fisher} and \FIG{subfig:pos_15dim_union_semantic_fisher} show the results obtained from the principal word vectors generated with the union context and part-of-speech tags.
\FIG{subfig:pos_15dim_union_syntactic_fisher} shows that fairly high amount of syntactic discriminability can be obtained from the symmetric union context. 
We see a that the syntactic discriminability of word vector slightly increases as the parameter $k$ in the symmetric union context increases. 
However, \FIG{subfig:pos_15dim_union_semantic_fisher} shows that the semantic discriminability between the word vectors generated with the symmetric union context is completely insensitive to the parameter $k$. 
Among the principal word vectors generated with the part-of-speech tags and different types of union context, the bests results of the semantic discriminability is obtained from the forward union context with a small value of $k$. 
We see that there is an inverse relationship between the values of semantic discriminability obtained from the forward union context and the parameter $k$, \ie as the value of $k$ increases the value of semantic discriminability of word vectors decreases. 

We summarize our observation on the feature variables generated with the part-of-speech tags as follow.
In general, we see that the results obtained from part-of-speech tags do not act similar to the previous experiments with word forms. 
This shows that the importance of the contextual features on the quality of principal word vectors. 
The context direction, backward versus forward, does not act regularly on the syntactic and semantic discriminability of principal word vectors generated with the part-of-speech tags. 
We see that the backward neighbourhood contexts result in high values of syntactic discriminability but the backward window-based contexts result in fairly low values of syntactic discriminability. 
This observation is seen with the semantic discriminability of word vectors too. 
The window-based forward contexts result in highest amount of semantic discriminability but the neighbourhood forward contexts result in low value of semantic discriminability. 


In our last series of experiments with feature variables, we extract $15$ dimensional principal word vectors from the joint set of feature variables formed by the word forms and the part-of-speech tags. 
The word vectors are extracted with different types of context functions as before. 
\FIG{fig:word_utag_fisher} shows the values of syntactic and semantic discriminability obtained from these word vectors. 
In comparison with the previous experiments on word forms and part-of-speech tags represented in \FIG{fig:word_form_fisher} and \FIG{fig:pos_fisher}, we see that the results obtained from the joint feature variables act similar to the results obtained from the word forms shown in \FIG{fig:word_form_fisher} with slight improvements on the results. 
This shows that the joint approach of feature combination can take advantage of its constituent feature variables. 
For example, here, we see that the joint feature variable inherent the regularities of the context direction in feature variables formed by word forms and \emph{part} of the information provided by by the part-of-speech tags. 
\begin{figure}[htp]
  \begin{center}
    \subfloat[]{\scalebox{0.57} {
      \begin{tikzpicture}
  \begin{axis}[
      align = center, 
      title={neighbourhood - syntactic disc.},
      xlabel={$\tau$},
      xtick={0,2,4,6,8,10},
      ylabel={FDR},
      xmin=0, xmax=+11,
      ymin=0.1, ymax=1.1,
      legend cell align={left},
      legend style={at={(0.98,0.1)}, anchor=south east},
    ]

    \addplot
    coordinates{(1,0.974002) (2,0.918611) (3,0.914835) (4,0.912592) (5,0.913709) (6,0.866949) (7,0.897439) (8,0.892925) (9,0.917449) (10,0.901752) } ;
    \addlegendentry{backward} 

    \addplot
    coordinates{(1,0.791141) (2,0.898159) (3,0.632247) (4,0.718674) (5,0.660368) (6,0.663248) (7,0.695233) (8,0.707041) (9,0.614615) (10,0.677589)} ;
    \addlegendentry{forward} 

%
    \addplot[mark=none, red]
    coordinates{(0,0.1359) (11,0.1359)} ;
    \addlegendentry{baseline}

  \end{axis}
\end{tikzpicture} 
      \label{subfig:word_utag_15dim_neighbourhood_syntactic_fisher}
      }
    }
    \hfill
    \subfloat[]{\scalebox{0.57} {
      \begin{tikzpicture}
  \begin{axis}[
      align = center, 
      title={window-based - syntactic disc.},
      xlabel={$k$},
      xtick={0,2,4,6,8,10},
      ylabel={FDR},
      xmin=0, xmax=+11,
      ymin=0.1, ymax=1.1,
      legend cell align={left},
      legend style={at={(0.98,0.1)}, anchor=south east},
    ]
    
    \addplot
    coordinates{(1,0.973762) (2,0.979781) (3,0.979287) (4,0.976651) (5,0.979978) (6,0.981477) (7,0.982628) (8,0.983418) (9,0.983938) (10,0.984824)} ; 
    \addlegendentry{backward}
 
    \addplot
    coordinates{(1,0.790937) (2,0.800815) (3,0.811036) (4,0.835000) (5,0.856726) (6,0.863458) (7,0.870996) (8,0.875459) (9,0.877321) (10,0.878339)} ;
    \addlegendentry{forward}

    \addplot
    coordinates{(1,1.023688) (2,0.951298) (3,0.947294) (4,0.970324) (5,0.974865) (6,0.979142) (7,0.987959) (8,0.997217) (9,1.002659) (10,1.006233)} ;
    \addlegendentry{symmetric}

    \addplot[mark=none, red]
    coordinates{(0,0.1359) (11,0.1359)} ;
    \addlegendentry{baseline}


  \end{axis}
\end{tikzpicture} 
      \label{subfig:word_utag_15dim_window-based_syntactic_fisher}
      }
    }
    \hfill
    \subfloat[]{\scalebox{0.57} {
      \begin{tikzpicture}
  \begin{axis}[
      align = center, 
      title={union - syntactic disc.},
      xlabel={$k$},
      xtick={0,2,4,6,8,10},
      ylabel={FDR},
      xmin=0, xmax=+11,
      ymin=0.1, ymax=1.1,
      legend cell align={left},
      legend style={at={(0.98,0.1)}, anchor=south east},
    ]
    \addplot
    coordinates{(1,0.980986) (2,1.006339) (3,1.008313) (4,0.992405) (5,1.018183) (6,1.026426) (7,1.029866) (8,1.032633) (9,1.034294) (10,1.035675)} ; 
    \addlegendentry{backward}
 
    \addplot
    coordinates{(1,0.828842) (2,0.859784) (3,0.895125) (4,0.904345) (5,0.909424) (6,0.912654) (7,0.915975) (8,0.918427) (9,0.920246) (10,0.921551)} ;
    \addlegendentry{forward}

    \addplot
    coordinates{(1,1.062994) (2,0.994838) (3,1.008780) (4,1.012454) (5,0.999618) (6,0.998455) (7,0.998175) (8,0.997870) (9,0.997221) (10,0.997034)} ;
    \addlegendentry{symmetric}

    \addplot[mark=none, red]
    coordinates{(0,0.1359) (11,0.1359)} ;
    \addlegendentry{baseline}

  \end{axis}
\end{tikzpicture} 
      \label{subfig:word_utag_15dim_union_syntactic_fisher}
      }
    }
    \vfill
    \subfloat[]{\scalebox{0.57} {
      \begin{tikzpicture}
  \begin{axis}[
      align = center, 
      title={neighbourhood - semantic disc.},
      xtick={0,2,4,6,8,10},
      xlabel={$\tau$},
      ylabel={FDR},
      xmin=0, xmax=+11,
      ymin=0.02, ymax=0.122,
      scaled y ticks=base 10:2
      legend cell align={left},
      legend style={at={(0.98,0.1)}, anchor=south east},
    ]

    \addplot
    coordinates{(1,0.097701) (2,0.104721) (3,0.108116) (4,0.100483) (5,0.099569) (6,0.097153) (7,0.099781) (8,0.102145) (9,0.099548) (10,0.102961)} ;
    \addlegendentry{backward}

    \addplot
    coordinates{(1,0.103093) (2,0.104671) (3,0.098500) (4,0.098298) (5,0.101628) (6,0.103290) (7,0.101184) (8,0.101453) (9,0.099233) (10,0.099303)} ;
    \addlegendentry{forward}

    \addplot[mark=none, red]
    coordinates{(0,0.0252) (11,0.0252)} ;
    \addlegendentry{baseline}

  \end{axis}
\end{tikzpicture} 
      \label{subfig:word_utag_15dim_neighbourhood_semantic_fisher}
      }
    }
    \hfill
    \subfloat[]{\scalebox{0.57} {
      \begin{tikzpicture}
  \begin{axis}[
      align = center, 
      title={window-based - semantic disc.},
      xlabel={$k$},
      xtick={0,2,4,6,8,10},
      ylabel={FDR},
      xmin=0, xmax=+11,
      ymin=0.02, ymax=0.122,
      scaled y ticks=base 10:2
      legend cell align={left},
      legend style={at={(0.98,0.1)}, anchor=south east},
    ]
    \addplot
    coordinates{(1,0.097678) (2,0.104202) (3,0.104379) (4,0.104534) (5,0.102884) (6,0.102422) (7,0.102197) (8,0.102154) (9,0.101909) (10,0.101891)} ; 
    \addlegendentry{backward}
 
    \addplot
    coordinates{(1,0.102988) (2,0.104238) (3,0.102367) (4,0.103478) (5,0.102306) (6,0.101879) (7,0.101641) (8,0.101441) (9,0.101538) (10,0.101546)} ;
    \addlegendentry{forward}

    \addplot
    coordinates{(1,0.105422) (2,0.109466) (3,0.110444) (4,0.110278) (5,0.109783) (6,0.108659) (7,0.108096) (8,0.107942) (9,0.107918) (10,0.107986)} ;
    \addlegendentry{symmetric}

    \addplot[mark=none, red]
    coordinates{(0,0.0252) (11,0.0252)} ;
    \addlegendentry{baseline}

  \end{axis}
\end{tikzpicture} 
      \label{subfig:word_utag_15dim_window-based_semantic_fisher}
      }
    }
    \hfill
    \subfloat[]{\scalebox{0.57} {
      \begin{tikzpicture}
  \begin{axis}[
      align = center, 
      title={union - semantic disc.},
      xlabel={$k$},
      xtick={0,2,4,6,8,10},
      ylabel={FDR},
      xmin=0, xmax=+11,
      ymin=0.02, ymax=0.122,
      scaled y ticks=base 10:2
      legend cell align={left},
      legend style={at={(0.98,0.1)}, anchor=south east},
    ]
    \addplot
    coordinates{(1,0.098351) (2,0.101340) (3,0.103522) (4,0.103589) (5,0.103411) (6,0.103875) (7,0.104197) (8,0.104559) (9,0.104758) (10,0.104941)} ; 
    \addlegendentry{backward}
 
    \addplot
    coordinates{(1,0.101151) (2,0.104049) (3,0.104634) (4,0.105629) (5,0.106479) (6,0.107050) (7,0.107538) (8,0.107839) (9,0.108073) (10,0.108282)} ;
    \addlegendentry{forward}

    \addplot
    coordinates{(1,0.102744) (2,0.112522) (3,0.114450) (4,0.114475) (5,0.114052) (6,0.114083) (7,0.114098) (8,0.114046) (9,0.114001) (10,0.113854)} ;
    \addlegendentry{symmetric}

    \addplot[mark=none, red]
    coordinates{(0,0.0252) (11,0.0252)} ;
    \addlegendentry{baseline}

  \end{axis}
\end{tikzpicture} 
      \label{subfig:word_utag_15dim_union_semantic_fisher}
      }
    }
  \end{center}
  \caption{The syntactic (top) and semantic (bottom) Fisher discriminant ratio (FDR) of principal word vectors extracted with the joint set set of feature variables formed by the word forms and part-of-speech tags, and different types of context including the neighbourhood context (a,d), the window-based context (b,e), and the union context (c,f).}
  \label{fig:word_utag_fisher}
\end{figure}


In the last part of this section, we summarize the results obtained from the syntactic and semantic separability of the principal word vectors.
In addition to the neighbourhood context function and its two other extensions, the window-based context, and the union context, we also study the results obtained from the dependency context. 
\FIG{subfig:best_15dim_syntactic_fisher} shows the highest amount of syntactic discriminability obtained from the feature variables.
We see that in most of the cases the joint set of feature variables formed by the word forms and part-of-speech tags result in higher amount of syntactic discriminability than the features variables formed by the word forms or part-of-speech tags. 
We see that the dependency context does not yield to satisfactory results despite the expensive data preparation step to annotate the training corpus. 
This observation agrees with \cite{kiela2014systematic} where states that a window-based context with small length works better than dependency context if the vectors are extracted from a fairly large corpus. 
Interestingly, the feature variables formed with neighbourhood context and part-of-speech tags result in high amount syntactic discriminability which is as good as other more complicated contexts (\eg window-based context and union context) and contextual features (\eg joint word form and part-of-speech tag). 
This shows that in case of availability of a part-of-speech tagged corpus, one can efficiently generate a set of principal word vectors with high mount of syntactic separability using the backward neighbourhood context with large value of $\abs{\tau}$. 
\FIG{subfig:best_15dim_semantic_fisher} shows the best results of semantic discriminability obtained from the feature variables. 
We see that in most of the cases the word vectors generated with part-of-speech tags result in higher amount of semantic separability. 
Similar to the syntactic discriminability, the dependency context does not show any advantage over the other types of context. 
The best result of semantic discriminant ratio is obtained from the feature variables formed with the part-of-speech tags and window-based context with large window size $k$. 
Our final conclusion on this part is that the feature variables formed with different types of contextual features and context can be meaningful to different tasks.
\begin{figure}[htp]
  \begin{center}
    \subfloat[]{\scalebox{0.80} {
      \begin{tikzpicture}
\begin{axis}[
    ybar=0pt,
    draw opacity=0,
    bar width=0.2cm,
    enlarge x limits={abs=.5cm},
    enlarge y limits=false,
    title={Word Form - POS},
    symbolic x coords={BN, FN, BW, FW, SW, BU, FU, SU, D},
    x tick label style={
      font=\small,
      rotate=35,
      anchor=east,
      yshift=-1ex,
      name=label,
    },
    xtick=data,
    major x tick style = transparent,
    ylabel={FDR},
    ymajorgrids = true,
    ymax=1.1,
    ymin=0.0,
    legend cell align=left, legend style={ column sep=.5ex, font=\small },
    legend style={at={(0.16,1.02)},anchor=south west,legend columns=4},
]
\addplot
  coordinates {(BN, 0.9193) (FN, 0.8314) (BW, 0.9446) (FW, 0.8502) (SW, 0.9677) (BU, 0.9671) (FU, 0.8665) (SU, 1.0052) (D, 0.7281)};
\addlegendentry{Word Form}

\addplot
  coordinates {(BN, 1.0292 ) (FN, 0.9184) (BW, 0.8613) (FW, 0.9273) (SW, 0.9176) (BU, 0.9133) (FU, 0.9273) (SU, 1.0209) (D, 0.8656)};
\addlegendentry{POS}

\addplot
  coordinates {(BN, 0.97400 ) (FN, 0.89816) (BW, 0.98482) (FW, 0.87834) (SW, 1.0237) (BU, 1.0357) (FU, 0.92155) (SU, 1.0630) (D, 0.967662)};
\addlegendentry{Word Form $\times$ POS}
\end{axis}
\end{tikzpicture}
      \label{subfig:best_15dim_syntactic_fisher}
      }
    }
   \hfill
   \subfloat[]{\scalebox{0.80} {
      \begin{tikzpicture}
\begin{axis}[
    ybar=0pt,
    draw opacity=0,
    bar width=0.2cm,
    enlarge x limits={abs=.5cm},
    enlarge y limits=false,
    title={NER},
    symbolic x coords={BN, FN, BW, FW, SW, BU, FU, SU, D},
    x tick label style={
      font=\small,
      rotate=35,
      anchor=east,
      yshift=-1ex,
      name=label,
    },
    xtick=data,
    major x tick style = transparent,
    ylabel={FDR},
    ymajorgrids = true,
    ymin=0.02, ymax=0.122,
      scaled y ticks=base 10:2
    legend cell align=left, legend style={ column sep=.5ex, font=\small },
    legend style={at={(0.16,1.02)},anchor=south west,legend columns=4},
]
\addplot
  coordinates{(BN, 0.1093) (FN, 0.1046) (BW, 0.1066) (FW, 0.1078) (SW, 0.1108) (BU, 0.1045) (FU, 0.1085) (SU, 0.1137) (D, 0.1038)};
\addplot
  coordinates{(BN, 0.1108) (FN, 0.1154) (BW, 0.1068) (FW, 0.1202) (SW, 0.1116) (BU, 0.1068) (FU, 0.1167) (SU, 0.1088) (D, 0.086)};
\addplot
  coordinates{(BN, 0.10812) (FN, 0.10467) (BW, 0.10453) (FW, 0.10424) (SW, 0.11044) (BU, 0.10494) (FU, 0.10828) (SU, 0.11448) (D, 0.106258)};

\legend{Word Form, POS, Word Form $\times$ POS}
\end{axis}
\end{tikzpicture}
      \label{subfig:best_15dim_semantic_fisher}
      }
    }
  \end{center}
  \caption{The best values of fisher discriminant ratio (FDR) obtained from the principal word vectors with regard to word's part of speech tags (top) and named entities (bottom). The word vectors are extracted with two types of contetxual feature sets, word forms (left) and universal part-of-tags (right).}
  \label{fig:best_fisher}
\end{figure}

To sum up, both components of feature variables, the feature set and the context type, play important role on the discriminability of principal word vectors. 
A set of linguistically rich features (\eg part-of-speech tags) can leads to high amount of syntactic and semantic discriminability if they are used with proper context. 
The joint approach of feature combination is solution to take the advantage of different types of features. 
The results presented in this section suggest that the feature variables by themselves need to be studied more carefully. 
Although a lot of related researches are done in the area of computational semantic, we still see that most of those researches are limited to the number of features and the amount of computational resources (\ie CPU time and memory space) to process more complicated contexts and more informative features. 
The principal method of word embedding proposed in this paper provide efficient solutions about how to process such data in a systematic fashion. 
We leave the study about the more complicated and more informative features as future work. 


\subsection{Number of Dimensions}
\label{subsec:num_dim}
The number of dimensions, also known as dimensionality, of word vectors is a key factor to control amount of variation in the original contextual word vectors encoded into the principal word vectors.
The dimensionality of principal word vectors is smaller than or equal to the dimensionality of contextual word vectors, \ie number of feature variables.
Several ad hoc rules have been proposed by researched to find the smallest number of dimensions that retains most of the data variations \cite[Chapter~6.1]{jolliffe2002principal}. 
The cumulative percentage of total variance adopts the metric of \emph{total variance} to find the optimal number of dimensions. 
The total variance of a data matrix is defined as the sum of the eigenvalues of their covariance matrix. 
Denoting $\mbox{TV}_m$ as the total variance of a rank $m$ covariance matrix, the cumulative percentage of total variance look for the smallest value $k$ with $k<m$ for which we have $100\times\frac{\mbox{TV}_k}{\mbox{TV}_m}>\rho$. 
The threshold value $\rho$ is a task dependent parameter which usually takes the values between $70\%$ to $90\%$. 
\FIG{subfig:eigen_spectrum} shows the spectrum of top $1000$ eigenvalues of the covariance matrix of contextual word vectors trained with our default setting. 
The steep descent in the eigenvalues shows that most of the data variation is encoded into the first few dimensions. 
\FIG{subfig:cumsum_eigen} shows the cumulative percentage of total variance of principal word vectors with $m=1000$. 
As shown, the first few dimensions, $10$ dimensions, account for more than $99\%$ of the total variance of the data. 
So, according to the rule of cumulative percentage of total variance, the optimal number of dimensions $k$ should be smaller than $10$. 
However, as we will see later, a large amount of information is encoded into the other dimensions too that significantly affect the spread and discriminability of the principal word vectors. 
This shows that the rule of cumulative percentage of total variance does not give us the best value of $k$. 
\begin{figure}
  \begin{center} 
    \subfloat[]{\scalebox{0.8} {
        \label{subfig:eigen_spectrum}
        \input{eigen_spectrum.tex}
      }
    }
    \hfill
    \subfloat[]{\scalebox{0.8} {
        \label{subfig:cumsum_eigen}
        \input{cumsum_eigen.tex}
      }
    }
  \end{center}
  \caption{(a) The spectrum of top $1000$ eigenvalues of the covariance matrix of principal word vectors. (b) The cumulative percentage of total variance (TV).}
  \label{fig:dimensionality_eigen_spectrum}
\end{figure}

In order to mitigate the problem with the cumulative percentage of total variance and to provide better view over the data variation, we propose to use the Log-eigenvalue diagram \cite{craddock69} for finding the optimal number of dimensions. 
In this approach, the decision about the number of dimensions $k$ is made on the basis of the logarithm of the eigenvalues not the eigenvalues themselves. 
The idea behind this approach is that the logarithm of the eigenvalues corresponding to \emph{noise} in the original data should be small and close to each other. 
In other words, if we plot the diagram of the logarithm of ascendantly sorted eigenvalue, called LEV diagram, we should look for ``a point beyond which the graph becomes approximately a straight line'' \cite[p~118]{jolliffe2002principal}.

The LEV approach is closely connected to the logarithm of generalized variance used to measure the spread of the principal word vectors. 
As shown in \EQ{eq:log_gv}, the generalized variance of a set of $k$ dimensional principal word vectors is equal to the cumulative sum of the logarithm of top $k$ eigenvalues. 
Similar to the cumulative percentage of total variance, one can define the cumulative percentage of the logarithm of generalized variance as $100\times\frac{\mbox{LGV}_k}{\mbox{LGV}_m}$, where $k$ and $m$ are positive integers with $k<m$, and look for the optimal value of $k$.  
\FIG{subfig:log_eigen_spectrum} shows the LEV diagram of the $1000$ dimensional principal word vectors, and \FIG{subfig:cumsum_log_eigen} shows the values of cumulative percentage of the logarithm of generalized variance of the principal word vectors, called LGV diagram. 
Although we see a sharp drop in LEV diagram (\FIG{subfig:log_eigen_spectrum}), the LGV diagram increases in a linear way with a gentle slope of around $0.1$ (see \FIG{subfig:cumsum_log_eigen}). 
This shows that the big reduction in the spectrum of eigenvalues, as we have previously seen in \FIG{subfig:eigen_spectrum}, doesn't necessarily mean that the corresponding dimensions with small eigenvalues encode noisy information and should be eliminated. 
In fact, our experiments on the high dimensional word vectors show that increasing the number of dimensions significantly improves the spread and the discriminability of the principal word vectors. 
This agrees with our observation in the LGV diagram which shows that none of the dimensions corresponding to the top $1000$ eigenvalues encode noisy data. 
\begin{figure}
  \begin{center} 
    \subfloat[]{\scalebox{0.8} {
        \label{subfig:log_eigen_spectrum}
        \input{log_eigen_spectrum.tex}
      }
    }
    \hfill
    \subfloat[]{\scalebox{0.8} {
        \label{subfig:cumsum_log_eigen}
        \input{cumsum_log_eigen.tex}
      }
    }
  \end{center}
  \caption{(a) The LEV diagram and (b) the LGV diagram of principal word vectors with maximum $1000$ dimensions. The horizontal axis $k$ is the number of dimensions of principal word vectors (the number of principal components). LEV stands for the logaithm of eigenvalue and LGV stands for the logarithm of generalized variance.}
  \label{fig:dimensionality_log_eigen_spectrum}
\end{figure}

\FIG{subfig:dim_generalized_variance} shows the generalized variance of principal word vectors with different number of dimensions ranging from $1$ to $1000$. 
Not surprisingly, the values of generalized variance are linearly related to the cumulative percentage of LGV shown in \FIG{subfig:cumsum_log_eigen}. 
We see that the values of generalized variance are linearly affected by the number of dimensions. 
Increasing the number of dimensions $k$ leads to higher amount of generalized variance. 
\FIG{subfig:dim_discriminability} shows the syntactic and semantic discriminability of the principal word vectors with respect to the number of dimensions $k$. 
The figure shows that the amount of discriminability of principal word vectors is linearly affected by the number of dimensions. 
Both syntactic and semantic discriminability of the word vectors increases as the value of $k$ increases.
Nevertheless it is worth noting that higher number of dimensions require more computational resources to be generated and to be processed in the subsequent tasks too.
\begin{figure}
  \begin{center} 
    \subfloat[]{\scalebox{0.8} {
        \label{subfig:dim_generalized_variance}
        \input{dim_generalized_variance.tex}
      }
    }
    \hfill
    \subfloat[]{\scalebox{0.8} {
        \label{subfig:dim_discriminability}
        \begin{tikzpicture}
  \begin{axis}[
      align = center, 
      title={data discriminability},
      xlabel={$k$},
      ylabel={FDR},
      legend cell align={left},
      legend style={at={(0.02,0.98)}, anchor=north west},
    ]
  \addplot+[mark size=1.5pt]
    coordinates {
      (1, 0.204910) (5, 0.624470) (10, 0.861651) (15, 0.973079) (20, 1.089516) (25, 1.228093) (30, 1.334142) (35, 1.434523) (40, 1.475007) (45, 1.565912) (50, 1.631070) (55, 1.702127) (60, 1.792141) (65, 1.883740) (70, 2.043035) (75, 2.161665) (80, 2.210952) (85, 2.442748) (90, 2.595761) (95, 2.761697) (100, 3.044111) (110, 3.302771) (120, 3.615700) (130, 3.793339) (140, 4.112910) (150, 4.288025) (160, 4.496094) (170, 4.658883) (180, 4.895302) (190, 5.011339) (200, 5.170573) (210, 5.305592) (220, 5.429602) (230, 5.538509) (240, 5.649909) (250, 5.755041) (260, 5.860165) (270, 5.954041) (280, 6.104090) (290, 6.250845) (300, 6.354976) (310, 6.454539) (320, 6.574080) (330, 6.746541) (340, 6.916692) (350, 7.064182) (360, 7.221217) (370, 7.330132) (380, 7.460193) (390, 7.577572) (400, 7.708839) (410, 7.839349) (420, 7.957823) (430, 8.092524) (440, 8.209753) (450, 8.314485) (460, 8.436372) (470, 8.557440) (480, 8.688490) (490, 8.817352) (500, 8.933565) (550, 9.562588) (600, 10.252976) (650, 10.917564) (700, 11.701519) (750, 12.466076) (800, 13.204614) (850, 14.001765) (900, 14.740978) (950, 15.648164) (1000, 16.448202)
    } ;
    \addlegendentry{syntactic disc.}

  \addplot+[mark size=1.5pt]
    coordinates {
      (1, 0.031748) (5, 0.046844) (10, 0.087213) (15, 0.098440) (20, 0.105682) (25, 0.114574) (30, 0.130789) (35, 0.136135) (40, 0.139520) (45, 0.144334) (50, 0.154580) (55, 0.167651) (60, 0.175684) (65, 0.182607) (70, 0.193128) (75, 0.197503) (80, 0.203176) (85, 0.207439) (90, 0.211628) (95, 0.218110) (100, 0.223583) (110, 0.236223) (120, 0.244138) (130, 0.255886) (140, 0.264249) (150, 0.272740) (160, 0.281721) (170, 0.288030) (180, 0.294904) (190, 0.302592) (200, 0.310981) (210, 0.317772) (220, 0.325508) (230, 0.333038) (240, 0.339085) (250, 0.346356) (260, 0.353639) (270, 0.360166) (280, 0.366700) (290, 0.373726) (300, 0.380034) (310, 0.386922) (320, 0.394812) (330, 0.400019) (340, 0.407043) (350, 0.414393) (360, 0.421020) (370, 0.428553) (380, 0.436416) (390, 0.445770) (400, 0.452267) (410, 0.458693) (420, 0.467754) (430, 0.475163) (440, 0.482710) (450, 0.489679) (460, 0.497117) (470, 0.503781) (480, 0.511516) (490, 0.519357) (500, 0.526332) (550, 0.564787) (600, 0.609773) (650, 0.647626) (700, 0.690825) (750, 0.738338) (800, 0.785153) (850, 0.828669) (900, 0.880953) (950, 0.929964) (1000, 0.983338)
    } ;
    \addlegendentry{semantic disc.}

  \end{axis}
\end{tikzpicture} 
      }
    }
  \end{center}
  \caption{(a) The logarithm of generalized variance of the $k$ dimensional principal word vectors. (b) The syntactic and semantic discriminability of the $k$ dimensional principal word vectors.}
  \label{fig:dimensionality_performance}
\end{figure}

\subsection{Weighting and Transformation}
\label{subsec:weight_and_trans}
In this section, we study how the principal word vectors are affected by the transformation step in Line~\ref{alg:gpca:transformation} at \ALG{alg:gpca}.
Our experiments in this part can be divided into two major types.
First is the experiments that are focused on the data spread and data discriminability. 
Second is the experiments that focus on the contribution of the word vectors on the word similarity benchmark and the dependency parsing. 
In the former series of experiments, the eigenvalue weighting matrix $\Lambda$ is set as in \EQ{eq:classic_Lambda}. 
In the second series of experiments, the matrix is set as in \EQ{eq:normal_Lambda}. 
This is because dependency parser used in these experiments uses a feed-forward neural network as its classifier, as mentioned in \SEC{sec:exp_setting}. 

The transformation is done through three parameters, the weight matrices $\Phi$, and $\Omega$ and the transformation function $\mathcal{f}$. 
As explained in \SEC{subsubsection:weight_matrices}, $\Omega$ is a weight matrix over the observations and $\Phi$ is a metric matrix used to scale the elements of contextual word vectors. 
In \TABLE{table:metric_matrix}, we proposed different diagonal matrices that can be used as $\Omega$ and $\Phi$. 
As explained in \SEC{subsubsection:trans_func}, the transformation function $\mathcal{f}$ adds some degree of non-linearity to the principal word embedding model. 
Although multiple transformation function can be tested, we restrict our experiments on the power transformation function.
The power transformation function is defined in two ways. 
First is to use a vector of power values whose elements correspond to the elements of contextual word vectors. 
Denoting $\boldsymbol{\mathcal{v}}=(\mathcal{v}_1,\dots,\mathcal{v}_m)$ as a contextual word vector, the power transformation function defined with the power vectors $\theta=(p_1,\dots,p_m)$ maps $\boldsymbol{\mathcal{v}}$ to $\mathcal{f}(\boldsymbol{\mathcal{v}}; \theta)=(\mathcal{v}_1^{p_1},\dots,\mathcal{v}_m^{p_m})$.
In this case, the optimal value of $\hat\theta$ which maximizes \EQ{eq:obj_trans_func} is estimated by simulated annealing with $p_i\in(0,1]$. 
Second is to use a single power value for all elements of the contextual word vectors.
Using the same notation as before, the power transformation function defined with the single power value $\theta=p$ maps $\boldsymbol{\mathcal{v}}$ to $\mathcal{f}(\boldsymbol{\mathcal{v}}; \theta)=(\mathcal{v}_1^{p},\dots,\mathcal{v}_m^{p})$.
The optimal value of power in this case too is estimated by simulated annealing through maximizing \EQ{eq:obj_trans_func} with $p\in(0,1]$. 

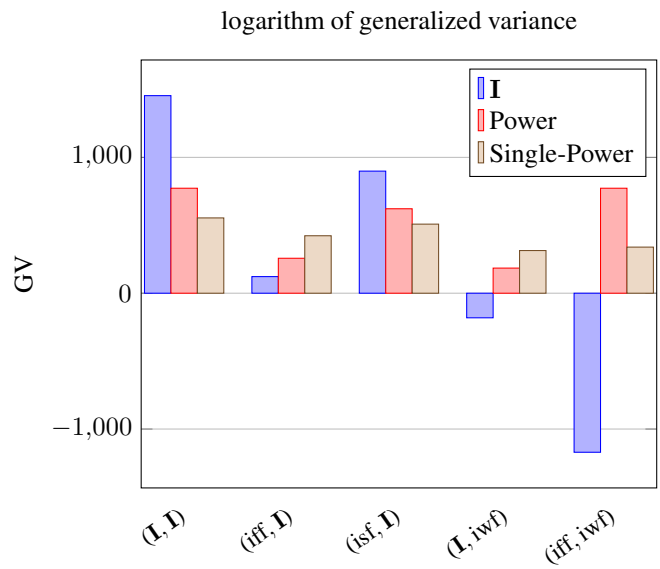
\begin{figure}
  \begin{center} 
      \pgfplotstableread[row sep=\\,col sep=&,comment chars=!]{
Weight-Matrix                         & I         & Logarithm & Hellinger & Normal & NormalSinglePower \\
{($\mathbf{I},\mathbf{I}$)}           & 1454.58   & 631.50318 & 806.02302 & 772.94 & 554.37815 \\
{($\mbox{iff},\mathbf{I}$)}           & 122.32    & 810.98112 & 198.41880 & 257.35 & 423.07776 \\
{($\mbox{isf},\mathbf{I}$)}           & 898.87    & 739.67827 & 614.86816 & 621.79 & 508.58249 \\
{($\mathbf{I},\mbox{iwf}$)}           & -181.04   & 832.61730 & 97.64829  & 184.91 & 314.12313 \\
{($\mbox{iff},\mbox{iwf}$)}           & -1170.93  & 885.08269 & -569.64304& 772.94 & 339.45462 \\
}\WEIGHTGENERALIZEDVARIANCETABLE 

\begin{tikzpicture}
\begin{axis}[
    ybar=0pt,
    xtick=data,
    xticklabels from table={\WEIGHTGENERALIZEDVARIANCETABLE}{Weight-Matrix},
   title={logarithm of generalized variance},
    x tick label style={
      font=\small,
      rotate=35,
      anchor=east,
      yshift=-1ex,
      name=label,
    },
    xtick=data,
    major x tick style = transparent,
    ylabel={GV},
    ymajorgrids = true,
    legend cell align=left, 
]
\addplot table[x expr=\coordindex, y=I]{\WEIGHTGENERALIZEDVARIANCETABLE} ;
\addlegendentry{$\mathbf{I}$}



\addplot table[x expr=\coordindex, y=Normal]{\WEIGHTGENERALIZEDVARIANCETABLE} ;
\addlegendentry{Power}

\addplot table[x expr=\coordindex, y=NormalSinglePower]{\WEIGHTGENERALIZEDVARIANCETABLE} ;
\addlegendentry{Single-Power}

\end{axis}
\end{tikzpicture}
  \end{center}
  \caption{The logarithm of generalized variance of principal word vectors generated with different types of weight matrices and transformation functions.}
  \label{fig:weight_log_generalized_variance}
\end{figure}

\FIG{fig:weight_log_generalized_variance} shows the values of generalize variance of principal word vectors with respect to different combinations of the weight matrices and the transformation function. 
In addition to the results obtained with the power transformation function, the figure shows the results obtained from the identity transformation function $\mathbf{I}$, which basically means no transformation. 
The identity transformation function $\mathbf{I}$ returns its input with no change. 
We see that the spread of principal word vectors is highly influenced by both the weight matrices and the transformation function. 
The highest amount of data spread is obtained from the principal word vectors with no transformation, $\Phi=\mathbf{I}$, $\Omega=\mathbf{I}$, and the transformation function $\mathcal{f}=\mathbf{I}$.
Most of this high amount of data spread is due to the imbalanced and excessive contribution of the top eigenvalues in \EQ{eq:log_gv}.
As elaborated in \SEC{subsubsection:trans_func}, the transformation function in \EQ{eq:obj_trans_func} reduces the excessive effect of the top eigenvalues through compressing the data distribution along the top eigenvectors and expanding the distribution along the other eigenvectors. 
This data normalization leads to some reduction in the generalized variance of the principal word vectors (see the bins related to $(\mathbf{I},\mathbf{I})$ and the power transformation function). 
This reduction in the data spread is due to the very large data compression along the top eigenvalues.
The second pair of bins in \FIG{fig:weight_log_generalized_variance} ($\Phi=\mbox{iff}$, and $\Omega=\mathbf{I}$) shows the effect of inverse of feature frequency on the data spread. 
We see that the weighting matrix leads to a large compression in the principal word vectors which is then slightly expanded by the power transformation function. 
The third pair of bins in \FIG{fig:weight_log_generalized_variance} ($\Phi=\mbox{isf}$, and $\Omega=\mathbf{I}$) shows how the spread of the principal word vectors are affected by the inverse of standard deviation of the feature variables. 
In other words, it shows how the spread of the principal word vectors is affected if we compute the principal word vectors from the correlation matrix of contextual word vectors instead of using their covariance matrix. 
As mentioned in \SEC{subsubsection:weight_matrices}, performing PCA on the correlation matrix instead of the covariance matrix is a solution to mitigate the imbalanced contribution of feature variables in the data spread. 
We see that the spread of principal word vectors generated from the correlation matrix with no transformation is close to the spread of principal word vectors generated from the covariance matrix with power transformation (see the blue bin in the first pair of bins in \FIG{fig:weight_log_generalized_variance}). 
In terms of data spread, this means that the effect of the power transformation function on the distribution of principal word vectors is as good as the effect of performing PCA on the correlation matrix.
However, as we will see later, this does not mean that they will necessarily result in the same amount of data discriminability. 
Performing the power transformation on the contextual word vectors normalized by the standard deviation of their feature variables reduces the spread of resulting principal word vectors. 
Part of this reduction is due to the randomness involved in the estimation of the entropies in \EQ{eq:obj_trans_func}.
Part of that can also be due to the fact the metric matrix $\Phi=\mbox{iff}$ normalizes the contextual word vectors along their basis vectors which are associated with the feature variables. 
However, the transformation function normalizes the data along the eigenvectors of their covariance matrix which are not necessarily equal to the basis vectors. 
The two remaining pairs of bins show the effect weight matrix $\Omega$ on the spread of principal word vectors.
The negative value of the data spread shows that most of the eigenvalues describing the data spread are smaller than one. 
This means that the spread of principal word vectors reduces drastically when we weight the observations with their inverse frequency. 
However, this negative effect of the weight matrix $\Omega$ on the spread of principal word vectors is largely cancelled out by the power transformation function. 
We see that the spread of the principal word vectors after performing power transformation is comparable to the other sets of word vectors.
In general we see that the power transformation normalizes normalizes the data in the expected way, \ie it compresses the over expanded data and expands highly massed data.

\begin{figure}
  \begin{center} 
    \subfloat[]{\scalebox{0.8} {
        \label{subfig:weight_pos_fisher}
        \pgfplotstableread[row sep=\\,col sep=&,comment chars=!]{
Trans-Function                        & I         & Logarithm & Hellinger & Normal  & NormalSinglePower \\
{($\mathbf{I},\mathbf{I}$)}           & 2.9757    & 3.317244  & 3.702663  & 3.9226  & 2.00878 \\
{($\mbox{iff},\mathbf{I}$)}           & 2.9426    & 2.651692  & 3.533038  & 3.2192  & 2.05397 \\
{($\mbox{isf},\mathbf{I}$)}           & 2.7938    & 1.341958  & 3.696216  & 3.5856  & 2.38434 \\
{($\mathbf{I},\mbox{iwf}$)}           & 0.013352  & 1.634448  & 0.135873  & 1.1251  & 0.78494 \\
{($\mbox{iff},\mbox{iwf}$)}           & 0.00000000& 1.533368  & 0.000748  & 1.2282  & 1.31823 \\
}\POSWEIGHTTRANSTABLE 

\begin{tikzpicture}
\begin{axis}[
    ybar=0pt,
    xtick=data,
    xticklabels from table={\POSWEIGHTTRANSTABLE}{Trans-Function},
   title={syntactic discriminability},
    x tick label style={
      font=\small,
      rotate=35,
      anchor=east,
      yshift=-1ex,
      name=label,
    },
    xtick=data,
    major x tick style = transparent,
    ylabel={FDR},
    ymajorgrids = true,
    ymax=4,
    ymin=0.0,
    legend cell align=left, legend style={ column sep=.5ex },
]
\addplot table[x expr=\coordindex, y=I]{\POSWEIGHTTRANSTABLE} ;
\addlegendentry{$\mathbf{I}$}
\addplot table[x expr=\coordindex, y=Normal]{\POSWEIGHTTRANSTABLE} ;
\addlegendentry{Power}
\addplot table[x expr=\coordindex, y=NormalSinglePower]{\POSWEIGHTTRANSTABLE} ;
\addlegendentry{Single-Power}

%
%
%
%

\end{axis}
\end{tikzpicture}
      }
    }
    \hfill
    \subfloat[]{\scalebox{0.8} {
        \label{subfig:weight_ner_fisher}
        \pgfplotstableread[row sep=\\,col sep=&,comment chars=!]{
Trans-Function                        & I         & Logarithm & Hellinger & Normal  & NormalSinglePower\\
{($\mathbf{I},\mathbf{I}$)}           & 0.223702  & 0.502478  & 0.344210  & 0.37913 & 0.48947 \\
{($\mbox{iff},\mathbf{I}$)}           & 0.206106  & 0.351612  & 0.306392  & 0.31040 & 0.44816 \\
{($\mbox{isf},\mathbf{I}$)}           & 0.178800  & 0.404778  & 0.287383  & 0.31413 & 0.44160 \\
{($\mathbf{I},\mbox{iwf}$)}           & 0.015327  & 0.402288  & 0.089260  & 0.27509 & 0.35734 \\
{($\mbox{iff},\mbox{iwf}$)}           & 0.000000  & 0.430770  & 0.001805  & 0.26117 & 0.48723 \\
}\NERWEIGHTTRANSTABLE 

\begin{tikzpicture}
\begin{axis}[
    ybar=0pt,
    title={semantic discriminability},
%
    xtick=data,
    xticklabels from table={\NERWEIGHTTRANSTABLE}{Trans-Function},
    x tick label style={
      font=\small,
      rotate=35,
      anchor=east,
      yshift=-1ex,
      name=label,
    },
    xtick=data,
    major x tick style = transparent,
    ylabel={FDR},
    ymajorgrids = true,
    ymax=0.5,
    ymin=0.0,
    legend cell align=left, 
]

\addplot table[x expr=\coordindex, y=I]{\NERWEIGHTTRANSTABLE} ;
\addlegendentry{$\mathbf{I}$}
\addplot table[x expr=\coordindex, y=Normal]{\NERWEIGHTTRANSTABLE} ;
\addlegendentry{Power}
\addplot table[x expr=\coordindex, y=NormalSinglePower]{\NERWEIGHTTRANSTABLE} ;
\addlegendentry{Single-Power}

%
%
%

\end{axis}
\end{tikzpicture}
      }
    }
  \end{center}
  \caption{The (a) syntactic and (b) semantic discriminability of principal word vectors generated with different types of weight matrices and transformation functions.}
  \label{fig:weighting_discriminability}
\end{figure}
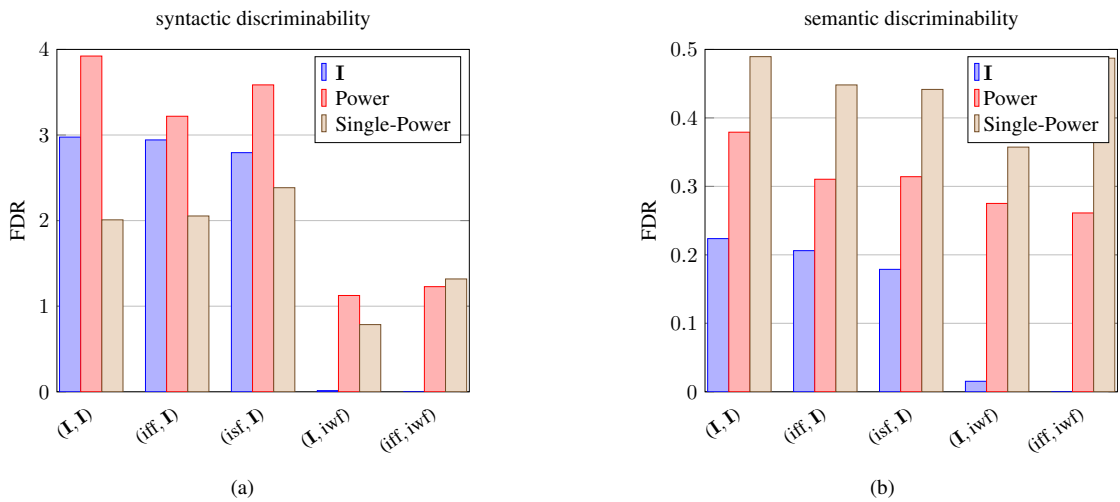

We study the effect of the weighting mechanism and the transformation function on the discriminability of principal word vectors. 
\FIG{fig:weighting_discriminability} shows the amount of syntactic and semantic discriminability of principal word vectors obtained from different weight matrices and transformation functions as above. 
As shown in the figures, the power transformation leads to a significant improvement in both syntactic and semantic discriminability of the principal word vectors. 
This improvement is more clear in the cases where the amount of discriminability is very small, $\Omega=\mbox{iwf}$. 
All weight matrices reduces the discriminability of principal word vectors. 
In fact the best the results are obtained from the identity weight matrices, $\Phi=\mathbf{I}$ and $\Omega=\mathbf{I}$, and the power transformation function. 
We see that the feature weighting, which is done by $\Phi$, is more meaningful to the syntactic and semantic discriminability of word vectors then the observation weighting, which is done by $\Omega$. 

In the remaining part of this section, we study how the weighting mechanism and the transformation function affect the performance of the principal word vectors on the third party tasks such as word the similarity benchmark and the dependency parsing.
These tasks are outlined in \SEC{sec:exp_setting}. 
As mentioned above, the eigenvalue weighting matrix $\Lambda$ in these experiments is set as in \EQ{eq:normal_Lambda} with $\alpha=0.1$.
The parameter $\alpha$ is set on the basis of the standard deviation of the initial weights of the neural network classifier used in the parser.
\FIG{fig:weight_sim_rank} summarizes the results obtained from the similarity benchmark. 
The vertical axis is the average of similarity correlations. 
As shown, the power transformations contribute more than the weight matrices to the task. 
The best results are obtained from the single-power transformation function with identity weight matrices. 
Almost the same result is also obtained from the single-power transformation with $\Phi=\mbox{iff}$ and $\Omega=\mathbf{I}$. 
Similar to the data discriminability, we see that the weighting matrix $\Omega=\mathbf{I}$ has negative effect on the results. 
Given these observations, one might ask if the observation weighting always has a negative effect on the performance of principal word vectors. 
We will show that the results completely depend on the task and different weight matrices might be useful for certain tasks. 
The study about the meaningfulness of weight matrices to certain tasks is out of the scape of this research and we leave it as future work. 
In this research, we experimentally show that different types of weight matrices are meaningful to different tasks. 
\begin{figure}[ht!]
  \begin{center}
    \pgfplotstableread[row sep=\\,col sep=&,comment chars=!]{
Trans-Function                        & I         & Logarithm & Hellinger & Normal  & NormalSinglePower \\
{($\mathbf{I},\mathbf{I}$)}           & 0.232     &           &           & 0.382     & 0.452 \\
{($\mbox{iff},\mathbf{I}$)}           & 0.175     &           &           & 0.221     & 0.451 \\
{($\mbox{isf},\mathbf{I}$)}           & 0.265     &           &           & 0.437     & 0.449 \\
{($\mathbf{I},\mbox{iwf}$)}           & 0.048     &           &           & 0.066     & 0.243 \\
{($\mbox{iff},\mbox{iwf}$)}           & 0.020     &           &           & 0.141     & 0.396 \\
}\WSIMWEIGHTTRANSTABLE 

\begin{tikzpicture}
\begin{axis}[
    ybar=0pt,
    xtick=data,
    xticklabels from table={\WSIMWEIGHTTRANSTABLE}{Trans-Function},
   title={word similarity},
    x tick label style={
      font=\small,
      rotate=35,
      anchor=east,
      yshift=-1ex,
      name=label,
    },
    xtick=data,
    major x tick style = transparent,
    ylabel={Avg. Correlation},
    ymajorgrids = true,
    ymax=0.5,
    ymin=0.0,
    legend cell align=left, legend style={ column sep=.5ex },
]
\addplot table[x expr=\coordindex, y=I]{\WSIMWEIGHTTRANSTABLE} ;
\addlegendentry{$\mathbf{I}$}
\addplot table[x expr=\coordindex, y=Normal]{\WSIMWEIGHTTRANSTABLE} ;
\addlegendentry{Power}
\addplot table[x expr=\coordindex, y=NormalSinglePower]{\WSIMWEIGHTTRANSTABLE} ;
\addlegendentry{Single-Power}

%
%
%
%

\end{axis}
\end{tikzpicture}
  \end{center}
  \label{fig:weight_sim_rank}
  \caption{The average of the results obtained from the word similarity benchmark.}
\end{figure}

\FIG{fig:weight_parsing} shows the parsing results obtained from the principal word vectors generated with different weighting matrices and transformation functions. 
The figure shows the unlabelled attachment scores obtained from the Stanford dependency parser \cite{chen14} on the development set of WSJ.
Similar to the previous experiments, we see that the best results are obtained from the single-power transformation function. 
However, unlike the previous experiments, we see that the best result is obtained from $\Phi=\mbox{iff}$ and $\Omega=\mbox{iwf}$. 
This observation confirms the importance of weighting matrices on different tasks. 
\begin{figure}[ht!]
  \begin{center}
    \pgfplotstableread[row sep=\\,col sep=&,comment chars=!]{
Trans-Function                        & I         & Logarithm & Hellinger & Normal  & NormalSinglePower \\
{($\mathbf{I},\mathbf{I}$)}           & 90.4      &           &           & 90.8     & 91.9 \\
{($\mbox{iff},\mathbf{I}$)}           & 90.5      &           &           & 90.7     & 91.6 \\
{($\mbox{isf},\mathbf{I}$)}           & 90.4      &           &           & 91.0     & 91.8 \\
{($\mathbf{I},\mbox{iwf}$)}           & 90.7      &           &           & 91.5     & 91.3 \\
{($\mbox{iff},\mbox{iwf}$)}           & 90.8      &           &           & 91.7     & 92.2 \\
}\DPARSEWEIGHTTRANSTABLE 

\begin{tikzpicture}
\begin{axis}[
    ybar=0pt,
    xtick=data,
    xticklabels from table={\DPARSEWEIGHTTRANSTABLE}{Trans-Function},
   title={dependency parsing},
    x tick label style={
      font=\small,
      rotate=35,
      anchor=east,
      yshift=-1ex,
      name=label,
    },
    xtick=data,
    major x tick style = transparent,
    ylabel={UAS},
    ymajorgrids = true,
    ymax=92.5,
    ymin=88.0,
    legend cell align=left, legend style={ column sep=.5ex },
    legend style={fill=none},
]
\addplot table[x expr=\coordindex, y=I]{\DPARSEWEIGHTTRANSTABLE} ;
\addlegendentry{$\mathbf{I}$}
\addplot table[x expr=\coordindex, y=Normal]{\DPARSEWEIGHTTRANSTABLE} ;
\addlegendentry{Power}
\addplot table[x expr=\coordindex, y=NormalSinglePower]{\DPARSEWEIGHTTRANSTABLE} ;
\addlegendentry{Single-Power}

\end{axis}
\end{tikzpicture}
  \end{center}
  \label{fig:weight_parsing}
  \caption{The unlabelled attachment on the development set of WSJ.}
\end{figure}

\subsection{Comparison}
In this section, we compare the results obtained from the principal word vectors and other sets of word embeddings collected by popular methods of word embedding such as word2vec \cite{NIPS2013_5021}, GloVe \cite{pennington2014glove}, HPCA \cite{lebret-collobert:2014:EACL}, and random indexing (RI) \cite{sahlgren2006word}. 
We use word2vec in two modes, the continuous bag of words CBOW and the skip gram SGRAM.
We use our implementation of HPCA and RI.
The comparison is on the basis of 
\begin{enumerate} 
  \item the spread of word vectors in terms of the logarithm of generalized variance (see \EQ{eq:log_gv}),
  \item the syntactic and the semantic of discriminability of the word vectors,
  \item the performance of the word vectors in the word similarity benchmark, and
  \item the contribution of the word vectors in the task of dependency parsing.
\end{enumerate}
All sets of word vectors are extracted from the same raw corpus described in \SEC{sec:exp_setting}. 
Except for the set of word vectors extracted by word2vec, all embeddings are trained with the same setting, backward neighbourhood with length $1$, \ie the context refer to the immediate preceding word and the contextual features are the word forms.
word2vec is trained with symmetric neighbourhood context of length one, \ie the context forms with the immediate preceding and succeeding words. 
The number of iterations in GLOVE is set to $50$, in word2vec is set to $1$. 
All methods are trained with $5$ threads, if multi-threading is supported.

\TABLE{table:comparison} summarizes the results. 
The table is divided into two parts. 
The last four rows show the results obtained from the principal word vectors with different combinations of weighting matrices $\Phi$, $\Omega$, and transformation function $\mathcal{f}$ which are represented by the triple $(\Phi,\Omega,\mathcal{f})$. 
Among the all combinations of $(\Phi,\Omega,\mathcal{f})$, we have chosen the combinations which result in highest results in each of the evaluation metrics. 
The comparison on the parsing is made on the test set of the WSJ journal. 

\begin{table}[htbp]
  \begin{tabularx}{\textwidth}{@{}XYYYYYY@{}}
    \hline
     & \multicolumn{1}{Y}{Log. GV} & \multicolumn{1}{Y}{Syn. Disc.} & \multicolumn{1}{Y}{Sem. Disc.} & \multicolumn{1}{Y}{Sim. Corr.} & \multicolumn{1}{Y}{UAS} & \multicolumn{1}{Y}{LAS} \\ \hline
     RI                                   & 202           & 2.6           & 0.2           & 0.2 & 90.5 & 88.2 \\ 
     HPCA                                 & 347           & 2.1           & \textbf{0.5}  & 0.2 & 90.7 & 88.6 \\ 
     CBOW                                 & 622           & 0.8           & 0.1 & 0.5 & \textbf{92.1} & \textbf{90.1} \\ 
     SGRAM                                & 564           & 0.9           & 0.2 & \textbf{0.6} & \textbf{92.1} & \textbf{90.0} \\ 
     GLOVE                                & 525           & 0.9           & 0.1           & 0.5 & 91.9 & 89.9 \\ \hline \hline
     $(\mathbf{I},\mathbf{I},\mathbf{I})$ & \textbf{1454} & 3.0           & 0.2           & 0.2 & 89.9 & 87.6 \\ 
     $(\mathbf{I},\mathbf{I},\mbox{p})$   & 772        & \textbf{3.9}  & 0.4           & 0.4 & 90.5 & 88.3 \\ 
     $(\mathbf{I},\mathbf{I},\mbox{sp})$  & 554         & 2.0           & \textbf{0.5}  & 0.5 & 91.5 & 89.4 \\ 
     $(\mbox{iff},\mbox{iwf},\mbox{sp})$  & 339         & 1.3           & \textbf{0.5}  & 0.4 & 91.9 & 89.9 \\ 
  \end{tabularx}
  \caption{The comparison between principal word vectors and other sets of word vectors. The results obtained from the principal word vectors are shown in the second part of the table, below the double line. The triples show the certain settings of parameters $(\Phi,\Omega,\mathcal{f})$, where $\Phi$ and $\Omega$ are the weighting matrices, and $\mathcal{f}$ is the transformation function. $\mathcal{f}=\mbox{p}$ and $\mathcal{f}=\mbox{sp}$ refer to the power transformation function with vector of power values and single power value respectively. UAS and LAS stand for the unlabelled and labelled attachment scores respectively. The parsing results are related to the test set of WSJ.}
  \label{table:comparison}
\end{table}

The results show that the spread of principal word vectors generated with $(\mathbf{I},\mathbf{I},\mathbf{I})$ is significantly higher than the other sets of word vectors. 
However, this set of word vectors shows poor performance on the other evaluation metrics. 
In terms of syntactic discriminability, we see that the principal word vectors trained with identity weight matrices and the vector power transformation results in highest value of syntactic discriminability. 
The best results of semantic discriminability are obtained from HPCA and the principal word vectors with single power transformation function. 
In terms of word similarities, we see that principal word vectors are on par with the vectors generated with word2vec and GloVe. 
The principal word vectors and the vectors generated by GloVe result in the same parsing scores which are significantly higher than the results obtained from the RI and HPCA word vectors. 
However, we see that the parsing scores are slightly less than the results obtained from the word2vec. 
In order to see if the slight superiority of word2vec is due to chance, we perform an statistical significance test on the parsing results using the \namecite{Berg-Kirkpatrick:2012:EIS:2390948.2391058} 's method. 
\TABLE{table:p-value} shows the \emph{p-value} under the null hypothesis $H_0$: 
\emph{A is not better than B}, where $A$ refers to the set of principal word vectors generated with $(\mbox{iff},\mbox{iwf},\mbox{sp})$ and $B$ can be any of the other word embeddings methods mentioned above. 
The table shows that the null hypothesis is rejected with high confidence for RI and HPCA, but not for other methods.
This confirms the superiority of principal word vectors to HPCA and RI, and rejects the superiority of principal word vectors to the other methods. 
It also shows that the superiority of word2vec to the principal word vectors is not statistically significant. 
\begin{table}[htbp]
  \begin{center}
    \begin{tabularx}{\textwidth}{XXXXXXX}
    \hline
        & RI      & HPCA    & CBOW  & SGRAM & GLOVE  \\ \hline
    $p$ & $0.00$  & $0.00$  & $0.55$  & $0.60$  & $0.65$ \\ \hline
  \end{tabularx}
    \caption{p-value of the null hypothesis $H_0$: \emph{principal word vectors are not better than the other sets of word vectors on the task of dependency parsing}.}
  \end{center}
  \label{table:p-value}
\end{table}

Now we turn our attention to the efficiency of the word embedding methods. 
In addition to the mathematical formulation of a problem implemented in a software, the efficiency of the software system depends on many other factors such as memory management, parallelization, and how the system is divided into subsystems. 
Not all of the word embedding methods we listed above follows the same software architecture. 
Some these methods such as HPCA, and GloVe divide the task of word embedding into three main sub-tasks. 
First is to count the frequency of seeing in the training corpus. 
Second is to build a co-occurrence matrix.
Third is to extract word vectors from the co-occurrence matrix through performing a dimensionality reduction technique. 
If our implementation of HPCA, the first two steps are implemented as two single threaded programs and the third step is implemented as a multi-threaded program. 
Each of these tasks are implemented in HPCA as individual multi-threaded programs. 
GloVe implements the first two tasks with two single threaded programs but the third task  is implemented with a multi-threaded program. 
The division between the second and the third tasks enables these methods to perform the algorithm of dimensionality reduction once on the co-occurrence matrix which models the entire training corpus. 
On the other side, both RI and word2vec perform the second and the third task together which can be processed it in a multi-threaded way. 
This makes these methods highly efficient in terms of memory usage, since they don't need to explicitly build the co-occurrence matrix in the memory. 
However, this efficiency in the memory usage is in the cost of CPU time. 
This because both methods need to update their internal states, representing the parameters of the methods, after seeing a batch of words. 
Nevertheless, these methods does not provide any clear solution to make use of extra memory in order to reduce the CPU time. 

In our experiments, word2vec and GloVe are used with no change, but HPCA and RI are reimplemented as below.
In our implementation of HPCA, the first two steps, counting the vocabularies and building the co-occurrence matrix, are implemented as two individual single threaded programs, and the third step, dimensionality reduction, is implemented as a multi-threaded program. 
We implement random indexing (RI) as a random projection method, which extract a set of word vectors trough random projection of a co-occurrence matrix. 
We use the same method as in HPCA to build the co-occurrence matrix. 
Once the co-occurrence matrix is built, the low dimensional word vectors are generated through performing a sparse matrix product which can be done very fast. 

Principal word embedding acts similar to GloVe and our implementation of HPCA and RI. 
We first count the marginal frequency of words (and features if the corpus is annotated).
Then we build a co-occurrence matrix.
Finally, we build the low dimensional word vectors from this matrix using \ALG{alg:pwvec}. 
The first two steps are implemented with a single threaded problem.
The third step is implemented in Octave compiled with SuitSparse and OpenBLAS. 
The SuitSparse library is used to speed up the process of the sparse matrix operations in Line~\ref{alg:rsvd:smapling} and Line~\ref{alg:rsvd:qr} in \ALG{alg:rsvd}. 
The multi-threading functionality of OpenBLAS speeds up the basic matrix operations and the singular value decomposition in Line~\ref{alg:rsvd:svd} in \ALG{alg:rsvd}. 

In order to provide a fair comparison between these methods, we compare the methods with regard to their performance on the third step, the dimensionality reduction. 
\TABLE{table:cmp1} shows the time required to perform the dimensionality reduction with each of the methods. 
Since word2vec performs both scanning and dimensionality reduction together, we exclude it from this table. 
\begin{table}[htbp]
  \begin{center}
    \begin{tabularx}{\textwidth}{XXXXXXX}
    \hline
            & RI                & HPCA   & GLOVE    & PWE\\ \hline
    Sec.    & $\mathbf{180}$    & $480$  & $8040$   & $900$ \\ \hline
  \end{tabularx}
  \caption{The amount of time (seconds) required by each of the word embedding methods to perform the dimensionality reduction. PWE refer to the principal word embedding method introduced in this paper.}
  \end{center}
  \label{table:cmp1}
\end{table}
We see that the most efficient methods is RI. 
The principal word embedding is almost two times slower than HPCA. 
All of these methods require the same amount of time to scan the training corpus and to build the co-occurrence matrix. 
This takes around $2$ hours to build the vocabulary list and the co-occurrence matrix. 
On the other side, word2vec needs more than $10$ hours to generate the word vectors. 
So, principal word embedding is faster than GLOVE and word2vec but slower than RI and HPCA. 
This together with what shown in \TABLE{table:p-value} confirm that principal word embedding is an efficient method of word embedding which can generate a set of word vectors a good as GloVe and word2vec. 

\section{Conclusion}
\label{sec:conclusion}
Principal word embedding is a method of word embedding which generates a set of word vectors from a corpus through principal component analysis of contextual word vectors. 
A contextual word vector is defined as a representation of a word whose elements count the frequency of seeing the word in different contextual environments formed in a corpus. 
We have shown that the distribution of these vectors is not suitable for performing the classic principal component analysis.
Hence, a generalized version of principal component analysis is proposed to reshape the distribution of the word vectors through maximizing the entropy of contextual word vectors. 
The concept of corpus is also generalized in such a way to cover both raw and annotated corpora. 
To this end, we formulating both context and contextual feature (\ie annotation symbols into a random variable called features variable. 
We have shown that there is a direct relationship between the number of feature variables and the distribution of the contextual word vectors, hence the need for an \emph{adaptive} method of transformation.
In addition to these, we formulate a quadratic transformation which provides for weighting both the word observations and contextual features. 

These generalizations lead to a flexible method of word embedding that can generate word vectors with different types of contexts and contextual features using a reasonable amount of computational resources. 
In order to provide a better picture about the distribution and the type of information encoded in to the principal word vectors, two metrics are introduced to measure the amount of spread and the amount of discriminability of word vectors. 
We have shown that the spread and the separability of the word vectors highly depends on the type of feature variables. 
Our experiments on the dimensionality of word vectors show that the syntactic discriminability of word vectors is more sensitive to the number of dimensions of principal word vectors than their semantic discriminability. 
In comparison with the other methods of word embedding, the contribution of the principal word vectors to certain tasks is as good as or better than other sets of word vectors generated with the popular methods of word embedding. 
In terms of efficiency, the principal word embedding is faster than other methods of word embedding such as word2vec and GloVe, and slightly slower than the methods such as Hellinger PCA and random indexing.

\section{Future Work}
\label{sec:future_work}

The principal word vectors are formed by linear combination of the feature variables associated with a set of contextual features. 
This means that the dimensions of principal word vectors are interpretable in terms of the contextual word vectors. 
As a future work, this hints at providing linguistically motivate interpretations for each dimension of the principal word vectors. 
The linguistically motivated study of principal word vectors can also be led toward answering the questions such as \emph{why a certain type of feature variables result in higher amount of syntactic or semantic discriminability?}
The answers to this question are connected to another research area where one can investigate \emph{what type of annotations (contextual features) and contexts are more suitable for certain tasks?}
For example, for the task of part-of-speech tagging, we may face this question whether a set of word vectors extracted from a tagged corpus work better than a set of word vectors extracted from a raw corpus. 
One may also ask whether the dependency context work better than the neighbourhood context for the task of dependency parsing. 
Similar research question can be raised about the transformation function. 
Our experiments in this research were focused only on the power transformation. 
It would be interesting to see \emph{if other transformation functions such as Logarithm work as good as or better than the power transformation function}. 

Another research area is \emph{to replace the generalized PCA through SVD, with other approaches of principal component analysis such as the Hebbian PCA, the local PCA, and the kernel PCA.}
The Hebbian PCA enables the method to generate a set of word vectors without explicitly forming the contextual (co-occurrence) matrix. 
In this case, the method is expected to uses less amount of memory which is desirable in many environments. 
The local PCA provide for investigating the clusters of feature variables and the kernel PCA is helpful to see if different types of kernels can be meaningful to different tasks. 
Since the kernel matrix is in the quadratic order of the number of words, the classic kernel PCA might not be the best choice. 
In this case, the online methods of kernel PCA, such as Hebbian kernel PCA, seem to be more suitable since they do not form the kernel matrix implicitly. 

\starttwocolumn
\bibliographystyle{compling}
\bibliography{references}

\end{document}